\documentclass[lettersize,journal,compsoc]{IEEEtran}
\usepackage{amsmath,amsfonts}
\usepackage{algorithmic}
\usepackage{algorithm}
\usepackage{array}
\usepackage[caption=false,font=normalsize,labelfont=sf,textfont=sf]{subfig}
\usepackage{textcomp}
\usepackage{stfloats}
\usepackage{url}
\usepackage{verbatim}
\usepackage{graphicx}
\usepackage{cite}
\usepackage{booktabs} % For formal tables
\usepackage[table,xcdraw]{xcolor}
\usepackage{multirow}
\usepackage{dsfont}
\usepackage{bbm}
\usepackage{url}
\usepackage{ragged2e}
\usepackage{color,xcolor}
\usepackage{booktabs}
\usepackage{amsthm}
\usepackage{mathrsfs}
\usepackage{bm}
\usepackage{amssymb}
\usepackage{caption}
\usepackage{makecell}
\usepackage{balance}
\usepackage[colorlinks,
linkcolor=red,
anchorcolor=blue,
citecolor=blue
]{hyperref}
\usepackage{graphicx}
\usepackage{multirow}
\usepackage{siunitx}
\usepackage[normalem]{ulem}
\useunder{\uline}{\ul}{}
\usepackage{cuted}
\usepackage{utfsym}
\def\eg{e.g.}
\def\ie{i.e.}

\newcommand{\secref}[1]{Section~\ref{#1}}

\newcommand{\tabref}[1]{Table~\ref{tab:#1}}
\newcommand{\twotabref}[2]{Tables~\ref{tab:#1} and \ref{tab:#2}}

\newcommand{\eqnref}[1]{Eqn.~\ref{#1}}
\newcommand{\firef}[1]{Fig.~\ref{#1}}
\newcommand{\fourfigref}[4]{Figs.~\ref{#1},\ref{#2},\ref{#3},~and~\ref{#4}}
\newcommand{\appref}[1]{Appendix~\ref{app:#1}}

\renewcommand{\raggedright}{\leftskip=0pt \rightskip=0pt plus 0cm}

\newcommand{\rev}[1]{{#1}}
\newcommand{\pa}[1]{{#1}}

\definecolor{forestgreen}{rgb}{0.13, 0.55, 0.13}
\newcommand{\green}[1]{\textcolor{forestgreen}{#1}}
\newcommand{\red}[1]{\textcolor{red}{#1}}

\begin{document}

\title{Revisiting Essential and Nonessential Settings of Evidential Deep Learning}

\author{Mengyuan~Chen,
	Junyu~Gao,
	and~Changsheng~Xu,~\IEEEmembership{Fellow,~IEEE}
        % <-this % stops a space
    \IEEEcompsocitemizethanks{
       	\IEEEcompsocthanksitem Mengyuan Chen, Junyu Gao and Changsheng Xu are with National Lab of Pattern Recognition, Institute of Automation, Chinese Academy of Sciences, Beijing 100190, P. R. China, and with University of Chinese Academy of Sciences, Beijing, China, and also with Peng Cheng Laboratory, ShenZhen, China (e-mail: chenmengyuan2021@ia.ac.cn; junyu.gao@nlpr.ia.ac.cn; csxu@nlpr.ia.ac.cn).}
%\thanks{This paper was produced by the IEEE Publication Technology Group. They are in Piscataway, NJ.}% <-this % stops a space
\thanks{Manuscript received April 19, 2021; revised August 16, 2021.}}

\markboth{Journal of \LaTeX\ Class Files,~Vol.~14, No.~8, August~2021}%
{Shell \MakeLowercase{\textit{et al.}}: A Sample Article Using IEEEtran.cls for IEEE Journals}

\IEEEpubid{0000--0000/00\$00.00~\copyright~2021 IEEE}

\IEEEtitleabstractindextext{
\begin{abstract}
\raggedright{Evidential Deep Learning (EDL) is an emerging method for uncertainty estimation that provides reliable predictive uncertainty in a single forward pass, attracting significant attention.
Grounded in subjective logic, EDL derives Dirichlet concentration parameters from neural networks to construct a Dirichlet probability density function (PDF), modeling the distribution of class probabilities.
Despite its success, EDL incorporates several nonessential settings:
In model construction, (1)~a commonly ignored prior weight parameter is fixed to the number of classes, while its value actually impacts the balance between the proportion of evidence and its magnitude in deriving predictive scores.
In model optimization, (2)~the empirical risk features a variance-minimizing optimization term that biases the PDF towards a Dirac delta function, potentially exacerbating overconfidence.
(3)~Additionally, the structural risk typically includes a KL-divergence-minimizing regularization, whose optimization direction extends beyond the intended purpose and contradicts common sense, diminishing the information carried by the evidence magnitude.
Therefore, we propose Re-EDL, a simplified yet more effective variant of EDL, by relaxing the nonessential settings and retaining the essential one, namely, the adoption of projected probability from subjective logic.
Specifically, Re-EDL treats the prior weight as an adjustable hyperparameter \pa{rather than} a fixed scalar, and directly optimizes the expectation of the Dirichlet PDF provided by deprecating both the variance-minimizing optimization term and the \pa{divergence regularization term}. 
Extensive experiments and state-of-the-art performance validate the effectiveness of our method.
The source code is available at \url{https://github.com/MengyuanChen21/Re-EDL}.}
\end{abstract}

}

\maketitle

\section{Introduction}

%Estimating uncertainty allows models to provide an indication of how confident they are in their predictions, which plays a crucial role in decision-making of high-risk domains such as medical treatment~\cite{begoli2019need,abdar2022need} and autonomous driving~\cite{feng2018towards,choi2019gaussian}, improving training efficiency in scenarios of active learning~\citesun2022evidential,nguyen2022measure} and reinforcement learning~\cite{ahn2017uncertainty,an2021uncertainty}, detecting anomaly/out-of-distribution (OOD) testing samples~\cite{d2021pixel,linmans2023predictive}, etc. 
%However, the mainstream methods to obtain predictive uncertainty of deep learning models, such as Bayesian neural networks~\cite{hernandez2015probabilistic,dusenberry2020efficient} and deep ensembles~\cite{lakshminarayanan2017simple}, have substantial memory or computational demands, which constrain their suitability for real-time, large-scale industrial deployment.
%This limitation drives the interest of researchers in exploring methods which can achieve accurate uncertainty estimation without additional cost.

\IEEEPARstart{I}{n} high-risk domains such as autonomous driving and medical analysis, it is imperative for models to reliably convey the confidence or uncertainty level of their predictions~\cite{choi2019gaussian,abdar2022need}.
%However, previous research suggests that most modern deep neural networks (DNNs), especially when trained for classification via supervised learning, exhibit poor calibration, tending predominantly towards over-confidence~\cite{guo2017calibration}.
Despite effective uncertainty quantification methods based on Bayesian theory and ensemble techniques have been developed, these mainstream methods necessitate multiple forward passes in the inference phase~\cite{blundell2015weight,dusenberry2020efficient,gal2016dropout,lakshminarayanan2017simple,wen2020batchensemble,egele2022autodeuq}, imposing substantial computational burdens that hamper their widespread industrial adoption.
This limitation drives the interest of researchers in exploring how to achieve high-quality uncertainty estimation with minimal additional cost.

%Estimating uncertainty allows models to indicate how confident they are in their predictions, which plays a crucial role in decision-making of high-risk domains such as autonomous driving~\cite{choi2019gaussian} and medical treatment~\cite{abdar2022need}.
%The paramount challenge for deep neural nets in achieving reliable uncertainty estimation is the notorious over-confidence issue of modern network architectures.
%Previous research~\cite{guo2017calibration} demonstrates that most modern deep neural networks, especially when trained for classification in a supervised learning setting, are poorly calibrated, primarily in the direction of over-confidence.
%%Specifically, classification models exhibit undue confidence in misclassified or out-of-distribution samples, since they are optimized by one-hot encoded labels which concentrate the probability mass in a single class with zero entropy, thereby leaving no room for uncertainty regarding the input~\cite{thulasidasan2019mixup}.
%However, the mainstream methods of uncertainty quantification based on deep neural nets necessitate multiple forward passes in the inference phase~\cite{blundell2015weight,dusenberry2020efficient,gal2016dropout,lakshminarayanan2017simple,wen2020batchensemble}, imposing considerable computational burdens that hamper their widespread industrial adoption.
%This limitation drives the interest of researchers in exploring how to achieve high-quality uncertainty estimation with minimal additional cost.

Evidential deep learning (EDL)~\cite{sensoy2018evidential} is such a newly arising single-forward-pass uncertainty estimation method, which has attracted increasing attention for its success in various pattern recognition tasks~\cite{amini2020deep,bao2021evidential,qin2022deep,chen2022dual,oh2022improving,sun2022evidential,park2022active,sapkota2022adaptive,gao2023collecting}.
%Drawing upon the theory of subjective logic~\cite{josang2001logic,josang2016subjective}, EDL harnesses both the proportion of collected evidence among classes and their magnitude value to calibrate the predictive scores to better reflect the actual likelihood of correctness, thereby effectively mitigating model over-confidence and enhancing uncertainty estimation.
Drawing upon the subjective logic theory~\cite{josang2001logic,josang2016subjective}, \pa{EDL employs deep neural networks to derive Dirichlet concentration parameters, constructing a Dirichlet distribution that models the distribution of class probabilities and enables high-quality uncertainty estimation}.
%harnesses both the proportion of collected evidence among classes and their magnitude value
%to achieve high-quality uncertainty estimation, effectively mitigating model over-confidence on out-of-distribution samples.
Specifically, in $C$-class classification, EDL models the distribution of class probability~$\bm p_X$ with a constructed Dirichlet distribution $\text{Dir}(\bm p_X, \bm \alpha_X)$, whose concentration parameter vector $\bm \alpha_X(x)$ is given by
\begin{equation}
\label{bijective-edl}
\small
\bm \alpha_X(x) = \bm e_X(x) + C\cdot\bm a_X(x),
\quad\forall x\in\mathbb X=\{1,2,...,C\},
\end{equation}
where the base rate $\bm a_X$ is typically set as a uniform distribution over $\mathbb{X}$, and its scalar coefficient $C$ serves as a parameter termed as a \textit{prior weight}.
\IEEEpubidadjcol
Note that to keep the notation uncluttered, we use $\bm \alpha_X(x)$ as a simplified expression of $\bm \alpha_X(X=x)$, and similarly for $\bm e_X(x)$ and $\bm a_X(x)$.
The random variable $X$ denotes the class index of the input sample, and $\bm e_X(x)$ signifies the amassed evidence for the sample's association with class~$x$.
%The based rate $\bm a_X$ is typically assumed as a uniform distribution when no prior information is available.
%$W$ is a scalar termed prior weight, and in the EDL framework, it is simply set equal to the number of classes, $C$.
Thereafter, for model optimization, the traditional EDL method integrates the mean square error (MSE) loss over the class probability $\bm p_X$, which is assumed to follow the above Dirichlet distribution, thus deriving the empirical risk (average loss over training samples) as
\begin{equation}
\small
\label{edl-loss}
\begin{aligned}
	\mathcal{L}_\text{edl-emp}
	=&\frac{1}{|\mathcal{D}|}\sum_{(\bm z, \bm y)\in \mathcal{D}}\mathbb{E}_{\bm p_X\sim\text{Dir}(\bm p_X, \bm \alpha_X)}\left[\Vert \bm y - \bm p_X\Vert_2^2\right] \\
	=&\frac{1}{|\mathcal{D}|}\sum_{(\bm z, \bm y)\in \mathcal{D}}\sum_{x\in\mathbb X}\left(\bm y_x -\mathbb{E}_{\bm p_X\sim\text{Dir}(\bm p_X, \bm \alpha_X)}[\bm p_X(x)]\right)^2 \\
	&+ \text{Var}_{\bm p_X\sim\text{Dir}(\bm p_X, \bm \alpha_X)}[\bm p_X(x)],
\end{aligned}
\end{equation}
where the training set $\mathcal{D}$ consists of sample features and their one-hot labels denoted $(\bm z, \bm y)$\footnote{In deep learning, the sample feature is usually denoted by $\bm x$. However, to preclude ambiguity with $x$ as the value of the random variable $X$, we employ $\bm z$ instead to denote the sample feature. Random variable $X$, label $\bm y$, and feature $\bm z$ all pertain to the same input sample.}, and $\bm y_x$ refers to the $x$-th element of $\bm y$.
In addition, the structural risk (loss with extra regularization to mitigate over-fitting) of EDL-related methods typically include an additional regularization $\mathcal{L}_\text{kl}$,
% which has the following specific form:
\begin{equation}
	\small
	\label{kl-loss}
	\mathcal{L}_\text{kl}
	=\frac{1}{|\mathcal{D}|}\sum_{(\bm z, \bm y)\in \mathcal{D}}\text{KL}\left(\text{Dir}(\bm p_X, \tilde{\bm \alpha}_X), \text{Dir}(\bm p_X, \bm 1)\right),
\end{equation}
where $\bm 1$ denotes a $C$-dimensional ones vector, and $\tilde{\bm \alpha}_X=\bm y + (\bm 1 - \bm y)\odot \bm \alpha_X$ represents a modified Dirichlet parameter vector where the target class value is set to 1.
$\mathcal{L}_\text{kl}$ expects to minimize the Kullback-Leibler divergence between a uniform distribution and a modified Dirichlet distribution (hereafter referred to as \textit{KL-Div-minimizing} for short), thereby suppressing evidence of non-target categories.
%Besides, EDL adopts an annealing coefficient $\mu_t=\min(1.0, t/10)\in[0,1]$ for this regularization, where $t$ is the training epoch index.
%A more detailed introduction of the subjective logic theory and the traditional EDL method will be provided in \secref{preliminary}.

Despite the remarkable success of EDL, we argue that the existing EDL-based methods incorporate nonessential settings in \pa{both model construction and model optimization}.
These settings have been widely accepted by deep learning researchers; however, they are not intrinsically mandated by the mathematical framework of subjective logic and, based on our research, have minimal impact on enhancing uncertainty estimation in most cases.
Specifically, in \pa{model construction}, \textbf{(1)}
%the constructed Dirichlet distribution $\text{Dir}(\bm p_X, \bm \alpha_X)$,
the commonly ignored prior weight parameter in \eqnref{bijective-edl} governs the balance between capitalizing on the proportion of evidence and its magnitude when deriving predictive scores. However, EDL prescribes this parameter's value to be equivalent to the number of classes, potentially resulting in highly counter-intuitive outcomes.
Therefore, we advocate for setting the prior weight parameter as a free hyperparameter in the neural network to adapt to complex application cases.
\pa{In model optimization}, \textbf{(2)}~the empirical risk given by \eqnref{edl-loss} includes a variance-minimizing optimization term, which encourages the Dirichlet PDF modeling the distribution of probabilities to approach a Dirac delta function which is infinitely high and infinitesimally thin, or in other words, requires an infinite amount of evidence of the target class, thus further intensifying the over-confidence issue.
\textbf{(3)}~When designing the structural risk, EDL-related works commonly incorporate a KL-Div-minimizing regularization term $\mathcal{L}_\text{kl}$, as formulated in \eqnref{kl-loss}.
Our analysis suggests that its optimization direction extends beyond the intended purpose and contradicts common sense, diminishing the information carried by evidence magnitude.
%often hindering uncertainty estimation.
Therefore, we advocate for directly optimizing the expectation of the Dirichlet distribution towards one-hot labels, deprecating both the variance-minimizing optimization term and the \pa{KL-divergence} regularization to obtain more reliable predictive scores.
Note that our relaxations strictly adhere to the subjective logic theory.

Relaxing the nonessential EDL settings that are not mandated by subjective logic and often bring minimal benefit, we naturally become curious about the truly essential setting that contributes to the uncertainty estimation capability of EDL.
In this work, we identify the adoption of projected probability from subjective logic as the essential setting.
Specifically, compared to the traditional softmax operation, projected probability introduces an extra parameter to class scores and utilizes an output activation function characterized by a more gradual growth rate than the Exp function, both of which lead to more effective preservation of the magnitude information carried by model output logits.

In summary, as presented in \firef{method-graph}, we develop a simplified yet more effective EDL variant by deprecating the nonessential settings while retaining the essential ones.
To distinguish it from R-EDL~\cite{chen2024r}, the conference version which also relaxes EDL settings, we refer to the method introduced in this work as \textbf{Re-EDL}.
This name highlights its derivation from \textbf{Re}visiting EDL settings, while also indicating its relevance to the earlier R-EDL.
Our contributions include:
\vspace{-0.3em}
\begin{itemize}
\item An analysis of the commonly ignored prior weight parameter which balances the trade-off relationship between leveraging the proportion and magnitude of evidence in the subjective logic framework.

\item An analysis of the benefits of directly applying the MSE loss to the expected value of the constructed Dirichlet distribution as the empirical risk, rather than integrating the MSE loss over the class probabilities $\bm p_X$ drawn from this Dirichlet distribution.

\item An analysis of the optimization direction and practical impact of the KL-Div-minimizing regularization, which is commonly adopted in EDL’s structural risk but considered by us as nonessential since it often hinders uncertainty quantification by causing information loss in evidence amplitude.

\item An exploration of the truly essential setting in EDL, namely, the adoption of projected probability from subjective logic theory, which contributes to the superior uncertainty estimation capability within the context of our simple Re-EDL formulation.

\item Experiments on multiple benchmarks for uncertainty estimation, which comprehensively demonstrate the effectiveness of our proposed method under the classical, few-shot, video-modality, and noisy settings.
\end{itemize}

A preliminary version of this work~\cite{chen2024r} has been accepted for a Spotlight presentation at ICLR 2024.
In this paper, we extend our previous work both theoretically and empirically.
\textbf{Theoretically}, (1) \pa{we analyze that the commonly adopted KL-Div-minimizing regularization is a nonessential EDL setting, since its optimization extends beyond the intended purpose and contradicts common sense, typically causing information loss in evidence amplitude and hindering uncertainty estimation.}
(2) Additionally, \pa{we analyze that replacing traditional softmax classification head with projected probability is a truly essential EDL setting which contributes to the superior uncertainty estimation, since the components of projected probability better preserve the amplitude information of model logits.}
(3) \pa{Therefore, we propose Re-EDL by relaxing the nonessential EDL settings and retaining only the essential one, which achieves impressive simplicity and uncertainty estimation capability.}
\textbf{Empirically}, (4) we \pa{highlight a series of in-depth experiments which support our arguments regarding the essential and nonessential EDL settings}.
(5) Furthermore, we \pa{include more recent baseline methods and} expand the original experimental setup from \cite{chen2024r} by including four additional OOD datasets to ensure a comprehensive evaluation.
Key baseline methods are reproduced under consistent training settings, enhancing the fairness of comparisons.
Additional introductions, ablations, and visualizations are provided.

Due to space limitations, derivations, proofs, \pa{additional introductions, and extra results} are provided in Appendix.

\section{Related Work}
\label{related}

\noindent\textbf{Theoretical Extensions of EDL.}
A comprehensive introduction to EDL~\cite{sensoy2018evidential} is provided in \secref{edl}.
Here, we offer a brief overview of the subsequent developments of EDL.
%The theoretical advancements of EDL can be categorized into classification and regression based on the nature of the tasks.
%In classification tasks, 
Early research primarily focuses on enhancing the model's uncertainty estimation capabilities by incorporating additional OOD samples.
For instance, \cite{sensoy2020uncertainty} employs generative models to obtain proximal OOD samples, using the latent space of a variational autoencoder (VAE)~\cite{doersch2016tutorial} as a proxy for semantic similarity.
\cite{hu2021multidimensional} extends this approach by leveraging both proximal and distant OOD samples to further improve uncertainty estimation. 
Besides, \cite{nagahama2023learning,zhao2019quantifying} explores the use of pre-prepared OOD samples to enhance the performance of EDL models.
Recent advancements in EDL classification theory have primarily focused on exploring alternative methods for evidence collection.
For instance, \cite{shen2023post} introduces an innovative evidence collection paradigm that gathers evidence from multiple intermediate layers, rather than relying solely on the final network layer.
RED~\cite{pandey2023learn} conducts a deep analysis of evidential activation functions and proposes a novel regularizer that effectively addresses existing limitations.
$\mathcal{I}$-EDL, proposed by \cite{deng2023uncertainty}, incorporates the Fisher information matrix to assess the evidence informativeness carried by samples.
Additionally, HENN~\cite{li2024hyper} proposes a generalized variant of EDL, extending the multinomial subjective opinion characterized by EDL to a hypernomial version.
Furthermore, \cite{kandemir2022evidential} combines EDL, neural processes, and neural Turing machines to propose the Evidential Tuning Process, which shows stronger performances than EDL but requires a rather complex memory mechanism.
Compared with previous efforts, our method is the first to revisit essential and nonessential EDL settings, leading to a simplified yet superior EDL variant.
% while strictly adhering to subjective logic theory.
%while strictly adhering to the subjective logic theory.

\noindent\textbf{Widespread Applications of EDL.}
EDL has been widely applied across various downstream application fields, \pa{including computer vision~\cite{sensoy2020uncertainty,zhao2019quantifying,hu2021multidimensional,ghesu2019quantifying,ji2024spectral,shi2024evidential,wang2022uncertainty,lou2023elfnet,bao2021evidential,zhang2023learning,zhao2023open,zhu2022towards,bao2022opental,sun2022evidential1,chen2023uncertainty}, natural language processing~\cite{zhang2023ner,ren2022evidential,ren2023uncertaintyevent}, cross-modal learning~\cite{qin2022deep,li2023dcel,shao2024dual,li2024prototype}, and other scientific subjects such as medicine~\cite{ghesu2019quantifying}, physics~\cite{tan2023single}, chemistry~\cite{vazquez2024outlier}, etc.}
%For instance, DEAR~\cite{bao2021evidential} achieves impressive performances on open-set action recognition by proposing a novel model calibration method to regularize the EDL training.
%PAU~\cite{li2024prototype} induces accurate uncertainty for cross-modal retrieval by taking the variations of similarities between a visual instance and constructed learnable prototypes as belief masses.
\pa{From the perspective of machine learning paradigms,} except for supervised learning, EDL has also achieved success within active learning~\cite{shi2020multifaceted,park2022active,sun2022evidential}, transfer learning~\cite{chen2022evidential,aguilar2023continual,pei2024evidential,zhang2024revisiting}, reinforcement learning~\cite{yang2023uncertainty,wang2023deep}, weakly-supervised learning~\cite{gao2023vectorized,chen2023uncertainty,chen2023cascade}, few-shot learning~\cite{su2023hsic}, etc.
Besides, Deep Evidential Regression (DER) \cite{amini2020deep, soleimany2021evidential} successfully extends the application field of EDL to  regression by incorporating evidential priors into the Gaussian likelihood function, thereby enhancing uncertainty modeling in regression networks.
%\cite{} offers further insights into the empirical effectiveness of DER, even in the presence of over-parameterized representations of uncertainty.
Moreover,\cite{meinert2021multivariate,ma2021trustworthy,meinert2023unreasonable,ye2024uncertainty,wu2024evidence,oh2022improving,duan2024evidential,pandey2023evidential} also provides valuable explorations in DER.

% Additionally, Zhao et al. (2020) introduce a multi-source uncertainty framework that synergizes with DST for semi-supervised node classification using Graph Neural Networks (GNNs). Bao et al. (2022) advance a comprehensive framework for Open Set Temporal Action Localization (OSTAL), built upon the EDL methodology.

\noindent\textbf{Other single-model uncertainty methods based on DNNs.}
In addition to EDL-related works, various single-model methods exist for estimating predictive uncertainties.
Efficient ensemble methods~\cite{wen2020batchensemble, dusenberry2020efficient}, which cast a set of models under a single one, show state-of-the-art performances on large-scale datesets. While these methods are parameter-efficient, they necessitate multiple forward passes during inference.
Bayesian Neural Networks (BNNs)\cite{ritter2018scalable, izmailov2021bayesian} model network parameters as random variables and quantify uncertainty through posterior estimation while suffering from a significant computational cost. 
A widely-recognized method is Monte Carlo Dropout~\cite{gal2016dropout}, which interprets the dropout layer as a random variable following a Bernoulli distribution, and training a neural network with such dropout layers can be considered an approximation to variational inference. 
Two other notable single-forward-pass methods, DUQ~\cite{van2020uncertainty} and SNGP~\cite{liu2020simple}, introduce distance-aware output layers using radial basis functions or Gaussian processes.
Although nearly competitive with deep ensembles in OOD benchmarks, these methods entail extensive modifications to the training procedure and lack easy integration with existing classifiers.
Another group of efficient uncertainty methods are Dirichlet-based uncertainty (DBU) methods, to which EDL also belongs.
Prominent DBU methods encompass KL-PN \cite{malinin2018predictive}, RKL-PN~\cite{malinin2019reverse}, PostN~\cite{charpentier2020posterior}, and NatPN~\cite{charpentier2021natural}, which vary in both the parameterization and the training strategy of the Dirichlet distribution.
%Prominent DBU methods encompass KL-PN \cite{malinin2019reverse} optimizes a prior network with dual Kullback-Leibler (KL) divergence terms; RKL-PN~\cite{malinin2019reverse} incorporates reverse KL divergence to circumvent issues associated with undesired multi-modal target distributions; and Posterior Network~\cite{charpentier2020posterior} employs normalizing flows obviate the need for OOD data during training. 
Compared to these preceding methods, our approach combines the benefits of exhibiting favorable performances, being single-forward-pass, parameter-efficient, and easily integrable.
% and capable of distinguishing various types of uncertainty.

\noindent\textbf{Comparing Subjective Logic with other Uncertainty Reasoning Frameworks.} Please refer to \appref{comparison}.

\section{Preliminary: From Subjective Logic to Evidential Deep Learning}
\label{preliminary}
The essence of Evidential Deep Learning (EDL) lies in employing DNNs as analysts within the framework of subjective logic.
In this section, we briefly introduce core concepts of subjective logic theory (\secref{subjective-logic}), and outline the primary steps involved in developing EDL (\secref{edl}).
This introduction aims to differentiate between the fundamental theoretical requirements and the optional practical implementations, facilitating the subsequent discussion on the essential and nonessential settings of EDL (\secref{redl}).

\subsection{Subjective Logic Theory}
\label{subjective-logic}
Just as the names of \textit{binary} logic and \textit{probabilistic} logic imply, an argument in binary logic must be either true or false, and while probabilistic logic allows for probabilities within the range $[0,1]$ to express partial true.
However, both binary logic and probabilistic logic deal with definite arguments and do not provide a mechanism to express uncertainty or indifference, such as saying \textit{``I don't know"}.
To address this limitation, subjective logic~\cite{josang2001logic,josang2016subjective} extends probabilistic logic by explicitly including uncertainty about probabilities in the formalism.
Specifically, an argument in subjective logic, also called a \textit{subjective opinion}, is formalized as follows:

\textbf{Definition 1 (Subjective opinion).}
Given a categorical random variable $X$ on the domain $\mathbb{X}$, a subjective opinion over $X$ is defined as the ordered triplet ${\bm \omega}_X=({\bm b}_X, u_X, {\bm a}_X)$, where ${\bm b}_X$ is a \textit{belief mass} distribution over $X$, $u_X$ is a \textit{uncertainty mass}, ${\bm a}_X$ is a \textit{base rate}, aka a prior probability distribution over $X$, and the additivity requirements $\sum_{x\in\mathbb{X}} \bm b_{X}(x) + u_X = 1$ and $\sum_{x\in\mathbb{X}} \bm a_{X}(x) = 1$ are satisfied.

Belief mass $\bm b_{X}(x)$ assigned to a singleton value $x\in\mathbb{X}$ expresses support for the statement $X=x$ being TRUE, and uncertainty mass can be interpreted as belief mass assigned to the entire domain.
Therefore, subjective logic also provides a well-defined \textit{projected probability}, which follows the additivity requirement of traditional probability theory, by reassigning the uncertainty mass into each singleton of domain $\mathbb X$ according to the base rate $\bm a_X$ as follows:

\textbf{Definition 2 (Projected probability).}
%Let ${\bm \omega}_X=({\bm b}_X, u_X, {\bm a}_X)$ be a subjective opinion.
The projected probability ${\bm P}_X$ of the subjective opinion ${\bm \omega}_X=({\bm b}_X, u_X, {\bm a}_X)$ is defined by ${\bm P}_X(x) = {\bm b}_X(x) + {\bm a}_X(x) u_X$, $\forall x\in \mathbb{X}$. Note that the additivity requirement $\sum_{x\in\mathbb{X}} \bm P_{X}(x) = 1$ is satisfied.

Furthermore, the subjective logic theory points out that, if the base rate $\bm a_X$ and a parameter termed prior weight, denoted as $W$, is given, there exists a bijection between a multinomial opinion and a Dirichlet probabilistic density function (PDF) as presented in Theorem 1.
This relationship emerges from interpreting second-order uncertainty by probability density, and plays an important role in the formalism of subjective logic since it provides a calculus reasoning with PDFs.
The proof is provided in \appref{theorem1}.

\textbf{Theorem 1 (Bijection between subjective opinions and Dirichlet PDFs).}
Consider a random variable $X$ defined on the domain $\mathbb{X}$, and let ${\bm \omega}_X=({\bm b}_X, u_X, {\bm a}_X)$ represent a subjective opinion.
Denote by $\bm p_X$ a probability distribution over $\mathbb{X}$, and let the Dirichlet PDF with concentration parameter $\bm \alpha_X$ be denoted as $\text{Dir}({\bm p_X, \bm \alpha_X})$, where $\bm \alpha_X(x) \geq 0$ and $\bm p_X(x)\neq0$ when $\bm \alpha_X(x) <1$.
Given the base rate $\bm a_X$, there exists a bijection $F$ that maps the subjective opinion $\bm \omega_X$ to the Dirichlet PDF $\text{Dir}({\bm p_X, \bm \alpha_X})$ as follows:
\begin{equation}
\label{bijective1}
\small
\begin{aligned}
F:{\bm \omega}_X &= ({\bm b}_X, u_X, {\bm a}_X) \mapsto \\
\text{Dir}({\bm p_X, \bm \alpha_X}) &=
\frac{\Gamma\left(\sum_{x\in\mathbb{X}}\bm \alpha_X(x)\right)}
{\prod_{x\in\mathbb{X}}\Gamma(\bm \alpha_X(x))}
\prod_{x\in\mathbb{X}}\bm p_X(x)^{\bm \alpha_X(x)-1},
\end{aligned}
\end{equation}
where $\Gamma$ denotes the Gamma function, and $\bm \alpha_X$ satisfies that
\begin{equation}
	\small
\label{bijective2}
\bm \alpha_X(x) = \frac{\bm b_X(x) W}{u_X} + \bm a_X(x)W,
\quad\forall x\in\mathbb X,
\end{equation}
and $W\in\mathbb{R}_+$ is a scalar called a prior weight, whose setting will be further discussed in \secref{relax-dirichlet}.

\subsection{Evidential Deep Learning}
\label{edl}
Based on subjective logic, \cite{sensoy2018evidential} proposes a single-forward-pass uncertainty estimation method named Evidential Deep Learning (EDL), which lets deep neural networks play the role of analysts to give belief mass and uncertainty mass of samples. For example, in $C$-class classification, the belief mass $\bm b_X$ and uncertainty mass $u_X$ of the input sample, whose category index is a random variable $X$ taking values $x$ from the domain $\mathbb X=[1,...,C]$, are given by
\begin{equation}
	\small
	\label{edl-mass}
	\bm b_X(x) = \frac{\bm e_X(x)}{S_X},\quad u_X= \frac{C}{S_X},\quad S_X=\sum_{x\in \mathbb X}\bm e_X(x) + C.
\end{equation}
%for $\forall x\in \mathbb X$.
Specifically,  $\bm e_X(x)$, which denotes the \textit{evidence} of the random variable $X$ taking the value $x$, is the $x$-th element of the evidence vector $\bm e_X=f(g(\bm z))\in\mathbb R^C_+$, where $\bm z$ is the feature of the input sample, $g$ is a deep neural network, $f$ is a non-negative output activation function (\eg, softplus), sometimes also called \textit{evidence function}, and the scalar $C$ in this equation serves as the prior weight.

According to Theorem 1, there exists a bijection between the Dirichlet PDF denoted $\text{Dir}_X(\bm p_X, \bm \alpha_X)$ and the opinion $\bm \omega_X=(\bm b_X, u_X, \bm a_X)$ if the requirement in \eqnref{bijective2} is satisfied.
Substituting \eqnref{edl-mass} into \eqnref{bijective2} and setting the prior weight $W$ in \eqnref{bijective2} as $C$, we obtain the relationship between the parameter vector of the Dirichlet PDF and the collected evidence in EDL, as expressed by \eqnref{bijective-edl}.
%\begin{equation}
%	\label{bijective-edl}
%	\bm \alpha_X(x) = \bm e_X(x) + \bm a_X(x)W,
%	\quad\forall x\in\mathbb X.
%\end{equation}
Moreover, since EDL sets the base rate $\bm a_X(x)$ as a uniform distribution,
%and the prior weight $W$ as the cardinality of domain $\mathbb X$, aka the class number $C$,
the relationship given by \eqnref{bijective-edl} can be further simplified into $\bm \alpha_X(x)	= \bm e_X(x) + 1$, $\forall x\in \mathbb X$.

To perform model optimization, EDL integrates the MSE loss function over the class probability $\bm p_X$ which is assumed to follow the Dirichlet PDF specified in the bijection, thus derives the empirical risk given by \eqnref{edl-loss}.
Moreover, the structural risk of EDL-related methods typically includes a KL-Div-minimizing regularization, formulated in \eqnref{kl-loss}, to suppress evidence of non-target categories.
The detailed derivations are provided in \appref{optimization-objective-app}.
In inference, EDL utilizes the projected probability $\bm P_X$ (refer to Definition~2),
\begin{equation}
	\small
	\label{edl-probability}
	\bm P_X(x)={\bm b}_X(x) + {\bm a}_X(x) u_X=\frac{\bm e_X(x) + 1}{S_X}=\frac{\bm \alpha_X(x)}{S_X},
\end{equation}
%where $S_X=\sum_{x\in \mathbb X}\bm e_X(x) + C=\sum_{x\in \mathbb X}\bm \alpha_X(x)$,
as the predictive scores, and uses \eqnref{edl-mass} to calculate the uncertainty mass $u_X$ as the uncertainty of classification.
%\begin{equation}
%\label{edl-uncertainty}
%u_X= \frac{C}{\sum_{x\in \mathbb X}\bm e_X(x) + C}=\frac{C}{S_X},
%\end{equation}
%where $S_X=\sum_{x\in \mathbb X}\bm e_X(x) + C$ is the sum of $\bm \alpha_X(x)$ over $x\in\mathbb X$.

\section{Re-EDL: Revisit Essential and Nonessential Settings of EDL}
\label{redl}

Despite the significant success of EDL, we argue that the existing EDL-based methodologies (\secref{edl}) retain several rigid settings, which, while widely accepted, are not intrinsically mandated within subjective logic (\secref{subjective-logic}) and typically offer minimal benefit to uncertainty quantification.
Specifically, in this section, we identify the following nonessential EDL settings:
In model construction, \textbf{(1)} a \textbf{prior weight} parameter is fixed to the number of classes, suppressing the model's ability to adjust the impact of evidence magnitude and its proportion on predictions (\secref{relax-dirichlet});
\pa{In model optimization}, \textbf{(2)}~for the \textbf{empirical risk} (average loss over training samples), traditional EDL adopts the expected value of the MSE loss over the constructed Dirichlet distribution rather than directly applying MSE to the expected value of the Dirichlet distribution.
This results in an additional variance-minimizing term, which exacerbates overconfidence. (\secref{simplified-loss});
\textbf{(3)}~For the \textbf{structural risk} (loss with extra regularization terms), a commonly adopted KL-Div-minimizing regularization on non-target evidence limits model's complexity but hampers uncertainty estimation in most situations (\secref{kl-div-loss}).
After relaxing the above nonessential EDL settings while strictly adhering to subjective logic, we further provide an in-depth discussion about the truly essential EDL settings which contribute to the superior uncertainty estimation capability (\secref{essential}).

\vspace{-0.5em}
\subsection{Relaxing the Rigid Setting of Fixing Prior Weight in \pa{Model} Construction}
\label{relax-dirichlet}
In this subsection, we elucidate how the prior weight $W$ balances between leveraging the proportion and magnitude of evidence to compute predictive scores. Conclusively, we argue against the rigidity of fixing $W$ to the class number and propose viewing it as an adjustable hyperparameter.

The nomenclature of \textbf{prior weight} comes from the expression of \eqnref{bijective-edl}. The scalar coefficient $C$, functioning as the prior weight $W$, denotes the weight of the base rate $\bm a_X$, which is alternatively termed the \textbf{prior distribution}.
In Theorem 1, it should be noted that the bijection between subjective opinions and Dirichlet PDFs is only specified when the base rate $\bm a_X$ and the prior weight $W$ are given.
Typically, in the absence of prior information, we default to setting the base rate as a uniform distribution over the domain $\mathbb{X}$, \ie, $\bm a_X(x)=1/|X|=1/C$, $\forall x\in\mathbb X$.
However, the setting of the prior weight $W$ is worth further discussion.

\begin{figure}[]
	\centering
	\includegraphics[width=1.0\linewidth]{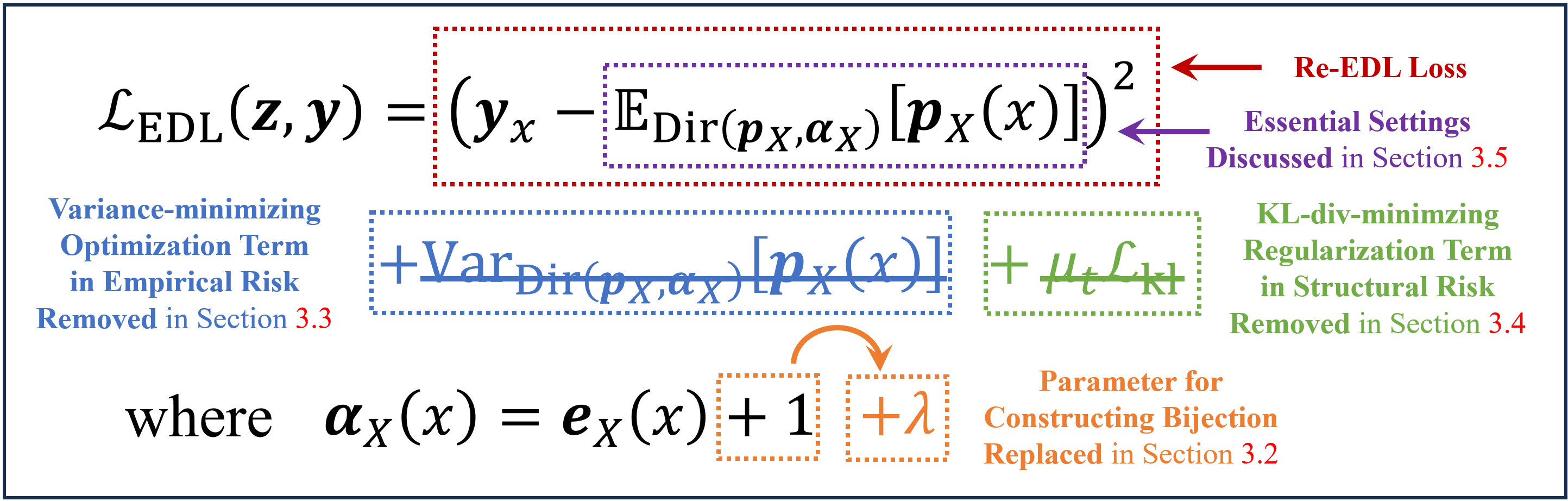}
	\vspace{-0.5em}
	\caption{A conceptual framework diagram illustrating the essential and nonessential settings discussed in \secref{redl}. By relaxing the nonessential settings while retaining the essential ones, the proposed Re-EDL method achieves more superior uncertainty estimation with reduced complexity.}
	\label{method-graph}
	\vspace{-0.5em}
\end{figure}

We argue that fixing the prior weight to the cardinality of the domain, which is widely adopted by EDL researchers, is not intrinsically mandated by subjective logic and may result in counter-intuitive results.
For example, a 100-class classification task forces $W=100$. Even though the neural net gives an extreme evidence distribution $\bm e=[100,0,0,....,0]\in\mathbb R^{100}_+$, EDL will reach the prediction that the probability of the sample belonging to Class 1 is $P=(100+1)/(100+100)\approx0.5$ by \eqnref{edl-probability}, which is highly counter-intuitive.
The underlying reason is that the value of $W$ dictates the degree to which the projected probability is influenced by the \textbf{magnitude} of the evidence or contrarily the \textbf{proportion} of the evidence.
To elucidate this point more clearly, we first revisit \eqnref{edl-mass} and \eqnref{edl-probability} without fixing the prior weight $W$ to $C$. In this way, we can obtain a generalized form of the projected probability $\bm P_X$ as
\begin{equation}
\small
\label{sll-probability}
\bm P_X(x)={\bm b}_X(x) + {\bm a}_X(x) u_X=\frac{\bm e_X(x) + \frac{W}{C}}{\sum_{x^\prime\in \mathbb X}\bm e_X(x^\prime) + W}.
\end{equation}
When the prior weight $W$ is set to zero, the projected probability $\bm P_X$ in \eqnref{sll-probability} degenerates to a conventional probability form, which solely relies on the proportion of evidence among classes and is unaffected by their magnitude, as scaling the evidence by a constant coefficient has no impact on $\bm P_X$.
However, when $W$ is not zero, we have
\begin{equation}
\label{prob-upper-bound}
\small
\bm P_X(x)
\leq \frac{\bm e_X(x) + \frac{W}{C}}{\bm e_X(x) + W}
=1 - (1 - \frac1C)\cdot\frac{1}{\bm e_X(x)/W + 1},
\end{equation} 
where the equlity holds only if $\sum_{x^\prime\in\mathbb X, x^\prime\neq x}\bm e_X(x^\prime)=0$.
\eqnref{prob-upper-bound} indicates that with extreme evidence distributions, where evidence for all classes except class $x$ is zero, the upper bound of $\bm P_X(x)$ is determined by the ratio of the evidence for class $x$ to the prior weight $W$. 
In other words, given the prior weight $W$, the upper bound of $\bm P_X(x)$ purely relies on the magnitude of $\bm e_X(x)$, with a lower magnitude creating a larger gap between $\bm P_X(x)$'s upper bound and $1$.

From the two cases presented above, it becomes evident that the value of $W$ determines the extent to which the projected probability $\bm P_X(x)$ is influenced by the magnitude and proportion of evidence respectively.
Specifically, a small $W$ implies that $\bm P_X(x)$ is predominantly influenced by the proportion of evidence distribution, whereas a large $W$ leads $\bm P_X(x)$ to mainly considering the magnitude of the evidence while overlooking the evidence proportion.

%\cite{sensoy2018evidential} demonstrated that both the magnitude and the proportion of model outputs are valuable for obtaining well-calibrated predictive results.
%Intuitively speaking, for any specific case, there should exist an optimal value for $W$ which can balance the inherent trade-off between leveraging the proportion of evidence and its magnitude to obtain predictive scores best reflecting the actual likelihood of correctness, thus minimizing the model over-confidence.
Intuitively speaking, for any specific case, there should exist an optimal value for $W$ which can balance the inherent trade-off between leveraging the proportion of evidence and its magnitude to obtain predictive scores minimizing model over-confidence.
%on misclassified and out-of-distribution samples.
However, it is unlikely that such an optimal value is universally applicable to all scenarios, given the myriad of complex factors influencing the network's output.
Hence, we advocate for relinquishing the rigidity of assigning the number of classes to $W$, but instead, treating $W$ as an adjustable hyperparameter within the neural network.
Therefore, as presented in \firef{method-graph}, we revisit \eqnref{bijective2} to derive a generalized form of the concentration parameter $\bm \alpha_X$ of the constructed Dirichlet PDF:
\begin{equation}
\small
\label{alpha-evidence}
\bm \alpha_X(x) = \left(\frac{\bm e_X(x)}{W} + \frac{1}{|\mathbb X|}\right)W
= \bm e_X(x) + \lambda,
\end{equation}
where $\lambda = W/C\in\mathbb R_+$ is a hyperparameter.
Note that the projected probability retains the same form as in \eqnref{edl-probability}, \ie, $\bm P_X(x)=\bm \alpha_X(x)/S_X$, while the form of uncertainty mass is transformed into $u_X=\lambda C/S_X$.
%Note that both the projected probability and the uncertainty mass retain the same form as in \eqnref{edl-probability}, \ie, $\bm P_X(x)=\bm \alpha_X(x)/S_X$ and $u_X=C/S_X$, when represented by $\bm \alpha_X(x)$ and $S_X$.

\vspace{-0.5em}
\subsection{Deprecating the Variance-minimizing Optimization Term in Empirical Risk}
\label{simplified-loss}

With the above generalized form of the concentration parameter, this subsection elaborates on our simplified EDL empirical risk, which directly optimizes the expectation of the constructed Dirichlet distribution, \ie, the projected probability $\bm P_X$.
In contrast, traditional EDL uses the MSE loss's expectation over the Dirichlet distribution as empirical risk, resulting in an additional variance-minimizing optimization term that potentially exacerbates overconfidence.

With the generalized setting of $\bm \alpha_X$ in \eqnref{alpha-evidence}, the projected probability $\bm P_X$ has the following variant:
\begin{equation}
	\small
\label{projected}
\bm P_X(x) = \frac{\bm \alpha_X(x)}{S_X}=\frac{\bm e_X(x) + \lambda}{\sum_{x^\prime\in \mathbb X}\bm e_X(x^\prime) + C\lambda},\quad\forall x\in \mathbb X.
\end{equation}
Consequently, by substituting the class probability in traditional MSE loss with the projected probability $\bm P_X$ in \eqnref{projected}, we seamlessly derive our empirical risk denoted $\mathcal{L}_\text{re-edl-emp}$:
\begin{equation}
\label{sll-loss}
\small
\begin{aligned}
	\mathcal{L}_\text{re-edl-emp}
	&=\frac{1}{|\mathcal{D}|}\sum_{(\bm z, \bm y)\in \mathcal{D}}\sum_{x\in\mathbb X} \left(\bm y_x - \bm P_X(x)\right)^2.
\end{aligned}
\end{equation}
Regarding the reason for adopting this formulation, we contend that the projected probability $\bm P_X$ has the unique property of alleviating the overconfidence typically arising from optimization toward the hard one-hot labels $\bm y$.
%Fortunately, the subjective logic theory offers an objective suitable for optimization toward the hard one-hot labels $\bm y$ without inducing significant over-confidence, that is, the projected probability $\bm P_X=\bm b_X + \bm a_X u_X$.
As previously noted, the projected probability $\bm P_X$ harnesses both the magnitude and proportion of collected evidence to more accurately represent the actual likelihood of a given output.
From an optimization perspective, compared to the proportion of evidence among classes, \ie, $\bm e_X(x)/\sum_x \bm e_X(x)$, or the belief mass $\bm b_X$, the projected probability $\bm P_X$ has more tolerance towards the existence of the uncertainty mass $u_X$, since $u_X$ also contributes to the projected probability $\bm P_X$ according to the base rate $\bm a_X$.
In other words, the item $\bm a_X u_X$ alleviates the urgency of the projected probability $\bm P_X$ tending to the one-hot label $\bm y$ when the model has not collected enough evidence, since the uncertainty mass $u_X$ is inversely proportional to the total amount of evidence, thus mitigating the over-confidence issue to some extent.
%Specifically, with the generalized setting of $\bm\alpha_X$ in \secref{relax-dirichlet}, the optimization objective with respect to the projected probability has the following form:
%\begin{equation}
%	\label{sll-loss}
%	\begin{aligned}
%		\mathcal{L}_\text{r-edl}
%		&=\frac{1}{|\mathcal{D}|}\sum_{(\bm z, \bm y)\in \mathcal{D}}\sum_{x\in\mathbb X} \left(\bm y_x - \bm P_X(x)\right)^2 \\
%		&=\frac{1}{|\mathcal{D}|}\sum_{(\bm z, \bm y)\in \mathcal{D}}\sum_{x\in\mathbb X} \left(\bm y_x - \frac{\bm e_X(x) + \lambda}{\sum_{x\in \mathbb X}\bm e_X(x) + C\lambda}\right)^2.
%	\end{aligned}
%\end{equation}

Meanwhile, \eqnref{sll-loss} can be interpreted as encouraging the expectation of the Dirichlet distribution to converge to the provided label, since the bijection introduced in Theorem~1 has been established on the following identity:
\begin{equation}
\label{base-identity}
\bm P_X(x)=\mathbb{E}_{\bm p_X\sim\text{Dir}(\bm p,\bm \alpha)}[\bm p_X(x)],
\end{equation}
which can be easily derived from \eqnref{projected} and the property of Dirichlet distributions.
Therefore, by substituting \eqnref{base-identity} into \eqnref{sll-loss} and then comparing it with \eqnref{edl-loss}, we can find that the essential difference between the two empirical risks is that, EDL uses \textbf{the expectation of the traditional MSE loss over the constructed Dirichlet PDF}, while our proposed Re-EDL directly applies \textbf{MSE on the expectation of the Dirichlet PDF}.
Therefore, \pa{as shown in \firef{method-graph}}, the optimization term $\mathcal{L}_\text{var}$, which attempts to minimize the variance of the Dirichlet distribution, is deprecated:
\begin{equation}
\small
\label{variance}
\begin{aligned}
\mathcal{L}_\text{var}
&=\frac{1}{|\mathcal{D}|}\sum_{(\bm z, \bm y)\in \mathcal{D}}\sum_{x\in\mathbb X}\text{Var}_{\bm p_X\sim\text{Dir}(\bm p_X, \bm \alpha_X)}[\bm p_X(x)] \\
&=\frac{1}{|\mathcal{D}|}\sum_{(\bm z, \bm y)\in \mathcal{D}}\frac{S^2_X - \sum_{x\in\mathbb X}\bm\alpha_X^2(x)}{S_X^2(S_X + 1)}.
\end{aligned}
\end{equation}

Let us delve deeper into this variance-minimizing optimization term.
When the variance of a Dirichlet distribution is close to zero, the Dirichlet probability density function is in the form of a Dirac delta function which is infinitely high and infinitesimally thin.
%In other words, the Dirichlet distribution which models the distribution of first-order probabilities would gradually degenerate to a traditional point estimation of first-order probabilities when its variance approaches zero.
Consequently, in the entire training phase, $\mathcal{L}_\text{var}$ 
%tends to suppress the uncertainty mass to zero and simultaneously 
keeps requiring an infinite amount of evidence of the target class, which further intensifies the serious over-confidence issue we seek to mitigate.
From another perspective, the Dirichlet distribution which models the distribution of first-order probabilities would gradually degenerate to a traditional point estimation of first-order probabilities when its variance approaches zero, thus losing the advantage of subjective logic in modeling second-order uncertainty.
Therefore, we posit that omitting $\mathcal{L}_\text{var}$ contributes to alleviating the over-confidence issue which commonly results in suboptimal uncertainty estimation, while preserving the merits of subjective logic. 
Our ablation study further corroborates this assertion.

\vspace{-0.5em}
\subsection{Delving into KL-divergence-minimizing Regularization on Non-target Evidence in Structural Risk}
\label{kl-div-loss}

Originating from the pioneer work~\cite{sensoy2018evidential}, a Kullback-Leibler (KL) divergence-minimizing regularization is commonly incorporated into the structural risk of EDL-related methods~\cite{ren2022evidential,qin2022deep,aguilar2023continual,deng2023uncertainty,pei2024evidential,zhang2024revisiting,han2022trusted,xie2023exploring,fan2023flexible,li2023dcel,sapkota2023adaptive}, suppressing the evidence for non-target classes.
\pa{While this regularization can marginally improve classification accuracy with a carefully tuned coefficient and enhance generalization on noisy data, we regard it as nonessential, since it is not mandated by subjective logic and typically brings minimal benefit on uncertainty estimation.}
%While it can marginally improve classification accuracy with a carefully tuned coefficient and enhance generalization on noisy test data, its necessity as a standard process in EDL remains debatable.
In this subsection, we initially present the vanilla form of this regularization and its variant with our relaxation proposed in \secref{relax-dirichlet}, and then offer an analysis of both the \pa{optimization direction} and practical impact, supported by experimental results.

In the traditional EDL method, after the target class value of the Dirichlet parameter vector being set to 1, the KL divergence between the modified Dirichlet distribution and a uniform distribution is expected to be minimized by an auxiliary regularization, which has the following form:
\begin{equation}
	\small
	\label{vanilla-kl}
	\mathcal{L}_\text{kl}
	=\frac{1}{|\mathcal{D}|}\sum_{(\bm z, \bm y)\in \mathcal{D}}\text{KL}\left(\text{Dir}(\bm p_X, \tilde{\bm \alpha}_X), \text{Dir}(\bm p_X, \bm 1)\right),
\end{equation}
where $\bm 1$ denotes a $C$-dimensional ones vector, and $\tilde{\bm \alpha}_X=\bm y + (\bm 1 - \bm y)\odot \bm \alpha_X$ represents the modified Dirichlet parameter vector.
Besides, an annealing coefficient $\mu_t=\min(1.0, t/10)\in[0,1]$ is adopted for this regularization, where $t$ is the training epoch index.
In the case of adopting the relaxation of the prior weight $W$ in \eqnref{alpha-evidence}, the Dirichlet distribution bijective to the vacuous opinion $\hat{\bm \omega}_X=(\bm b_X=\bm 0, u_X=1)$ is no longer a uniform distribution.
Specifically, it is parameterized by $\lambda\cdot\bm 1$, and $\mathcal{L}_\text{kl}$ is modified into:
\begin{equation}
	\label{redl-kl}
	\small
	\begin{aligned}
	\mathcal{L}_\text{kl}(\lambda)
	&=\frac{1}{|\mathcal{D}|}\sum_{(\bm z, \bm y)\in \mathcal{D}}\text{KL}\left(\text{Dir}(\bm p_X, \tilde{\bm \alpha}^\lambda_X), \text{Dir}(\bm p_X, \lambda\cdot\bm 1)\right) \\
	&\approx\frac{1}{|\mathcal{D}|}\sum_{(\bm z, \bm y)\in \mathcal{D}} \bigg(\log\frac{\Gamma(\tilde{S}^\lambda_X)}{\prod_{x\in\mathbb X}\Gamma\left(\tilde{\bm \alpha}_X^\lambda(x)\right)} + \\
	&\sum_{x\in\mathbb X}\left(\tilde{\bm \alpha}_X^\lambda(x)-\lambda\right)
	\left(\psi\left(\tilde{\bm \alpha}_X^\lambda(x)\right) - \psi(\tilde{S}^\lambda_X)\right)\bigg),
	\end{aligned}
\end{equation}
%which is adopted by R-EDL~\cite{chen2024r}.
where $\Gamma$ and $\psi$ denote the gamma and digamma functions,
$\tilde{\bm \alpha}_X^\lambda=\lambda\bm y + (\bm 1 - \bm y)\odot \bm \alpha_X$,
%represents a modified Dirichlet parameter vector whose value of the target class has been set to $\lambda$,
and $\tilde{S}_X^\lambda=\sum_{x\in\mathbb X} \tilde{\bm \alpha}^\lambda_X(x)$.
The detailed derivation is in \appref{optimization-objective-app}.
\pa{With the introduction complete, we proceed to conduct an in-depth analysis of the KL-Div-minimizing regularization from following aspects.}

%Despite \cite{deng2023uncertainty} elucidates the effectiveness of this regularization with the PAC-Bayesian theory~\cite{mcallester1998some}, some works have questioned its pratical effectiveness and many other EDL applications deprecate it.
%This regularization term has demonstrated promising empirical results and has been elucidated by \cite{deng2023uncertainty} using the PAC-Bayesian theory~\cite{mcallester1998some}.

\textbf{Theoretically, does the optimization direction of this regularization align with the common sense?}
%Probably no.
It is unlikely to be the case.
As introduced by \cite{sensoy2018evidential}, this regularization is motivated by the principle that \textit{``we prefer the total evidence to shrink to zero for a sample if it cannot be correctly classified"}.
However, the actual impact of $\mathcal{L}_\text{kl}$ extends far beyond the motivation:
Since both \eqnref{vanilla-kl} and \eqnref{redl-kl} achieve their unique minimum value when $\tilde{\bm \alpha}_X = \bm 1$ or $\tilde{\bm \alpha}_X^\lambda = \lambda \cdot \bm1$, this regularization not only affects misclassified samples but also encourages the evidence for all non-target classes to approach zero for all training samples.
As clearly shown in \tabref{evidence}, the magnitudes of non-target evidence diminishes substantially when $\mathcal{L}_\text{kl}$ is adopted.
%However, the ratio of target evidence to the sum of evidence for all non-target categories surpasses 100 with $\mathcal{L}_\text{kl}$, while without  $\mathcal{L}_\text{kl}$ the ratio is about 25.
We argue that this optimization direction contradicts common sense, particularly on challenging samples.
For instance, it is reasonable for a network model or a human analyst to gather some evidence supporting the classification of a poorly written digit ``3" as an ``8" or categorizing a dog-like ``cat" as a ``dog" due to their shared visual patterns.
In other words, since there is an inevitable overlap in visual patterns among different categories (otherwise, classification would be extremely easy), the situation where all non-target evidence, as a proxy of non-target belief mass, shrinks to zero should not occur.
%When non-target evidence vanishes, the information carried by their magnitude is also lost.

\begin{table}[]
	\small
	\centering
	\caption{The influence of $\mathcal{L}_\text{kl}$ on evidence allocation and model performances on CIFAR-10. The annealing coefficient $\mu_t=\min(1.0, t/10)$ is adopted.
%		OOD evidence is averaged over six OOD datasets, and results are averaged over 5 runs.
	}
	\vspace{-0.5em}
	\label{tab:evidence}
	\renewcommand{\arraystretch}{1.05}
	\resizebox{\linewidth}{!}{
		\begin{tabular}{@{}c|c|cccc|cc@{}}
			\toprule
			\multirow{2}{*}{Methods} & \multirow{2}{*}{$\mathcal{L}_\text{kl}$} & \multicolumn{4}{c|}{Evidence Per Testing Sample} & \multicolumn{2}{c}{Performance} \\ \cmidrule(l){3-8} 
			&  & \begin{tabular}[c]{@{}c@{}}Target\\ Evidence\end{tabular} & \begin{tabular}[c]{@{}c@{}}Non-target\\ Evidence\end{tabular} & \multicolumn{1}{c|}{\begin{tabular}[c]{@{}c@{}}Target /\\ Total\end{tabular}} & \begin{tabular}[c]{@{}c@{}}OOD\\ Evidence\end{tabular} & Cls Acc & OOD Detect \\ \midrule
			\multirow{2}{*}{EDL} & \checkmark & 282.85 & 1.74 & \multicolumn{1}{c|}{99.39\%} & 28.95 & 88.48 & 82.32 \\
			& \usym{2717} & 1122.65 & 49.13 & \multicolumn{1}{c|}{95.81\%} & 470.99 & 90.20 & 84.50 \\ \midrule
			\multirow{2}{*}{Re-EDL} & \checkmark & 69.80 & 0.61 & \multicolumn{1}{c|}{99.13\%} & 9.52 & 90.09 & 83.73 \\
			& \usym{2717} & 829.51 & 31.56 & \multicolumn{1}{c|}{96.33\%} & 302.34 & 90.13 & 85.46 \\ \bottomrule
	\end{tabular}}
\vspace{-0.5em}
\end{table}

\textbf{\pa{Empirically, how does this regularization influence evidence distributions and uncertainty estimation?}}
As presented in \tabref{evidence}, when $\mathcal{L}_\text{kl}$ is adopted, the total evidence for non-target categories is suppressed to nearly zero.
Since the direct optimization target of the loss function is the probability distribution, the magnitude of the target class evidence also decreases due to the influence of minimal non-target class evidence.
However, the ratio of the target class evidence to total evidence increases, which is positively correlated with the projected probability $\bm P_X$, rising from about 96\% without the use of $\mathcal{L}_\text{kl}$ to over 99\% with it.
Therefore, the increase in this ratio indicates a substantial escalation in the model's overconfidence, which is specifically reflected in the decrease in OOD detection performance shown in \tabref{evidence} and \tabref{kl}.
From an informational perspective, when the evidence for non-target classes vanishes, the potentially useful information carried by the magnitude of evidence is also lost, thereby impairing the model's capability.

\textbf{\pa{Does this regularization always lead to worse results?}}
No, not always.
While in most cases we observe a decrease in OOD detection performance when adopting $\mathcal{L}_\text{kl}$,
%as shown in \tabref{kl}, 
this regularization may bring slight improvements on classification accuracy with a carefully tuned coefficient.
Additionally, our experiments demonstrates that this regularization provides strong generalization ability when encountering noisy testing data.
%We deduce that this benefit arises from $\mathcal{L}_\text{kl}$ constraining the model's output (both target and non-target evidence) within a narrower range, thereby reducing the model's complexity and mitigating the risk of overfitting.
We deduce that this benefit arises from $\mathcal{L}_\text{kl}$ functioning similarly to standard L1/L2 regularization.
While L1/L2 regularization directly limits the magnitude of model parameters to reduce model complexity, $\mathcal{L}_\text{kl}$ indirectly constrains model parameters by suppressing the magnitude of model's outputs (both target and non-target evidence) within a narrower range, thus also reducing complexity and mitigating overfitting.
Furthermore, existing studies~\cite{aguilar2023continual,shi2020multifaceted,pandey2022multidimensional} have explored variants of regularization on non-target evidence, which possess unique properties and have been proven effective in various applications.
Despite that we refer to it as a nonessential setting and, \pa{as shown in \firef{method-graph}}, derive the Re-EDL method by removing it, we believe this type of regularization is effective in certain cases and warrants further exploration.

\vspace{-0.5em}
\subsection{What are the Essential Settings of EDL?}
\label{essential}

In \secref{relax-dirichlet}, we challenge the rigidity of fixing the prior weight to the class number, advocating instead for treating it as an adjustable hyperparameter.
In \secref{simplified-loss}, we propose a simplified empirical risk by deprecating a variance-minimizing optimization term in the traditional EDL loss.
In \secref{kl-div-loss}, we analyze the \pa{disadvantages of}
%discrepancy between the intended purpose and the actual effect of 
the commonly used KL-Div-minimizing regularization on non-target evidence.
%that are not mandated by subjective logic and typically bring minimal benefit, 
\pa{With these relaxations, our EDL variant, Re-EDL, simply optimizes the expectation of the constructed Dirichlet distribution, also known as the projected probability, using the given one-hot labels, without any additional regularization.}
%As illustrated in \firef{method-graph}, by relaxing the above nonessential settings, we derive a much simpler yet more effective EDL variant, Re-EDL.
\pa{The impressive simplicity of the Re-EDL formulation naturally leads to a fundamental question:}
\textbf{What is the essential EDL setting that contributes to its uncertainty estimation capability?}

%Incorporating our relaxations, the optimization of Re-EDL is carried out by aligning the expectation of the constructed Dirichlet distribution, aka the projected probability, with the one-hot label, without the need for additional regularization.
Our experiments shows that, simply replacing the traditional softmax probability with the projected probability in traditional CE and MSE loss functions results in obvious improvements ($>4\%$) in OOD detection performances \pa{(refer to \tabref{loss-form})}.
Therefore, in this subsection, we argue that \textbf{the adoption of the projected probability} is the essential setting of EDL, and delve into its differences with the traditional softmax probability.
For the convenience of following discussion, we first present their formulations here:
\begin{equation}
	\small
	\label{softmax-prob}
	\text{Softmax Probability}_x = \frac{\exp(l_x)}{\sum_{x^\prime\in\mathbb{X}}\exp(l_{x^\prime})},
\end{equation}
%\begin{equation}
%	\text{Projected Prob}_x = \frac{\overbrace{\text{softplus}}^\text{Evidence function}(l_x) + \lambda}{\sum_{x^\prime\in\mathbb{X}}\text{softplus}(l_{x^\prime}) + \underbrace{|\mathbb{X}|\cdot\lambda}_\text{Prior weight}}
%\end{equation}
\begin{equation}
	\small
	\label{projected-prob}
	\begin{aligned}
	&\text{Projected Probability}_x =  \underbrace{\frac{\text{softplus}(l_x)}{\sum_{x^\prime\in\mathbb{X}}\text{softplus}(l_{x^\prime}) + |\mathbb{X}|\cdot\lambda}}_{\text{Belief mass}\: \bm b_X(x)} \\
	&+ \underbrace{\quad\frac{1}{|\mathbb{X}|}\quad}_{\text{Base rate}\:\bm a_X(x)}\cdot\quad \underbrace{\frac{|\mathbb{X}|\cdot\lambda}{\sum_{x^\prime\in\mathbb{X}}\text{softplus}(l_{x^\prime}) + |\mathbb{X}|\cdot\lambda}}_{\text{Uncertainty mass}\:u_X} \\
	&=\frac{\text{softplus}(l_x) + \lambda}{\sum_{x^\prime\in\mathbb{X}}\left(\text{softplus}(l_{x^\prime}) + \lambda\right)},
	\end{aligned}
\end{equation}
where $l_x$ represents the $x$-th value of the output logit vector.

After simplification, the differences between \eqnref{softmax-prob} and \eqnref{projected-prob} lies in:
\textbf{(1)}~Class scores in the projected probability include an additional parameter~$\lambda$;
\textbf{(2)}~The projected probability \pa{replaces the exponential function with softplus for activation}.
As shown in \tabref{essential}, both modifications result in an obvious improvement in OOD detection performance compared to the softmax classification head, without incurring any additional computational cost.
We argue that the underlying reasons for the improvement brought by the two modifications are unified: \textbf{it better utilizes the magnitude information carried by the model's output logits}.

First, the extra parameter $\lambda$ ensures that the information carried by the magnitude of logits is partially preserved during the normalization of the predicted probability distributions.
As discussed in \secref{relax-dirichlet}, in the EDL framework, the value of prior weight determines the extent to which the projected probability is influenced by the magnitude and proportion of evidence, respectively.
When the prior weight $W$ vanishes to zero ($\lambda=W/|\mathbb{X}|=0$), the projected probability is solely determined by the proportion of the evidence distribution, disregarding their magnitudes.
However, this is exactly what consistently occurs in the softmax probability lacking the extra parameter: class scores $[e^5,e^3,e^2]$ are indistinguishable from $[e^0,e^{-2},e^{-3}]$ after normalization, despite the significant difference in their magnitudes.
By introducing an appropriate parameter $\lambda$ to the class scores, as demonstrated in \tabref{essential}, the model obtains the opportunity to leverage the magnitude information carried by logits, which potentially benefits uncertainty estimation.

Secondly, \pa{projected probability replaces the exponential function with an activation function that has a more gradual growth rate, mitigating the severe overconfidence issue in softmax probability,} which heavily favors one class over others.
Due to the nature of exponentiation, the softmax operation tends to amplify larger logits and suppress smaller ones, transforming logit distributions with diverse magnitude information to homogeneous, nearly one-hot probability distributions.
In contrast, softplus offers a gentler growth rate and better preserving magnitude information of logits.

Theoretically, softplus is not the only choice for the evidence function.
Subjective logic only requires the belief mass to be within the range of 0 to 1, meaning the evidence needs to be non-negative.
Hence, in the EDL framework, this function only needs to ensure non-negativity, and options like ReLU, softplus, and even Exp all meet this criterion.
However, Exp's exponential growth rate transforms probability distributions into forms close to one-hot, while ReLU is rarely used for activation in the output layer of modern neural networks because it crudely zeros out all negative logits and causes serious information loss.
For implementation, we deduce that non-negative functions similar to softplus, i.e., those that maintain a gradual but non-zero growth rate over the entire real number domain, are likely to be more effective in preserving magnitude information.

In summary, the essential setting of EDL is replacing \pa{softmax} with the projected probability, which adds a proper extra parameter to class scores and uses \pa{an output activation function with a more gradual growth rate} instead of Exp, thereby better preserving useful magnitude information of logits and improving uncertainty estimation.

\vspace{-0.5em}
\section{Experiments}
\label{experiment}

\subsection{Experimental Setup}

\textbf{Baselines.} Following \cite{deng2023uncertainty}, we focus on comparing with other Dirichlet-based uncertainty methods, including the traditional \textbf{EDL}~\cite{sensoy2018evidential}, \textbf{$\mathcal{I}$-EDL}~\cite{deng2023uncertainty}, \textbf{RED}~\cite{pandey2023learn}, \textbf{R-EDL}~\cite{chen2024r},
%\textbf{KL-PN}~\cite{malinin2018predictive}, \textbf{RKL-PN}~\cite{malinin2019reverse},
\textbf{PostN}~\cite{charpentier2020posterior}, and \textbf{NatPN}~\cite{charpentier2021natural}.
Additionally, we present the results of the representative single-forward-pass method \textbf{DUQ}~\cite{van2020uncertainty}, the popular Bayesian uncertainty method \textbf{MC Dropout}~\cite{gal2016dropout}, and the \textbf{Deep Ensemble}~\cite{lakshminarayanan2017simple} method using 5 and 25 model instances for reference.
For experiments concerning video-modality data, following \cite{bao2021evidential}, we compare our methods with: \textbf{OpenMax}~\cite{bendale2016towards}, \textbf{MC Dropout}, \textbf{BNN SVI}~\cite{krishnan2018bar},  \textbf{RPL}~\cite{chen2020learning}, and \textbf{DEAR}~\cite{bao2021evidential}.

\noindent\textbf{Datasets.} Refer to \appref{datasets} for a detailed introduction.

\begin{table*}[]
	\small
	\centering
	\caption{Performance comparison evaluated by classification accuracy and AUPR scores for OOD detection on CIFAR-10. $\rightarrow$X indicates using X as OOD data. Results are averaged over 5 runs, with mean and standard deviation reported, except for the deep ensemble method which uses 25 model instances.}
	\label{tab:classical}
	\vspace{-0.5em}
	\renewcommand{\arraystretch}{1.1}
	\resizebox{\linewidth}{!}{
		\begin{tabular}{@{}c|cc|ccccccc@{}}
			\toprule
			\multirow{2}{*}{Method} & \multicolumn{2}{c|}{CIFAR10} & $\rightarrow$SVHN & $\rightarrow$CIFAR100 & $\rightarrow$GTSRB & $\rightarrow$LFWPeople & $\rightarrow$Places365 & $\rightarrow$Food101 & Mean \\
			& Cls Acc & Mis Detect & OOD Detect & OOD Detect & OOD Detect & OOD Detect & OOD Detect & OOD Detect & OOD Detect \\ \midrule
			MC Dropout~\cite{gal2016dropout} & \textbf{90.16$\pm$0.23} & {\ul 98.86$\pm$0.06} & 78.40$\pm$3.88 & 85.39$\pm$0.58 & 83.51$\pm$2.32 & 87.98$\pm$2.74 & 67.63$\pm$2.34 & 76.07$\pm$2.58 & 79.83$\pm$2.41 \\
			EDL~\cite{sensoy2018evidential} & 88.48$\pm$0.32 & 98.74$\pm$0.07 & 82.32$\pm$1.21 & 87.13$\pm$0.26 & 84.57$\pm$1.26 & 89.26$\pm$1.70 & 70.46$\pm$0.77 & {\ul 80.18$\pm$0.69} & 82.32$\pm$0.98 \\
			DUQ~\cite{van2020uncertainty} & 89.39$\pm$0.13 & 97.98$\pm$0.34 & 81.44$\pm$4.63 & 85.38$\pm$0.37 & 83.35$\pm$4.30 & 88.96$\pm$4.06 & 66.20$\pm$3.61 & 75.87$\pm$4.40 & 80.20$\pm$3.56 \\
			PostN~\cite{charpentier2020posterior} & 87.82$\pm$0.06 & 97.46$\pm$0.06 & 83.76$\pm$0.46 & 87.07$\pm$0.94 & 84.83$\pm$1.67 & 88.76$\pm$3.08 & 71.79$\pm$2.36 & 78.83$\pm$2.53 & 82.51$\pm$1.84 \\
			NatPN~\cite{charpentier2021natural} & 87.73$\pm$0.09 & 97.53$\pm$0.05 & 83.56$\pm$0.38 & 86.98$\pm$0.75 & 84.76$\pm$2.03 & 88.85$\pm$2.86 & 71.14$\pm$2.10 & 79.04$\pm$2.66 & 82.39$\pm$1.80 \\
			RED~\cite{pandey2023learn} & 89.43$\pm$0.28 & 98.82$\pm$0.09 & 82.85$\pm$2.35 & {\ul 87.84$\pm$0.54} & 85.30$\pm$3.81 & 89.15$\pm$2.90 & 70.78$\pm$2.33 & 79.91$\pm$1.92 & 82.64$\pm$2.31 \\
			$\mathcal{I}$-EDL~\cite{deng2023uncertainty} & 88.38$\pm$0.15 & 98.71$\pm$0.11 & 84.97$\pm$2.11 & 86.31$\pm$0.32 & 84.79$\pm$2.13 & 89.34$\pm$0.98 & 68.92$\pm$1.21 & 77.75$\pm$2.07 & 82.01$\pm$1.47 \\
			\textbf{R-EDL}~\cite{chen2024r} & 90.09$\pm$0.31 & \textbf{98.98$\pm$0.05} & {\ul 85.00$\pm$1.22} & 87.73$\pm$0.31 & {\ul 87.25$\pm$0.69} & \textbf{90.79$\pm$1.15} & {\ul 71.97$\pm$0.69} & 79.64$\pm$2.36 & {\ul 83.73$\pm$1.07} \\
			\textbf{Re-EDL} & {\ul 90.13$\pm$0.25} & 98.81$\pm$0.05 & \textbf{89.94$\pm$1.40} & \textbf{88.31$\pm$0.16} & \textbf{90.53$\pm$2.04} & {\ul 89.71$\pm$2.08} & \textbf{73.42$\pm$1.05} & \textbf{80.83$\pm$1.72} & \textbf{85.46$\pm$1.41} \\ \midrule
			Deep Ensemble(5)~\cite{lakshminarayanan2017simple} & 92.55$\pm$0.14 & 99.32$\pm$0.03 & 84.77$\pm$1.42 & 89.15$\pm$0.15 & 88.01$\pm$1.22 & 90.40$\pm$0.80 & 73.89$\pm$0.62 & 80.60$\pm$1.00 & 84.47$\pm$0.87 \\
			\textbf{Re-EDL+Deep Ensemble(5)} & 92.76$\pm$0.11 & 99.34$\pm$0.02 & {\ul 92.64$\pm$0.35} & {\ul 90.55$\pm$0.13} & {\ul 92.52$\pm$0.91} & {\ul 92.08$\pm$0.96} & {\ul 77.31$\pm$0.27} & {\ul 84.96$\pm$0.42} & {\ul 88.35$\pm$0.51} \\
			Deep Ensemble(25) & {\ul 93.01} & {\ul 99.44} & 86.22 & 90.17 & 89.77 & 90.77 & 75.92 & 82.45 & 85.88 \\
			\textbf{Re-EDL+Deep Ensemble(25)} & \textbf{93.27} & \textbf{99.46} & \textbf{93.45} & \textbf{91.01} & \textbf{93.07} & \textbf{92.29} & \textbf{78.13} & \textbf{85.82} & \textbf{88.96} \\ \bottomrule
	\end{tabular}}
\vspace{-0.5em}
\end{table*}

\subsection{Implementation Details}
\label{implementation-detail}
\textbf{Classical setting.}
Six datasets (\ie, SVHN~\cite{netzer2018street}, CIFAR-100~\cite{krizhevsky2009learning}, GTSRB~\cite{stallkamp2012man}, LFWPeople~\cite{huang2008labeled}, Places365~\cite{zhou2017places}, Food-101~\cite{bossard14}) are utilized as OOD data for CIFAR-10, while FMNIST~\cite{xiao2017fashion} and KMNIST~\cite{clanuwat2018deep} are used for MNIST.
In alignment with \cite{deng2023uncertainty,charpentier2020posterior}, VGG16 serves as the backbone network for CIFAR-10, and a ConvNet with three convolutional and three dense layers is employed for MNIST.
%FMNIST and KMNIST are utilized as OOD data for MNIST, while SVHN and CIFAR-100 are used for CIFAR-10.
Softplus is adopted as the evidence function.
The Adam optimizer is employed with a learning rate of $1\times10^{-4}$ for CIFAR-10, and a learning rate of $1\times10^{-3}$, decaying by $0.1$ every 15 epochs for MNIST.
The hyperparameter $\lambda$ is set to $0.8$ and $0.1$ for CIFAR-10 and MNIST, which is selected from the range [0.1:0.1:1.0] on the validation set.
The batch size is set to 64, and the training epoch is set to 200 and 60 for CIFAR-10 and MNIST.
Reported results are averaged over 5 runs.

\noindent\textbf{Few-shot setting.}
Following \cite{deng2023uncertainty}, we adopt a pre-trained WideResNet-28-10 network from \cite{yang2021free} to extract features and train a single dense layer for experiments under a challenging few-shot setting on the mini-ImageNet~\cite{vinyals2016matching} dataset, with the testing set of CUB~\cite{wah2011caltech} as OOD data.
We employ the $N$-way $K$-shot setting, with $N\in\{5,10\}$ and $K\in\{1,5,20\}$. 
Each few-shot episode comprises $N$ random classes and $K$ random samples per class for training, $\min(15,K)$ query samples per class from mini-ImageNet for classification and confidence estimation, and an equivalent number of query samples from the CUB dataset for OOD detection. Reported results are averaged over 10,000 episodes.
%Note that in the few-shot setting, we perform setting relaxations on $\mathcal{I}$-EDL to achieve stronger performances.
Softplus is used as the activation function to keep evidence non-negative.
The LBFGS optimizer is employed with the default learning rate $1.0$ for 100 epochs.
The hyperparameter $\lambda$ is also selected on the meta-validation set, as shown in \tabref{detail-fewshot} in Appendix.

\noindent\textbf{Video-modality setting.}
Following \cite{bao2021evidential}, we investigate the open-set action recognition task on UCF-101~\cite{soomro2012ucf101} using I3D as the backbone network. HMDB-51~\cite{kuehne2011hmdb} and MiT-v2~\cite{monfort2021multi} serve as sources of unknown samples. The hyperparameter $\lambda$ is set to 0.8, with a batch size of 8.

\noindent\textbf{Noisy setting}. To evaluate the model's generalization on noisy data, we generate noisy samples by adding zero-mean Gaussian noises with standard deviations of [0.025:0.025:0.200] to the testing samples of CIFAR-10.

\noindent\textbf{Other details.}
All experiments were conducted on NVIDIA RTX 3090 GPUs using Python 3.8 and PyTorch 1.12.
The source code link is provided in the abstract.

\begin{table*}[]
	\centering
	\caption{AUPR scores of OOD detection in the classical setting, measured by MP (Max projected probability), UM (Uncertainty Mass), DE (Differential Entropy), and MI (Mutual Information). Please note that this table provides detailed results for specific methods; for a comprehensive overview of all baselines, refer to \tabref{classical}.}
	\vspace{-0.5em}
	\label{tab:classical-ood-aupr-app}
	\renewcommand{\arraystretch}{1.1}
	\small
	\resizebox{\textwidth}{!}{
		\begin{tabular}{@{}c|llll|llll|llll@{}}
			\toprule
			\multirow{2}{*}{Method} & \multicolumn{4}{c|}{CIFAR10$\rightarrow$SVHN} & \multicolumn{4}{c|}{CIFAR10$\rightarrow$CIFAR100} & \multicolumn{4}{c}{CIFAR10$\rightarrow$GTSRB} \\
			& \multicolumn{1}{c}{MP} & \multicolumn{1}{c}{UM} & \multicolumn{1}{c}{DE} & \multicolumn{1}{c|}{MI} & \multicolumn{1}{c}{MP} & \multicolumn{1}{c}{UM} & \multicolumn{1}{c}{DE} & \multicolumn{1}{c|}{MI} & \multicolumn{1}{c}{MP} & \multicolumn{1}{c}{UM} & \multicolumn{1}{c}{DE} & \multicolumn{1}{c}{MI} \\ \midrule
			EDL & \multicolumn{1}{c}{82.30$\pm$1.17} & \multicolumn{1}{c}{82.32$\pm$1.21} & \multicolumn{1}{c}{82.30$\pm$1.18} & \multicolumn{1}{c|}{82.32$\pm$1.21} & \multicolumn{1}{c}{87.15$\pm$0.25} & \multicolumn{1}{c}{87.13$\pm$0.25} & \multicolumn{1}{c}{87.15$\pm$0.26} & \multicolumn{1}{c|}{87.13$\pm$0.26} & \multicolumn{1}{c}{84.59$\pm$1.27} & \multicolumn{1}{c}{84.57$\pm$1.26} & \multicolumn{1}{c}{84.60$\pm$1.27} & \multicolumn{1}{c}{84.58$\pm$1.26} \\
			$\mathcal{I}$-EDL & {\ul 85.11$\pm$2.29} & 84.97$\pm$2.11 & {\ul 85.12$\pm$2.29} & 84.98$\pm$2.12 & 86.36$\pm$0.31 & 86.31$\pm$0.32 & 86.37$\pm$0.31 & 86.31$\pm$0.32 & 84.90$\pm$2.15 & 84.79$\pm$2.14 & 84.91$\pm$2.15 & 84.80$\pm$2.14 \\
			\textbf{R-EDL} & 85.00$\pm$1.22 & {\ul 85.00$\pm$1.22} & 85.01$\pm$1.14 & {\ul 85.00$\pm$1.22} & \textbf{87.72$\pm$0.31} & {\ul 87.73$\pm$0.31} & {\ul 87.61$\pm$0.33} & {\ul 87.73$\pm$0.31} & {\ul 87.25$\pm$0.68} & {\ul 87.25$\pm$0.69} & {\ul 87.16$\pm$0.63} & {\ul 87.25$\pm$0.68} \\
			\textbf{Re-EDL} & \textbf{87.84$\pm$0.96} & \textbf{89.94$\pm$1.40} & \textbf{89.34$\pm$1.31} & \textbf{89.89$\pm$1.39} & {\ul 87.57$\pm$0.23} & \textbf{88.31$\pm$0.16} & \textbf{88.16$\pm$0.17} & \textbf{88.30$\pm$0.16} & \textbf{89.14$\pm$1.69} & \textbf{90.53$\pm$2.04} & \textbf{90.17$\pm$1.96} & \textbf{90.50$\pm$2.03} \\ \midrule
			\multirow{2}{*}{Method} & \multicolumn{4}{c|}{CIFAR10$\rightarrow$LFWPeople} & \multicolumn{4}{c|}{CIFAR10$\rightarrow$Places365} & \multicolumn{4}{c}{CIFAR10$\rightarrow$Food101} \\
			& \multicolumn{1}{c}{MP} & \multicolumn{1}{c}{UM} & \multicolumn{1}{c}{DE} & \multicolumn{1}{c|}{MI} & \multicolumn{1}{c}{MP} & \multicolumn{1}{c}{UM} & \multicolumn{1}{c}{DE} & \multicolumn{1}{c|}{MI} & \multicolumn{1}{c}{MP} & \multicolumn{1}{c}{UM} & \multicolumn{1}{c}{DE} & \multicolumn{1}{c}{MI} \\ \midrule
			EDL & \multicolumn{1}{c}{89.26$\pm$1.67} & \multicolumn{1}{c}{89.26$\pm$1.70} & \multicolumn{1}{c}{89.25$\pm$1.68} & \multicolumn{1}{c|}{89.26$\pm$1.70} & \multicolumn{1}{c}{70.47$\pm$0.78} & \multicolumn{1}{c}{70.46$\pm$0.77} & \multicolumn{1}{c}{70.48$\pm$0.78} & \multicolumn{1}{c|}{70.46$\pm$0.77} & \multicolumn{1}{c}{\textbf{80.19$\pm$0.69}} & \multicolumn{1}{c}{{\ul 80.18$\pm$0.69}} & \multicolumn{1}{c}{{\ul 80.19$\pm$0.69}} & \multicolumn{1}{c}{{\ul 80.18$\pm$0.69}} \\
			$\mathcal{I}$-EDL & 89.33$\pm$0.99 & 89.34$\pm$0.98 & 89.32$\pm$0.99 & 89.34$\pm$0.99 & 68.95$\pm$1.19 & 68.92$\pm$1.21 & 68.97$\pm$1.20 & 68.92$\pm$1.21 & 77.88$\pm$2.10 & 77.75$\pm$2.07 & 77.89$\pm$2.09 & 77.76$\pm$2.07 \\
			\textbf{R-EDL} & \textbf{90.79$\pm$1.15} & \textbf{90.79$\pm$1.15} & \textbf{90.82$\pm$1.14} & \textbf{90.79$\pm$1.15} & {\ul 71.97$\pm$0.69} & {\ul 71.97$\pm$0.69} & {\ul 71.84$\pm$0.70} & {\ul 71.97$\pm$0.69} & 79.64$\pm$2.36 & 79.64$\pm$2.36 & 79.57$\pm$2.38 & 79.64$\pm$2.36 \\
			\textbf{Re-EDL} & {\ul 89.75$\pm$2.08} & {\ul 89.71$\pm$2.08} & {\ul 89.72$\pm$2.09} & {\ul 89.71$\pm$2.08} & \textbf{72.27$\pm$1.03} & \textbf{73.42$\pm$1.05} & \textbf{73.09$\pm$1.04} & \textbf{73.39$\pm$1.05} & {\ul 79.97$\pm$1.61} & \textbf{80.83$\pm$1.72} & \textbf{80.56$\pm$1.67} & \textbf{80.81$\pm$1.71} \\ \bottomrule
	\end{tabular}}
\vspace{-0.5em}
\end{table*}

\subsection{Classical Setting}
\label{classical}
A classifier with reliable uncertainty estimation abilities should assign higher uncertainties to out-of-distribution (OOD) than in-distribution (ID) samples, assign higher uncertainties to misclassified than to correctly classified samples, as well as maintain comparable classification accuracy.
Therefore, we evaluate our method by OOD detection and misclassification detection in image classification, measured by the area under the precision-recall curve (AUPR) with labels 1 for ID data, and labels 0 for OOD data.
For the Dirichlet-base uncertainty methods, we use (the reciprocal of) uncertainty mass as the confidence scores,
while for methods which do not involve Dirichlet PDFs, we use the max probability.
As shown in \tabref{classical}, R-EDL and Re-EDL consistently exhibit superior performance across most metrics.
Specifically, when compared with the traditional EDL method~\cite{sensoy2018evidential} and the newly proposed $\mathcal{I}$-EDL~\cite{deng2023uncertainty}, R-EDL achieves absolute gains of 1.41\% and 1.72\% AUPR averaged over six OOD datasets, by relaxing the nonessential settings on prior weight and deprecating the variance-minimizing optimization term.
Re-EDL offers even greater improvements, with gains of 3.14\% and 3.45\%, by further deprecating the KL-Div-minimizing regularization on non-target evidence.
Moreover, as a single-forward-pass method, Re-EDL can be easily integrated into deep ensemble~\cite{lakshminarayanan2017simple} to enhance performance, albeit at a multiplicative computational cost.
According to \tabref{classical}, when incorporated into the deep ensemble method using 5 and 25 model instances, our proposed Re-EDL achieves enhancements of 3.88\% and 3.08\%, respectively, in the averaged AUPR of OOD detection, alongside modest improvements in classification accuracy and misclassification detection.
All results are averaged over five runs to mitigate effects of randomness.

\begin{figure}
	\centering
	\subfloat[PR curve]{
		\includegraphics[width=0.48\linewidth]{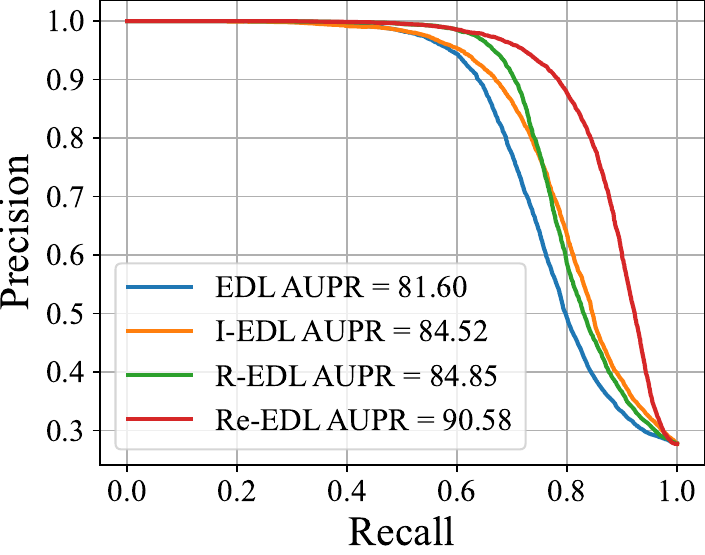}}
	\subfloat[ROC curve]{
		\includegraphics[width=0.48\linewidth]{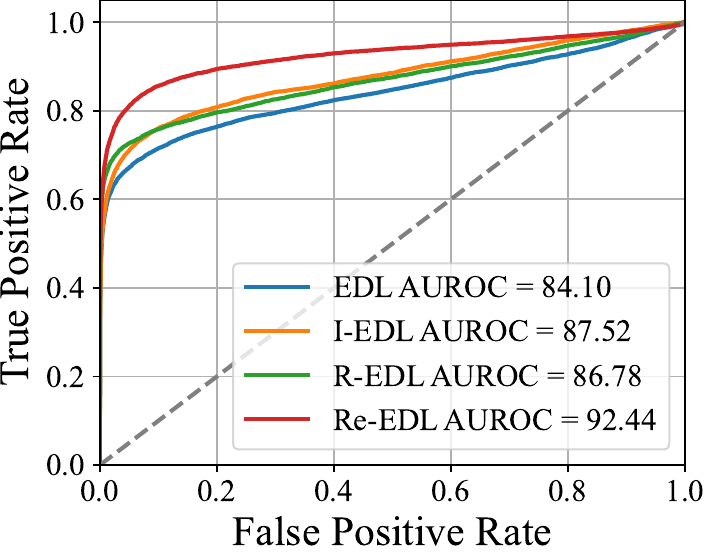}}
	\vspace{-0.5em}
	\caption{Precision-Recall (PR) curves and Receiver Operating Characteristic (ROC) curves of differentiating OOD data (SVHN) from ID data (CIFAR-10) in the classical setting. \pa{The adopted uncertainty measure is uncertainty mass (UM).}}
	\label{aupr-auroc-curve}
	\vspace{-0.5em}
\end{figure}

Moreover, we present AUPR scores of OOD detection using various uncertainty measures, including MP (Max projected Probability), UM (Uncertainty Mass), DE (Differential Entropy), and MI (Mutual Information), in \tabref{classical-ood-aupr-app}, and AUROC scores in \tabref{classical-ood-auroc-app} (\appref{additional-classical}). 
%which further validate the effectiveness of Re-EDL.
Derivations of these measures can be found in \appref{measures}.
\pa{In \firef{aupr-auroc-curve}, we present the Precision-Recall (PR) curves and Receiver Operating Characteristic (ROC) curves of differentiating OOD data (SVHN) from ID data (CIFAR-10) using UM as the uncertainty measure.
Curves with other measures are plotted in \firef{aupr-curve} and \firef{auroc-curve} (\appref{additional-classical}).}
\tabref{mnist} shows results on MNIST, with FMNIST and KMNIST as OOD data.
These results further validate the effectiveness of Re-EDL.

\begin{table}[]
	\small
	\centering
	\caption{Results on MNIST, averaged over 5 runs. ConvNet consisting of 3 conventional layers and 3 dense layers is adopted as the backbone network.}
	\vspace{-0.5em}
	\label{tab:mnist}
	\renewcommand{\arraystretch}{1.1}
	\resizebox{0.9\linewidth}{!}{
	\begin{tabular}{@{}c|cc|cc@{}}
		\toprule
		\multirow{2}{*}{Method} & \multicolumn{2}{c|}{MNIST} & $\rightarrow$KMNIST & $\rightarrow$FMNIST \\
		& Cls Acc & Mis Detect & OOD Detect & OOD Detect \\ \midrule
		MC Dropout & 99.30$\pm$0.05 & 99.98$\pm$0.02 & 98.09$\pm$0.12 & 98.73$\pm$0.41 \\
		DUQ & 98.65$\pm$0.12 & 99.97$\pm$0.03 & 98.52$\pm$0.11 & 97.92$\pm$0.60 \\
		PostN & 99.29$\pm$0.04 & 99.97$\pm$0.03 & 94.62$\pm$0.24 & 97.28$\pm$0.37 \\
		EDL & 98.22$\pm$0.31 & 99.98$\pm$0.00 & 96.31$\pm$2.03 & 98.08$\pm$0.42 \\
		$\mathcal{I}$-EDL & 99.21$\pm$0.08 & 99.98$\pm$0.00 & 98.33$\pm$0.24 & 98.86$\pm$0.29 \\
		\textbf{R-EDL} & {\ul 99.33$\pm$0.03} & {\ul 99.99$\pm$0.00} & {\ul 98.69$\pm$0.20} & {\ul 99.29$\pm$0.12} \\
		\textbf{Re-EDL} & \textbf{99.35$\pm$0.00} & \textbf{99.99$\pm$0.00} & \textbf{99.03$\pm$0.28} & \textbf{99.65$\pm$0.09} \\ \bottomrule
\end{tabular}}
\end{table}

\begin{table*}[]
	\centering
	\caption{Results of the few-shot setting for WideResNet-28-10 on mini-ImageNet, with the CUB dataset as OOD data.
%		The reported results are averaged over 10000 episodes.
	}
	\vspace{-0.5em}
	\label{tab:miniimagenet-cub-acc&conf&ood-aupr-main}
	\renewcommand{\arraystretch}{1.05}
	\small
	\resizebox{\textwidth}{!}{
		\begin{tabular}{@{}c|cccc|cccc|cccc@{}}
			\toprule
			\multirow{2}{*}{Method} & \multicolumn{4}{c|}{5-Way 1-Shot} & \multicolumn{4}{c|}{5-Way 5-Shot} & \multicolumn{4}{c}{5-Way 20-Shot} \\ \cmidrule(l){2-13} 
			& Cls Acc & Mis Detect & OOD-MP & OOD-UM & Cls Acc & Mis Detect & OOD-MP & OOD-UM & Cls Acc & Mis Detect & OOD-MP & OOD-UM \\ \midrule
			EDL & 61.00$\pm$0.11 & 75.34$\pm$0.12 & 66.78$\pm$0.12 & 65.41$\pm$0.13 & 80.38$\pm$0.08 & 92.09$\pm$0.05 & 74.46$\pm$0.10 & 76.53$\pm$0.14 & 85.54$\pm$0.06 & {\ul 97.05$\pm$0.02} & 80.01$\pm$0.10 & 79.78$\pm$0.12 \\
			$\mathcal{I}$-EDL & {\ul 63.82$\pm$0.10} & 80.33$\pm$0.11 & 71.79$\pm$0.12 & 74.76$\pm$0.13 & {\ul 82.00$\pm$0.07} & 93.61$\pm$0.05 & {\ul 82.04$\pm$0.10} & 82.48$\pm$0.10 & 88.12$\pm$0.05 & 96.98$\pm$0.02 & 84.29$\pm$0.09 & 85.40$\pm$0.09 \\
			\textbf{R-EDL} & \textbf{63.93$\pm$0.11} & \textbf{80.80$\pm$0.11} & {\ul 72.91$\pm$0.12} & {\ul 74.84$\pm$0.13} & 81.85$\pm$0.07 & {\ul 93.65$\pm$0.05} & \textbf{83.65$\pm$0.10} & \textbf{84.22$\pm$0.10} & \textbf{88.74$\pm$0.05} & \textbf{97.10$\pm$0.02} & {\ul 84.85$\pm$0.09} & {\ul 85.57$\pm$0.09} \\
			\textbf{Re-EDL} & 63.64$\pm$0.10 & {\ul 80.65$\pm$0.11} & \textbf{73.20$\pm$0.12} & \textbf{75.61$\pm$0.13} & \textbf{82.05$\pm$0.07} & \textbf{93.75$\pm$0.05} & 83.10$\pm$0.10 & {\ul 83.35$\pm$0.10} & {\ul 88.32$\pm$0.05} & 96.39$\pm$0.03 & \textbf{85.15$\pm$0.09} & \textbf{86.06$\pm$0.09} \\ \midrule
			\multirow{2}{*}{Method} & \multicolumn{4}{c|}{10-Way 1-Shot} & \multicolumn{4}{c|}{10-Way 5-Shot} & \multicolumn{4}{c}{10-Way 20-Shot} \\ \cmidrule(l){2-13} 
			& Cls Acc & Mis Detect & OOD-MP & OOD-UM & Cls Acc & Mis Detect & OOD-MP & OOD-UM & Cls Acc & Mis Detect & OOD-MP & OOD-UM \\ \midrule
			EDL & 44.55$\pm$0.08 & 61.68$\pm$0.10 & 59.19$\pm$0.09 & 67.81$\pm$0.12 & 62.50$\pm$0.08 & 84.35$\pm$0.06 & 71.06$\pm$0.10 & 76.28$\pm$0.10 & 69.30$\pm$0.09 & 93.15$\pm$0.03 & 74.50$\pm$0.08 & 76.89$\pm$0.09 \\
			$\mathcal{I}$-EDL & 49.37$\pm$0.07 & {\ul 67.54$\pm$0.09} & {\ul 71.60$\pm$0.10} & {\ul 71.95$\pm$0.10} & 67.89$\pm$0.06 & 85.52$\pm$0.05 & 80.63$\pm$0.11 & 82.29$\pm$0.10 & 78.59$\pm$0.04 & {\ul 93.32$\pm$0.03} & 81.34$\pm$0.07 & 82.52$\pm$0.07 \\
			\textbf{R-EDL} & {\ul 50.02$\pm$0.07} & 67.12$\pm$0.09 & \textbf{72.83$\pm$0.10} & \textbf{73.08$\pm$0.10} & \textbf{70.51$\pm$0.05} & \textbf{86.26$\pm$0.05} & {\ul 82.39$\pm$0.09} & \textbf{83.37$\pm$0.09} & \textbf{79.79$\pm$0.04} & \textbf{93.47$\pm$0.02} & {\ul 82.22$\pm$0.08} & {\ul 82.72$\pm$0.07} \\
			\textbf{Re-EDL} & \textbf{50.69$\pm$0.07} & \textbf{68.47$\pm$0.09} & 71.48$\pm$0.10 & 71.39$\pm$0.10 & {\ul 69.38$\pm$0.05} & {\ul 85.79$\pm$0.05} & \textbf{82.46$\pm$0.09} & {\ul 83.33$\pm$0.08} & {\ul 78.79$\pm$0.04} & 91.20$\pm$0.04 & \textbf{82.79$\pm$0.07} & \textbf{84.09$\pm$0.07} \\ \bottomrule
	\end{tabular}}
	\vspace{-0.5em}
\end{table*}

\subsection{Few-shot Setting}
\label{fewshot}

Next, we conduct more challenging few-shot experiments on mini-ImageNet to further demonstrate the effectiveness of our method.
As shown in \tabref{miniimagenet-cub-acc&conf&ood-aupr-main}, we report the averaged top-1  accuracy of classification and the AUPR scores of confidence estimation and OOD detection over 10,000 few-shot episodes.
As depicted in \tabref{miniimagenet-cub-acc&conf&ood-aupr-main}, R-EDL and Re-EDL achieve satisfactory performances on most $N$-way $K$-shot settings.
Specifically, comparing with the EDL and $\mathcal{I}$-EDL methods, R-EDL obtains absolute gains of 9.19\% and 1.61\% when evaluated by MP on OOD detection of the $5$-way $5$-shot task.
However, in this setting, Re-EDL fail to achieve further improvements over R-EDL as it does in the classical setting and subsequent video-modality setting.
\pa{We speculate that this may be because few-shot tasks demand stronger generalization capabilities due to the limited amount of training data.
Consequently, the improvement in generalization brought about by the KL-Divergence-minimizing regularization compensates for the performance loss it otherwise causes.}
However, after deprecating this regularization, our simpler Re-EDL still achieves performances comparable to those of R-EDL in the few-shot setting, which also suggests that the regularization is not an essential EDL setting.

\begin{table*}[]
	\centering
	\caption{Results of the noisy setting on CIFAR-10, averaged over 5 seeds. Noisy samples are generated by adding zero-mean Gaussian noises with standard deviations of [0.025:0.025:0.200] to the testing samples of CIFAR-10.}
	\label{tab:noise-complete}
	\vspace{-0.5em}
	\renewcommand{\arraystretch}{1.05}
	\small
	\resizebox{0.95\textwidth}{!}{
		\begin{tabular}{@{}c|ccc|ccc|ccc|ccc@{}}
			\toprule
			SD of Noise & \multicolumn{3}{c|}{0.025} & \multicolumn{3}{c|}{0.050} & \multicolumn{3}{c|}{0.075} & \multicolumn{3}{c}{0.100} \\ \midrule
			Method & Cls Acc & Noisy Detect & Avg & Cls Acc & Noisy Detect & Avg & Cls Acc & Noisy Detect & Avg & Cls Acc & Noisy Detect & Avg \\ \midrule
			EDL & 86.26$\pm$0.44 & 52.78$\pm$0.23 & 69.52 & 75.38$\pm$1.11 & {\ul 62.75$\pm$0.65} & 69.07 & 58.13$\pm$3.04 & {\ul 72.98$\pm$1.44} & 65.56 & 41.87$\pm$3.98 & 79.10$\pm$2.24 & 60.48 \\
			$\mathcal{I}$-EDL & 85.80$\pm$0.56 & 52.91$\pm$0.47 & 69.36 & 73.40$\pm$2.25 & 62.12$\pm$1.38 & 67.76 & 56.78$\pm$3.77 & 70.32$\pm$1.35 & 63.55 & 43.62$\pm$4.42 & 75.58$\pm$0.58 & 59.60 \\
			\textbf{R-EDL} & \textbf{87.54$\pm$0.35} & \textbf{53.48$\pm$0.55} & \textbf{70.51} & \textbf{76.35$\pm$1.56} & \textbf{64.19$\pm$1.02} & \textbf{70.27} & \textbf{60.19$\pm$3.03} & \textbf{75.33$\pm$1.43} & \textbf{67.76} & \textbf{45.70$\pm$3.67} & \textbf{82.97$\pm$1.54} & \textbf{64.34} \\
			\textbf{Re-EDL} & {\ul 87.37$\pm$0.64} & {\ul 53.18$\pm$0.18} & {\ul 70.28} & {\ul 75.76$\pm$1.72} & 62.69$\pm$0.71 & {\ul 69.22} & {\ul 59.68$\pm$3.04} & 72.94$\pm$1.34 & {\ul 66.31} & {\ul 44.22$\pm$2.60} & {\ul 80.24$\pm$1.89} & {\ul 62.23} \\ \midrule
			SD of Noise & \multicolumn{3}{c|}{0.125} & \multicolumn{3}{c|}{0.150} & \multicolumn{3}{c|}{0.175} & \multicolumn{3}{c}{0.200} \\ \midrule
			Method & Cls Acc & Noisy Detect & Avg & Cls Acc & Noisy Detect & Avg & Cls Acc & Noisy Detect & Avg & Cls Acc & Noisy Detect & Avg \\ \midrule
			EDL & 30.45$\pm$4.27 & 82.07$\pm$2.85 & 56.26 & 23.45$\pm$4.00 & 83.77$\pm$3.20 & 53.61 & 19.51$\pm$3.77 & 85.13$\pm$3.34 & 52.32 & 17.15$\pm$3.35 & 86.45$\pm$3.31 & 51.80 \\
			$\mathcal{I}$-EDL & {\ul 34.58$\pm$4.52} & 79.31$\pm$1.15 & {\ul 56.94} & \textbf{28.56$\pm$4.35} & 82.13$\pm$2.23 & 55.34 & \textbf{24.45$\pm$4.14} & 84.29$\pm$3.12 & 54.37 & \textbf{21.57$\pm$3.69} & 85.97$\pm$3.69 & 53.77 \\
			\textbf{R-EDL} & \textbf{35.16$\pm$2.97} & \textbf{87.62$\pm$1.49} & \textbf{61.39} & {\ul 28.47$\pm$1.87} & \textbf{90.50$\pm$1.55} & \textbf{59.48} & {\ul 24.07$\pm$2.08} & \textbf{92.32$\pm$1.86} & \textbf{58.20} & {\ul 20.83$\pm$2.50} & \textbf{93.50$\pm$2.24} & \textbf{57.16} \\
			\textbf{Re-EDL} & 33.85$\pm$1.69 & {\ul 84.61$\pm$2.29} & 59.23 & 27.00$\pm$1.61 & {\ul 87.16$\pm$2.65} & {\ul 57.08} & 22.83$\pm$2.59 & {\ul 88.72$\pm$3.10} & {\ul 55.78} & 19.90$\pm$3.34 & {\ul 89.76$\pm$3.68} & {\ul 54.83} \\ \bottomrule
	\end{tabular}}
	\vspace{-0.5em}
\end{table*}

\begin{table}[]
	\centering
	\caption{Results of video-modality setting for I3D backbone on UCF-101, with HMDB-51 and MiT-v2 as OOD data. Results of baselines are reported by \cite{bao2021evidential}.}
	\label{tab:osar}
	\vspace{-0.5em}
	\renewcommand{\arraystretch}{1.05}
	\small
	\resizebox{\linewidth}{!}{
		\begin{tabular}{@{}c|cc|cc@{}}
			\toprule
			\multirow{2}{*}{Method}          & \multicolumn{2}{c|}{UCF-101$\rightarrow$HMDB-51} & \multicolumn{2}{c}{UCF-101$\rightarrow$MiT-v2} \\
			& Open maF1           & Open Set AUC               & Open maF1         & Open Set AUC                                  \\ \midrule
			OpenMax               & 67.85$\pm$0.12             & 74.34             & 66.22$\pm$0.16             & 77.76            \\
			MC Dropout            & 71.13$\pm$0.15             & 75.07             & 68.11$\pm$0.20             & 79.14            \\
			BNN SVI               & 71.57$\pm$0.17             & 74.66             & 68.65$\pm$0.21             & 79.50            \\
			SoftMax               & 73.19$\pm$0.17             & 75.68             & 68.84$\pm$0.23             & 79.94            \\
			RPL                   & 71.48$\pm$0.15             & 75.20             & 68.11$\pm$0.20             & 79.16            \\
			DEAR                  & 77.24$\pm$0.18             & 77.08             & 69.98$\pm$0.23             & 81.54            \\ \midrule
			\textbf{R-EDL}          & {\ul 78.73$\pm$0.15}    & {\ul 77.94}    & {\ul 70.85$\pm$0.25}    & {\ul 82.26}   \\ 
			\textbf{Re-EDL}          & \textbf{78.92$\pm$0.13}    & \textbf{78.02}    & \textbf{71.03$\pm$0.21}    & \textbf{82.39}   \\ 
			\bottomrule
		\end{tabular}
	}
	\vspace{-0.5em}
\end{table}

\subsection{Video-modality Setting}
\label{osar}
We also assess our approach using video-modality samples~\cite{bao2021evidential,gao2020learning}, specifically on the open-set action recognition task. 
Following \cite{bao2021evidential}, we train models on UCF-101 training split and use the testing splits of HMDB-51 and MiT-v2 datasets as unknown sources.
Given that the SOTA method DEAR is predicated on EDL, we substitute its EDL implementation with our R-EDL and Re-EDL version.
As evidenced by \tabref{osar}, this modification yields enhanced performance, substantiating the efficacy of our methods.

\subsection{Noisy Setting}
\label{noisy}
In \secref{kl-div-loss}, we argue that the regularization $\mathcal{L}_\text{kl}$ constrains the magnitude of model outputs within a narrower range, thereby reducing complexity and mitigating overfitting to some extent.
Although $\mathcal{L}_\text{kl}$ \pa{typically} has a negative effect in experiments across the \pa{classical and video-modality} settings, it exhibits strong generalization capabilities when tested on noisy data.
Specifically, we introduce zero-mean isotropic Gaussian noise into the test split of the ID dataset to generate noisy samples.
\tabref{noise-complete} presents the classification accuracy and the AUPR scores for noisy detection across varying levels of Gaussian noise on CIFAR-10.
As indicated in \tabref{noise-complete}, R-EDL significantly outperforms both EDL and $\mathcal{I}$-EDL across nearly all noise levels, with its advantages becoming more pronounced as the noise intensity increases.
However, the performance of Re-EDL, which omits the KL-Div-minimizing regularization, generally falls below that of R-EDL.
This observation suggests that $\mathcal{L}_\text{kl}$ equips R-EDL with superior generalization abilities, enabling it to collect reliable evidence from noisy data, thereby leading to enhanced performances.
Nonetheless, Re-EDL still maintains superior performances compared to both EDL and $\mathcal{I}$-EDL.

%\begin{figure}[]
%	\centering
%	\includegraphics[width=\linewidth]{graph/noise.png}
%	\caption{The performance trends of EDL, $\mathcal{I}$-EDL, R-EDL, and Re-EDL, measured by the average of the classification accuracy and the AUPR score of OOD detection, across varying levels of Gaussian noise.}
%	\label{noise}
%\end{figure}

\vspace{-0.5em}
\subsection{Parameter Analysis}

\label{parameter}

\begin{figure}[]
	\centering
	\includegraphics[width=0.8\linewidth]{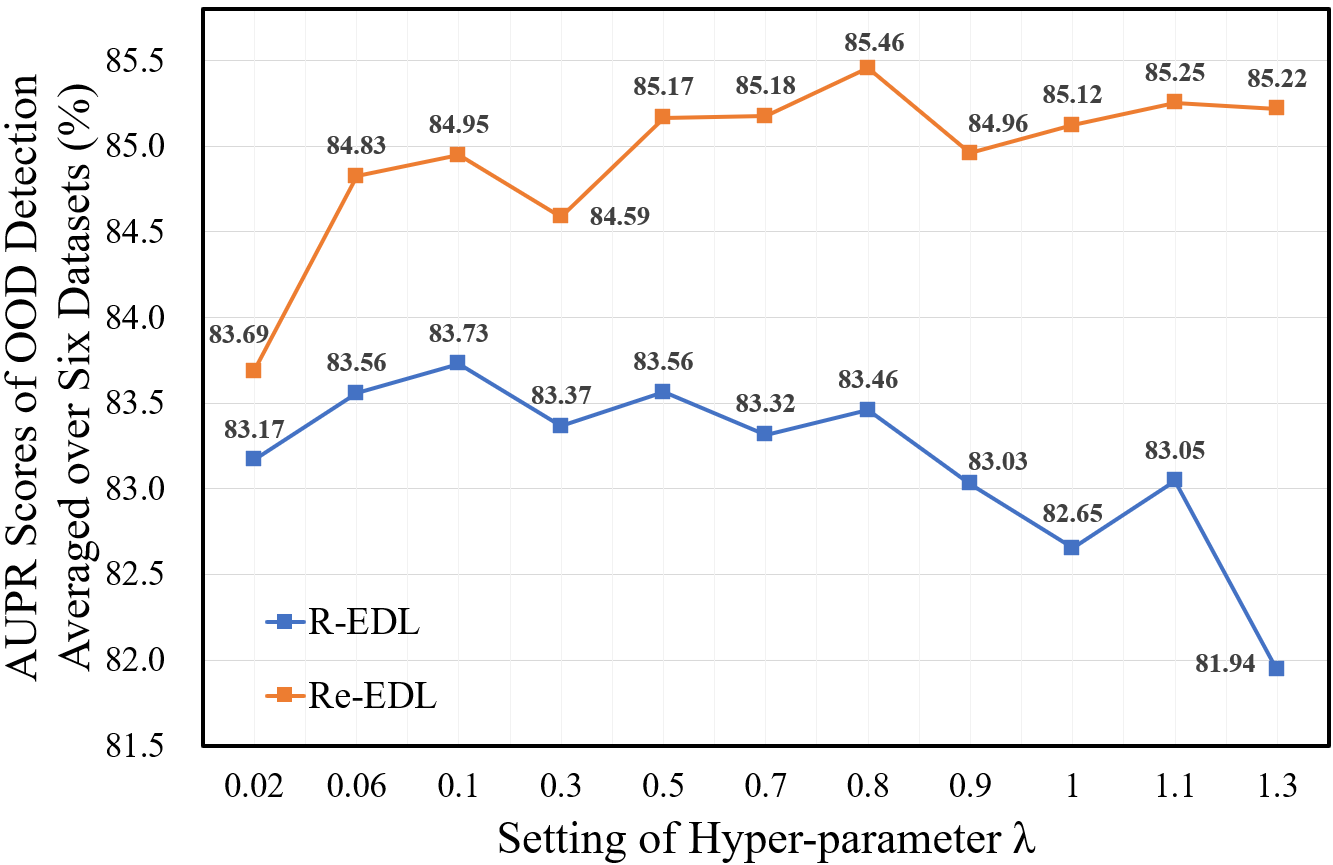}
	\vspace{-0.5em}
	\caption{Parameter analysis of the hyperparameter $\lambda$, evaluated by averaged AUPR for OOD detection on CIFAR-10.
		%		: SVHN, CIFAR-100, GTSRB, LFWPeople, and Places365.
	}
	\label{param}
	\vspace{-0.5em}
\end{figure}

\begin{table}[]
	\centering
	\caption{Performances on CIFAR-10 with varying values of the KL-Div-minimizing regularization coefficient $\mu$.
		Note that $\mu$ here is a scalar coefficient instead of an annealing one.
	}
	\vspace{-0.5em}
	\label{tab:kl}
	\renewcommand{\arraystretch}{1.05}
	\resizebox{\linewidth}{!}{
		\begin{tabular}{@{}ccc|ccc@{}}
			\toprule
			\multicolumn{3}{c|}{EDL} & \multicolumn{3}{c}{Re-EDL} \\ \midrule
			\multicolumn{1}{c|}{$\lg \mu$} & Cls Acc & OOD Detect & \multicolumn{1}{c|}{$\lg \mu$} & Cls Acc & OOD Detect \\ \midrule
			\multicolumn{1}{c|}{$-\infty$} & 90.20 & 84.50 & \multicolumn{1}{c|}{$-\infty$} & 90.13 & 85.46 \\
			\multicolumn{1}{c|}{-3.0} & 90.12 (\red{-0.08}) & 84.79 (\green{+0.29}) & \multicolumn{1}{c|}{-3.0} & 90.06 (\red{-0.07}) & 84.41 (\red{-1.05}) \\
			\multicolumn{1}{c|}{-2.3} & 90.21 (\green{+0.01}) & 83.97 (\red{-0.53}) & \multicolumn{1}{c|}{-2.3} & 90.19 (\green{+0.06}) & 83.81 (\red{-1.65}) \\
			\multicolumn{1}{c|}{-2.0} & 90.22 (\green{+0.02}) & 82.80 (\red{-1.70}) & \multicolumn{1}{c|}{-2.0} & 89.86 (\red{-0.27}) & 83.06 (\red{-2.40}) \\
			\multicolumn{1}{c|}{-1.3} & 90.15 (\red{-0.06}) & 82.80 (\red{-1.70}) & \multicolumn{1}{c|}{-1.3} & 90.40 (\green{+0.27}) & 83.20 (\red{-2.26}) \\
			\multicolumn{1}{c|}{-1.0} & 90.22 (\green{+0.02}) & 82.88 (\red{-1.62}) & \multicolumn{1}{c|}{-1.0} & 90.34 (\green{+0.21}) & 83.16 (\red{-2.30}) \\
			\multicolumn{1}{c|}{-0.3} & 89.47 (\red{-0.73}) & 82.86 (\red{-1.64}) & \multicolumn{1}{c|}{-0.3} & 24.29 (\red{-65.84}) & 51.43 (\red{-34.03}) \\
			\multicolumn{1}{c|}{0.0} & 74.99 (\red{-15.21}) & 76.13 (\red{-8.37}) & \multicolumn{1}{c|}{0.0} & 13.13 (\red{-77.00}) & 43.86 (\red{-41.60}) \\
			\multicolumn{1}{c|}{0.7} & 32.52 (\red{-57.68}) & 46.54 (\red{-37.96}) & \multicolumn{1}{c|}{0.7} & 12.05 (\red{-78.08}) & 40.33 (\red{-45.13}) \\ \bottomrule
	\end{tabular}}
\vspace{-0.5em}
\end{table}

\textbf{Hyperparameter $\lambda$ in projected probability.} We further investigate the effect of the hyperparameter $\lambda$.
\firef{param} presents the trend of variation in the average AUPR score for OOD detection on six OOD datasets as the hyperparameter $\lambda$ varies from 0.02 to 1.3.
The observations reveal findings in two aspects.
On one hand, examining the two curves individually, the hyperparameter $\lambda$ has a noticeable impact on OOD detection performance, with the effect varying by nearly 2\% within the range of 0.02 to 1.3. Fixing the prior weight to the number of classes, i.e., setting $\lambda$ to 1, typically fails to achieve optimal performance.
On the other hand, when comparing the performance gap between the two curves, removing $\mathcal{L}_\text{kl}$ consistently improves the performance of R-EDL, with the gap tending to widen as the value of $\lambda$ increases.
We speculate that as $\lambda$ increases, the prediction becomes more dependent on the magnitude of the evidence rather than its proportion.
Since $\mathcal{L}_\text{kl}$ compresses the magnitude of the evidence, potentially reducing the information it carries, the negative impact of $\mathcal{L}_\text{kl}$ on performance becomes more pronounced as $\lambda$ grows.

\noindent\pa{\textbf{Coefficient $\mu$ of regularization $\mathcal{L}_\text{kl}$.} Acute readers may wonder how the coefficient of the KL-Div-minimizing regularization affects performances.
In \tabref{kl} we present the results on CIFAR-10 with varying values of the coefficient $\mu$ in EDL and Re-EDL, which clearly demonstrate that while this regularization can slightly improve classification accuracy with a carefully tuned coefficient, it generally leads to worse uncertainty estimation as the coefficient $\mu$ increases.}

\subsection{Ablation Study}
\label{ablation}

\begin{table*}[]
	\centering
	\caption{Ablation study on the classical setting, with respect to the relaxations about treating $\lambda$ as a hyperparameter and deprecating the optimization term $\mathcal{L}_\text{var}$ and the regularization term $\mathcal{L}_\text{kl}$.}
	\label{tab:ablation}
	\vspace{-0.5em}
	\renewcommand{\arraystretch}{1.05}
	\small
	\resizebox{\textwidth}{!}{
		\begin{tabular}{@{}ccc|cc|ccccccc@{}}
			\toprule
			\multicolumn{3}{c|}{Nonessential Settings} & \multicolumn{2}{c|}{CIFAR10} & $\rightarrow$SVHN & $\rightarrow$CIFAR100 & $\rightarrow$GTSRB & $\rightarrow$LFWPeople & $\rightarrow$Places365 & $\rightarrow$Food101 & Mean \\
			$\lambda=1$ & $\mathcal{L}_\text{var}$ & $\mathcal{L}_\text{kl}$ & Cls Acc & Mis Detect & OOD Detect & OOD Detect & OOD Detect & OOD Detect & OOD Detect & OOD Detect & OOD Detect \\ \midrule
			\checkmark & \checkmark & \checkmark & 88.48$\pm$0.32 & 98.74$\pm$0.07 & 82.32$\pm$1.21 & 87.13$\pm$0.26 & 84.57$\pm$1.26 & 89.26$\pm$1.70 & 70.46$\pm$0.77 & 80.18$\pm$0.69 & 82.32$\pm$0.98 \\ \midrule
			- & \checkmark & \checkmark & 89.33$\pm$0.18 & {\ul 98.94$\pm$0.03} & 84.77$\pm$1.82 & 87.62$\pm$0.15 & 87.68$\pm$1.09 & {\ul 90.31$\pm$1.86} & 71.68$\pm$1.27 & {\ul 80.41$\pm$0.90} & 83.75$\pm$1.18 \\
			\checkmark & - & \checkmark & 88.25$\pm$0.29 & 98.73$\pm$0.04 & 84.30$\pm$1.56 & 86.90$\pm$0.33 & 84.44$\pm$0.33 & 88.40$\pm$2.18 & 69.79$\pm$1.56 & 80.33$\pm$0.98 & 82.36$\pm$1.43 \\
			\checkmark & \checkmark & - & {\ul 90.20$\pm$0.09} & 98.78$\pm$0.02 & 88.05$\pm$1.95 & 88.17$\pm$0.22 & 88.87$\pm$1.36 & 88.41$\pm$2.84 & \textbf{75.20$\pm$0.77} & 78.33$\pm$1.96 & 84.50$\pm$1.52 \\ \midrule
			- & - & \checkmark & 90.09$\pm$0.31 & \textbf{98.98$\pm$0.05} & 85.00$\pm$1.22 & 87.73$\pm$0.31 & 87.25$\pm$0.69 & \textbf{90.79$\pm$1.15} & 71.97$\pm$0.69 & 79.64$\pm$2.36 & 83.73$\pm$1.07 \\
			- & \checkmark & - & \textbf{90.25$\pm$0.23} & 98.78$\pm$0.07 & 87.30$\pm$1.38 & \textbf{88.45$\pm$0.39} & 87.99$\pm$3.05 & 89.07$\pm$1.84 & {\ul 74.86$\pm$0.85} & 79.22$\pm$3.13 & 84.48$\pm$1.77 \\
			\checkmark & - & - & 90.19$\pm$0.14 & 98.82$\pm$0.06 & {\ul 88.93$\pm$1.11} & 88.29$\pm$0.11 & {\ul 89.46$\pm$1.20} & 90.24$\pm$0.85 & 73.45$\pm$0.72 & 80.33$\pm$0.92 & {\ul 85.12$\pm$0.82} \\ \midrule
			- & - & - & 90.13$\pm$0.25 & 98.81$\pm$0.05 & \textbf{89.94$\pm$1.40} & {\ul 88.31$\pm$0.16} & \textbf{90.53$\pm$2.04} & 89.71$\pm$2.08 & 73.42$\pm$1.05 & \textbf{80.83$\pm$1.72} & \textbf{85.46$\pm$1.41} \\ \bottomrule
		\end{tabular}
	}
\vspace{-0.5em}
\end{table*}

\begin{table}[]
	\centering
	\caption{Ablation study of the essential EDL setting. Experimental settings keep consistent with Table 1.}
	\label{tab:essential}
	\vspace{-0.5em}
	\renewcommand{\arraystretch}{1.05}
	\resizebox{\linewidth}{!}{
		\begin{tabular}{@{}c|cc|cc|cc@{}}
			\toprule
			\multirow{2}{*}{Methods} & \multicolumn{2}{c|}{Class Score} & \multicolumn{2}{c|}{Activation Function} & \multicolumn{2}{c}{Performance} \\ \cmidrule(l){2-7} 
			& Vanilla & Extra $\lambda$ & Exponential & Softplus & Cls Acc & OOD Detect \\ \midrule
			Softmax & \checkmark & - & \checkmark & - & {\ul 90.10$\pm$0.25} & 81.30$\pm$2.36 \\
			Re-EDL ($\lambda=0$) & \checkmark & - & - & \checkmark & 90.08$\pm$0.10 & {\ul 83.48$\pm$2.11} \\ \midrule
			Re-EDL (Exp) & - & \checkmark & \checkmark & - & 89.81$\pm$0.18 & 82.39$\pm$2.03 \\
			Re-EDL & - & \checkmark & - & \checkmark & \textbf{90.13$\pm$0.25} & \textbf{85.46$\pm$1.41} \\ \bottomrule
	\end{tabular}}
\end{table}

\begin{table}[]
	\centering
	\caption{Validation of the essential EDL setting in the cross-entropy (CE) loss formulation by simply replacing softmax classification head with projected probability.}
	\label{tab:loss-form}
	\renewcommand{\arraystretch}{1.05}
	\small
	\vspace{-0.5em}
	\resizebox{0.48\textwidth}{!}{
		\begin{tabular}{@{}c|c|c|ccc@{}}
			\toprule
			\multirow{2}{*}{Loss} & \multirow{2}{*}{Method} & CIFAR10 & $\rightarrow$SVHN & $\rightarrow$CIFAR100 & Mean (six datasets) \\
			&  & Cls Acc & OOD Detect & OOD Detect & OOD Detect \\ \midrule
			\multirow{2}{*}{CE} & Softmax & 90.11$\pm$0.19 & 80.05$\pm$3.04 & 85.86$\pm$0.67 & 80.33$\pm$2.70 \\
			& Re-EDL & \textbf{90.48$\pm$0.25} & {\ul 85.54$\pm$1.29} & \textbf{88.65$\pm$0.37} & {\ul 84.56$\pm$1.00} \\ \midrule
			\multirow{2}{*}{MSE} & Softmax & 90.10$\pm$0.25 & 83.62$\pm$3.29 & 86.54$\pm$0.18 & 81.30$\pm$2.36 \\
			& Re-EDL & {\ul 90.13$\pm$0.25} & \textbf{89.94$\pm$1.40} & {\ul 88.31$\pm$0.16} & \textbf{85.46$\pm$1.41} \\ \bottomrule
	\end{tabular}}
\vspace{-0.5em}
\end{table}

\begin{table}[]
	\centering
	\caption{Comparison of common evidence functions.}
	\label{tab:activation}
	\renewcommand{\arraystretch}{1.05}
	\small
	\vspace{-0.5em}
	\resizebox{0.48\textwidth}{!}{
		\begin{tabular}{@{}c|c|ccc@{}}
			\toprule
			\multirow{2}{*}{\begin{tabular}[c]{@{}c@{}}Evidence\\ Functions\end{tabular}} & CIFAR10 & $\rightarrow$SVHN & $\rightarrow$CIFAR100 & Mean (six datasets) \\
			& Cls Acc & OOD Detect & OOD Detect & OOD Detect \\ \midrule
			ReLU & 83.56$\pm$6.10 & {\ul 84.35$\pm$4.59} & 85.29$\pm$2.73 & 81.81$\pm$3.37 \\
			Softplus & \textbf{90.13$\pm$0.25} & \textbf{89.94$\pm$1.40} & \textbf{88.31$\pm$0.16} & \textbf{85.46$\pm$1.41} \\
			Exp & {\ul 89.81$\pm$0.18} & 83.72$\pm$2.55 & {\ul 87.16$\pm$0.30} & {\ul 82.39$\pm$2.03} \\ \bottomrule
	\end{tabular}}
\vspace{-0.5em}
\end{table} 

\pa{\textbf{Nonessential EDL settings}.} As summarized in \tabref{ablation}, we evaluate the performance impact of relaxing individual or combined instances of the three nonessential EDL settings.
These settings are denoted as follows:
(1)~$\lambda=1$: the rigid setting of fixing prior weight to the number of classes, as discussed in \secref{relax-dirichlet};
(2)~$\mathcal{L}_\text{var}$: the variance-minimizing optimization loss, described in \secref{simplified-loss};
(3)~$\mathcal{L}_\text{kl}$: the KL-Div-minimizing regularization term, detailed in \secref{kl-div-loss}.
Note that if $\mathcal{L}_\text{kl}$ is retained, Re-EDL (row 8) reverts to the original R-EDL method (row 5). Furthermore, if both $\lambda=1$ and $\mathcal{L}_\text{var}$ are relaxed, R-EDL reverts to \pa{the traditional EDL method} (row 1).
As shown in rows 5, 6, and 7 of \tabref{ablation}, retaining any of the original settings results in a decline in OOD detection performance.
Specifically, when measured by the AUPR score for OOD detection averaged over six datasets, retaining any one of these settings reduces the performance of Re-EDL by 1.73\%, 0.98\%, and 0.34\%, respectively. 
Additionally, a comparison between row 1 and row 8 reveals that when all three settings are relaxed, the performance of Re-EDL surpasses the baseline method by 1.65\% on classification accuracy and 3.14\% on OOD detection AUPR. 
Therefore, relaxing these settings is effective and their combined application further optimizes performance.

\noindent\textbf{Components of essential EDL setting}.
In \secref{essential}, we argue that the essential setting of EDL is replacing softmax with projected probability, which adds an extra parameter $\lambda$ to class scores and employs an output activation function with a more gradual growth rate.
As shown in \tabref{essential}, both modifications help preserve the useful magnitude information of logits, thereby enhancing uncertainty estimation.

\noindent\textbf{Different loss formulations}.
Since we consider the use of projected probability from subjective logic to be the essential EDL setting, we also validate its effectiveness with the cross-entropy (CE) loss function, which is more commonly applied in classification tasks.
As shown in \tabref{loss-form}, using CE to optimize the projected probability instead of the softmax probability also leads to enhanced uncertainty estimation and comparable classification accuracy.

\noindent\textbf{Different evidence functions}.
In addition, we examine the effects of common evidence functions, including ReLU, Softplus, and Exp, as presented in \tabref{activation}.
To ensure a fair comparison, the hyperparameter $\lambda$ is consistently set to 1.
Besides, the input to the Exp function is constrained within the range of -10 to 10 using the clamp function to prevent numerical overflow.
Although Softplus is not the exclusive choice prescribed by subjective logic, it demonstrates superior performance in both classification and OOD detection when compared to ReLU and Exp, supporting our analysis in \secref{essential}.
Specifically, the exponential growth rate characteristic of the Exp function transforms probability distributions into forms resembling one-hot encoding, whereas ReLU crudely truncates all negative logits, both of which result in substantial information loss.

\subsection{More Experiment Results}
Due to space limitations, please refer to \appref{additional-experiment} for additional results in the classical setting, few-shot setting, and ablation study, as well as visualizations of PR and ROC curves, and uncertainty distributions using various metrics.

\section{Conclusion}
\label{conclusion}
\textbf{Summary.} 
We propose Re-EDL, a simplified yet more effective version of EDL, achieved by revisiting the essential and nonessential settings of the traditional method.
Our analysis yields insights in two key aspects.
On one hand, we identify the nonessential settings in traditional EDL, which include:
(1)~Fixing the prior weight parameter, which governs the balance between leveraging the proportion of evidence and its magnitude in deriving predictive scores, to the number of classes;
(2)~The empirical risk of EDL includes a variance-minimizing optimization term which encourages the Dirichlet PDF to approach a Dirac delta function, thereby heightening the risk of model overconfidence;
(3)~EDL's structural risk adopts a  KL-Div-minimizing regularization on non-target evidence, which extends its effect beyond the \pa{intended purpose and contradicts common sense}, hindering uncertainty estimation in most cases.
On the other hand, we identify the essential setting of EDL as the adoption of projected probability, which more effectively preserves the magnitude information of logits than the traditional softmax probability, thereby enhancing uncertainty estimation.
Building on these insights, Re-EDL treats the prior weight as an adjustable hyperparameter instead of fixing it to the class number, and directly optimizes the expectation of the Dirichlet PDF, phasing out both the variance-minimizing optimization term and the regularization on non-target evidence. 
Comprehensive experimental evaluations underscore the efficacy of our method.

\noindent\textbf{Deficiencies and Future directions.} This paper can be extended along several directions below.
(1)~While the crucial role of the prior weight in balancing the trade-off between leveraging the evidence proportion and the magnitude has been elucidated, the underlying mechanism dictating its optimal value warrants further investigation.
(2)~The optimization objective of Re-EDL can be interpreted as optimizing the expected value of the constructed Dirichlet PDF. While principled and effective, it is somewhat coarse. Future work could explore optimization goals considering other statistical properties of Dirichlet PDFs.
(3)~Although Re-EDL deprecates the traditional KL-Div-minimizing regularization, experiments still validate its benefits in certain aspects. 
Exploring regularization that simultaneously enhances generalization and uncertainty estimation is worthwhile.
In brief, we anticipate that Re-EDL, with its impressive simplicity, can establish a new baseline \pa{facilitating the single-forward-pass uncertainty quantification research}.
%(3) Dirichlet-based Uncertainty Methods (DUM), including EDL and R-EDL, employ Dirichlet PDFs to model the distribution of class probabilities. However, neither the projected probability nor the uncertainty mass fully encapsulate the nuances of the constructed Dirichlet distributions. Thus, developing uncertainty metrics more attuned to DUMs warrants further research.

\balance
\bibliographystyle{IEEEtran}
\bibliography{cite-short}

\begin{IEEEbiography}[{\includegraphics[width=1in,height=1.25in,clip,keepaspectratio]{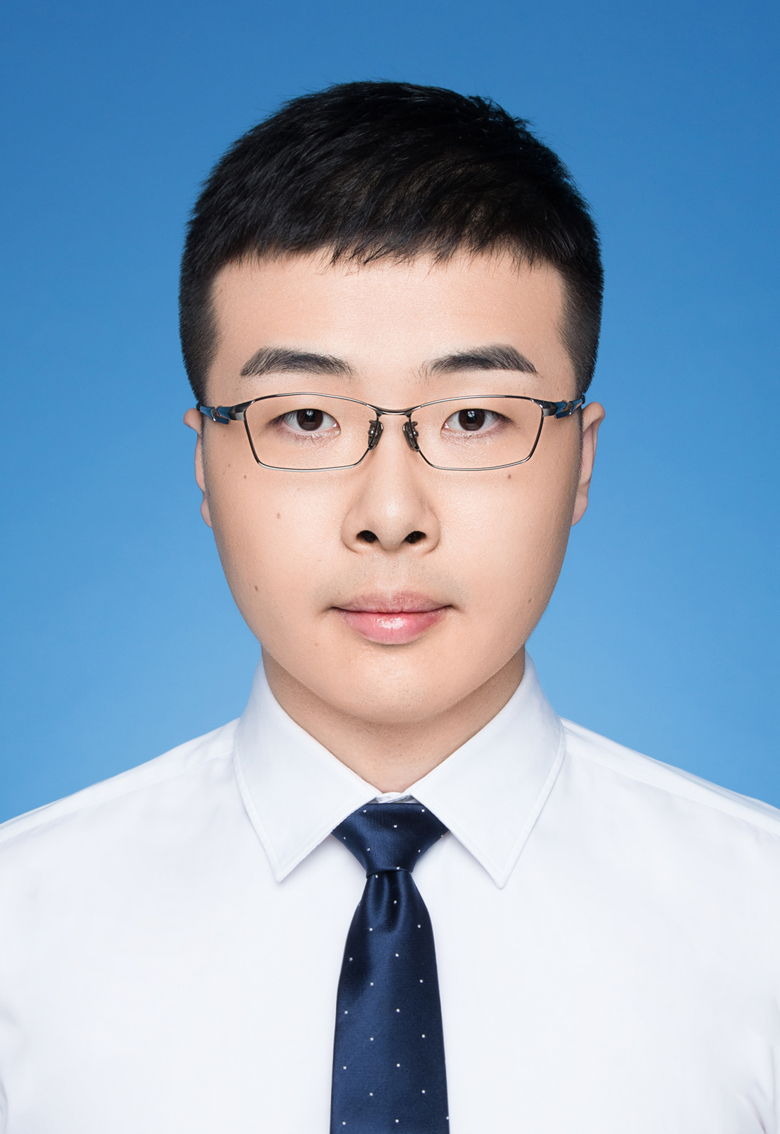}}]
%\begin{IEEEbiographynophoto}
	{Mengyuan Chen} is currently a Ph.D. candidate at the State Key Laboratory of Multimodal Artificial Intelligence Systems, Institute of Automation, Chinese Academy of Sciences, Beijing, China. His research interests include video understanding and uncertainty estimation.
\end{IEEEbiography}

\begin{IEEEbiography}[{\includegraphics[width=1in,height=1.25in,clip,keepaspectratio]{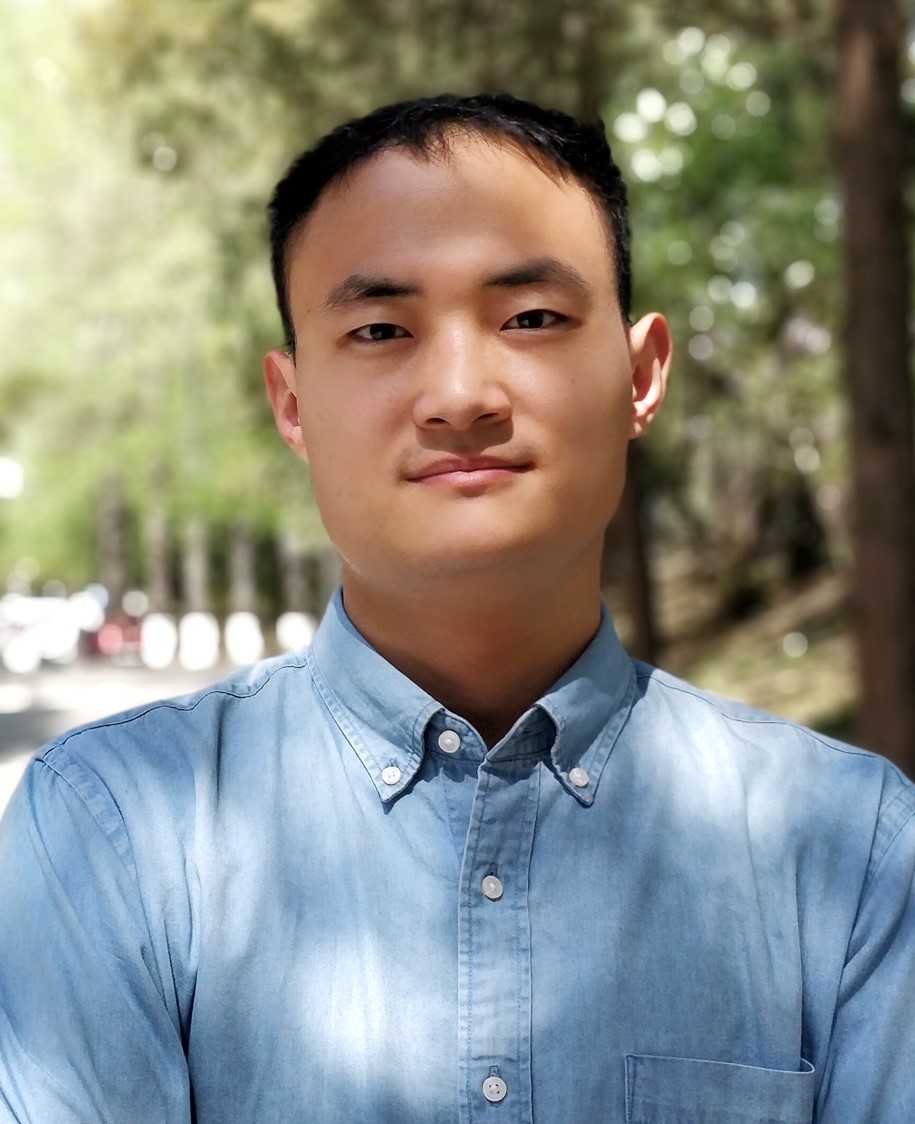}}]
%\begin{IEEEbiographynophoto}
	{Junyu Gao} is an Associate Professor at the Institute of Automation, Chinese Academy of Sciences, Beijing, China. His research interests include computer vision and embodied AI, especially video understanding, vision-and-language navigation, and cross-modal learning.
\end{IEEEbiography}

\begin{IEEEbiography}[{\includegraphics[width=1in,height=1.25in,clip,keepaspectratio]{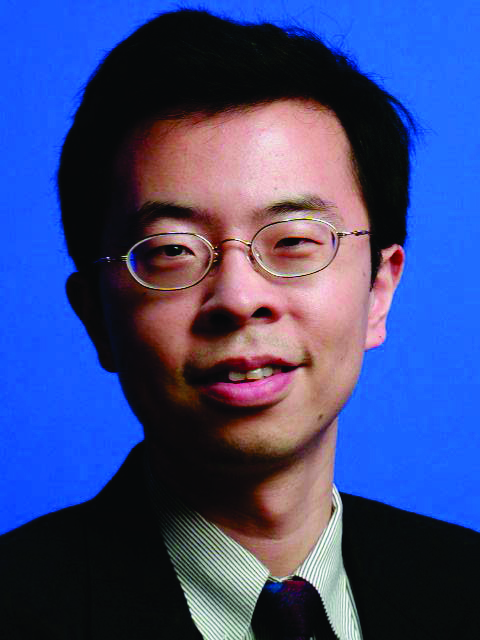}}]
%\begin{IEEEbiographynophoto}
	{Changsheng Xu} (M'97--SM'99--F'14) is a Professor in the State Key Laboratory of Multimodal Artificial Intelligence Systems (MAIS), Institute of Automation, Chinese Academy of Sciences. His research interests include multimedia content analysis/indexing/retrieval, pattern recognition and computer vision. He has hold 50 granted/pending patents and published over 400 refereed research papers in these areas.
%	Dr. Xu has served as associate editor, guest editor, general chair, program chair, area/track chair and TPC member for over 20 IEEE and ACM prestigious multimedia journals, conferences and workshops, including IEEE Trans. on Multimedia, ACM Trans. on Multimedia Computing, Communications and Applications and ACM Multimedia conference.
	He is IEEE Fellow, IAPR Fellow and ACM Distinguished Scientist.
\end{IEEEbiography}

\newpage

%\section*{Acknowledgments}
%This should be a simple paragraph before the References to thank those individuals and institutions who have supported your work on this article.

%{\appendix[Proof of the Zonklar Equations]
%Use $\backslash${\tt{appendix}} if you have a single appendix:
%Do not use $\backslash${\tt{section}} anymore after $\backslash${\tt{appendix}}, only $\backslash${\tt{section*}}.
%If you have multiple appendixes use $\backslash${\tt{appendices}} then use $\backslash${\tt{section}} to start each appendix.
%You must declare a $\backslash${\tt{section}} before using any $\backslash${\tt{subsection}} or using $\backslash${\tt{label}} ($\backslash${\tt{appendices}} by itself
% starts a section numbered zero.)}

\appendices

\section{Proof and Derivation}
\label{append}
This section provides the proof of Theorem 1 and the derivation of optimization objectives of EDL.
%and the derivation of the modified KL-divergence-based regularization term in R-EDL.

\subsection{Proof of Theorem 1}
\label{app:theorem1}

\textbf{Theorem 1 (Bijection between subjective opinions and Dirichlet PDFs).}
Let $X$ be a random variable defined in domain $\mathbb{X}$, and ${\bm \omega}_X=({\bm b}_X, u_X, {\bm a}_X)$ be a subjective opinion. $\bm p_X$ is a probability distribution over $\mathbb{X}$, and a Dirichlet PDF with the concentration parameter $\bm \alpha_X$  is denoted by $\text{Dir}({\bm p_X, \bm \alpha_X})$,
%\begin{equation}
%	\label{dirichlet}
%	\text{Dir}(\bm p_X,\bm \alpha_X)=
%	\frac{\Gamma\left(\sum_{x\in\mathbb{X}}\bm \alpha_X(x)\right)}
%	{\prod_{x\in\mathbb{X}}\Gamma(\bm \alpha_X(x))}
%	\prod_{x\in\mathbb{X}}\bm p_X(x)^{\bm \alpha_X(x)-1},
%\end{equation}
where $\bm \alpha_X(x) \geq 0$, and $\bm p_X(x)\neq0$ if $\bm \alpha_X(x) <1$.
Then, given the base rate $\bm a_X$, there exists a bijection $F$ between the opinion $\bm \omega_X$ and the Dirichlet PDF $\text{Dir}({\bm p_X, \bm \alpha_X})$:
\begin{equation}
	\small
	\label{bijective1-app}
	\begin{aligned}
	F:{\bm \omega}_X&=({\bm b}_X, u_X, {\bm a}_X) \mapsto \\ \text{Dir}({\bm p_X, \bm \alpha_X})&=
	\frac{\Gamma\left(\sum_{x\in\mathbb{X}}\bm \alpha_X(x)\right)}
	{\prod_{x\in\mathbb{X}}\Gamma(\bm \alpha_X(x))}
	\prod_{x\in\mathbb{X}}\bm p_X(x)^{\bm \alpha_X(x)-1},
	\end{aligned}
\end{equation}
where $\Gamma$ denotes the Gamma function, $\bm \alpha_X$ satisfies the following identity that
%\textbf{Theorem 1 (Bijection between subjective opinions and Dirichlet PDFs).}
%Let $X$ be a random variable in $\mathbb{X}$, where $\mathbb{X}$ is a domain consisting of $C$ mutually disjoint values.
%Let ${\bm \omega}_X=({\bm b}_X, u_X, {\bm a}_X)$ be a subjective opinion over the random variable $X$, and let $\bm p_X$ be a probability distribution over $\mathbb{X}$.
%The Dirichlet PDF denoted $\text{Dir}({\bm p_X, \bm \alpha_X})$ is expressed as:
%\begin{equation}
%	\label{dirichlet}
%	\text{Dir}(\bm p_X|\bm \alpha_X)=
%	\frac{\Gamma\left(\sum_{x\in\mathbb{X}}\bm \alpha_X(x)\right)}
%	{\prod_{x\in\mathbb{X}}\Gamma(\bm \alpha_X(x))}
%	\prod_{x\in\mathbb{X}}\bm p_X(x)^{\bm \alpha_X(x)-1},
%\end{equation}
%where $\bm \alpha_X(x) \geq 0$, with the restrictions that $\bm p_X(x)\neq0$ if $\bm \alpha_X(x) <1$.
%Then, if the base rate $\bm a_X$ is given, there exists a bijective mapping $F$ between the opinion $\bm \omega_X$ and the Dirichlet PDF $\text{Dir}({\bm p_X, \bm \alpha_X})$:
%\begin{equation}
%	\label{bijective1-app}
%	F:{\bm \omega}_X=({\bm b}_X, u_X, {\bm a}_X) \mapsto \text{Dir}({\bm p_X, \bm \alpha_X}).
%\end{equation}
%where 
\begin{equation}
	\label{bijective2-app}
	\bm \alpha_X(x) = \frac{\bm b_X(x) W}{u_X} + \bm a_X(x)W,
\end{equation}
$W\in\mathbb{R}_+$ is a given scalar representing a non-informative prior weight.

\textit{Proof.}
The proof of the bijection will be performed in two steps.
First, we will prove a Dirichlet distribution $\text{Dir}(\bm p_X,\bm \alpha_X)$ is uniquely specified by its parameters $\alpha_X$, aka there exists a bijective mapping between $\text{Dir}(\bm p_X,\bm \alpha_X)$ and $\bm \alpha_X$.
Then, we will prove the bijection between the Dirichlet parameters $\bm \alpha_X$ and the subjective opinion $\bm \omega_X$.
Therefore, the bijection between $\bm \omega_X$ and $\text{Dir}(\bm p_X,\bm \alpha_X)$ can be established due to the transitivity of bijection.

Step 1: To prove the mapping $F_1: \bm \alpha_X\mapsto\text{Dir}(\bm p_X,\bm \alpha_X)$ is bijective, we will prove it is both injective and surjective. The surjective property is obvious due to the mapping form. We use proof by contradiction to verify the injectivity as follows.

Assuming that there exists two Dirichlet distributions over the random variable $X$, which are parameterized by two different concentration parameter vectors $\bm \alpha_X$ and $\tilde{\bm\alpha_X}$ respectively, sharing exactly the same probability density function, \ie, there exists $x\in\mathbb{X}$, $\bm \alpha_X(x)\neq\tilde{\bm\alpha}_X(x)$, and for any $x \in \mathbb{X}$ and any $\bm p_X \in \mathcal{S}_{|\mathbb{X}|}$, the equation
\begin{equation}
	\small
	\begin{aligned}
	&\frac{\Gamma\left(\sum_{x\in\mathbb{X}}\bm \alpha_X(x)\right)}
	{\prod_{x\in\mathbb{X}}\Gamma(\bm \alpha_X(x))}
	\prod_{x\in\mathbb{X}}\bm p_X(x)^{\bm \alpha_X(x)-1} \\
	=&
	\frac{\Gamma\left(\sum_{x\in\mathbb{X}}\tilde{\bm\alpha_X}(x)\right)}
	{\prod_{x\in\mathbb{X}}\Gamma(\tilde{\bm\alpha_X}(x))}
	\prod_{x\in\mathbb{X}}\bm p_X(x)^{\tilde{\bm\alpha_X}(x)-1},
	\end{aligned}
\end{equation}
holds, where $\mathcal{S}_{|\mathbb{X}|}$ is a $|\mathbb{X}|$-dimensional unit simplex.
Taking the logarithm of both sides, we have
\begin{equation}
	\small
	\begin{aligned}
	&- \log(B(\bm \alpha_X(x))) + \sum_{x\in\mathbb{X}}(\bm \alpha_X(x) - 1)\log(\bm p_X(x)) \\
	=&
	- \log(B(\tilde{\bm \alpha}_X(x))) + \sum_{x\in\mathbb{X}}(\tilde{\bm \alpha}_X(x) - 1)\log(\bm p_X(x)),
	\end{aligned}
\end{equation}
where $B$ denotes a $|\mathbb{X}|$-dimensional beta function. Therefore, we have the following equation
\begin{equation}
	\small
	\sum_{x\in\mathbb{X}}\left(\bm \alpha_X(x) - \tilde{\bm \alpha}_X(x)\right)\log\left(\bm p_X(x)\right)
	=
	\log\left(\frac{B\left(\bm \alpha_X(x)\right)}{B\left(\tilde{\bm \alpha}_X(x)\right)}\right),
\end{equation}
for any $\bm p_X \in \mathcal{S}_{|\mathbb{X}|}$.
Since the above equation holds for any probability distribution $\bm p_X$, we have
\begin{equation}
	\label{homogenous}
	\sum_{x\in\mathbb{X}}\left(\bm \alpha_X(x) - \tilde{\bm \alpha}_X(x)\right)\log\left(\bm p_X(x) - \bm p^\prime_X(x)\right)
	= 0,
\end{equation}
for any $\bm p_X, \bm p^\prime_X \in \mathcal{S}_{|\mathbb X|}$.
The above equation can be regarded as a homogenous linear equation with $\bm \alpha_X(x) - \tilde{\bm \alpha}_X(x)$ as variables and $\log\left(\bm p_X(x) - \bm p^\prime_X(x)\right)$ as parameters.
Due to the arbitrariness of $\bm p_X$ and $\bm p^\prime_X$, and the property of homogeneous systems of linear equations, we know that \eqnref{homogenous} only has a particular solution, \ie, $\bm \alpha_X(x) - \tilde{\bm \alpha}_X(x)=0$ for any $x\in\mathbb X$, which violates our assumption.

Therefore, $F_1$ is injective and surjective, thus bijective.

Step 2: To prove the bijection between $\bm \omega_X$ and $\bm \alpha_X$, we also need to prove the mapping $F_2: \bm \omega_X\mapsto\bm \alpha_X$ is both injective and surjective. 
Since the base rate $\bm a_X$ and the non-informative prior weight $W$ in \eqnref{bijective2-app} are given, $F_2$ can be simplified to $(\bm b_X, u_X)\mapsto\bm \alpha_X$ with the formulation:
\begin{equation}
	\label{simplified-F2}
	\bm \alpha_X(x) = \frac{\bm b_X(x)}{u_X},\quad \forall x\in\mathbb{X}.
\end{equation}

First, we use proof by contradiction to verify the injection. Assuming that there exists two different sets of belief mass and uncertainty mass which corresponds to the same set of Dirichlet concentration parameters, aka there exists $(\bm b_X, u_X), (\tilde{\bm b}_X, \tilde{u}_X), \bm \alpha_X$, which satisfies
\begin{equation}
	\label{proof-by-contra}
	\bm \alpha_X(x) = \frac{\bm b_X(x)}{u_X} = \frac{\tilde{\bm b}_X(x)}{\tilde{u}_X},\quad \forall x\in\mathbb{X},
\end{equation}
and $\exists x\in\mathbb X$, $\bm b_X(x)\neq \tilde{\bm b}_X(x)$, or $u_X\neq \tilde {u}_X$. We take the summation of \eqnref{proof-by-contra} across all possible values of $x\in\mathbb X$ and utilize the additivity requirement $\sum_{x\in\mathbb{X}} \bm b_{X}(x) + u_X = 1$, then we will have
\begin{equation}
	\sum_{x\in\mathbb{X}}\bm \alpha_X(x) = \frac{1 - u_X}{u_X} = \frac{1 - \tilde{u}_X }{\tilde{u}_X}.
\end{equation}
Thus we reach $u_X=\tilde{u}_X$ and after using the relationship in \eqnref{proof-by-contra}, we will have $\bm b_X(x)=\tilde{\bm b}_X(x)$, $\forall x\in\mathbb X$.
Thereafter, our assumption is violated and thus $F_2$ is injective.

Second, we prove $F_2$ is surjective, aka for any Dirichlet parameter set $\bm \alpha_X$, there exists a set of $(\bm b_X, u_X)$ satisfying \eqnref{simplified-F2}.
By summing \eqnref{simplified-F2} over all values of $x\in\mathbb X$, we obtain the following formulation:
\begin{equation}
	S_X= \frac{1 - u_X}{u_X},
\end{equation}
where $S_X = \sum_{x\in\mathbb{X}}\bm \alpha_X(x)$. By reorganization and substituting $u_X$ into \eqnref{simplified-F2}, we have
\begin{equation}
	u_X= \frac{1}{S_X + 1},\quad \bm b_X(x)= \frac{\bm \alpha_X(x)}{S_X + 1},
\end{equation}
which satisfy all the requirements. Therefore, the mapping $F_2$ is surjective.

Finally, since $F_1$ and $F_2$ are both bijective, $F=F_1\circ F_2$ is also bijective.
\qed

Moreover, in cases of no prior information available, we generally set the base rate $\bm a_X(x)$ as uniform distribution, \ie, $\bm a_X(x)=\frac{1}{|\mathbb X|}$, $\forall x\in\mathbb X$, and \eqnref{bijective2-app} can be reorganized as
\begin{equation}
	\label{bijective3}
	\bm \alpha_X(x) = \left(\frac{\bm b_X(x)}{u_X} + \frac{1}{|\mathbb X|}\right)W,
	\quad\forall x\in\mathbb X,
\end{equation}
or equivalently as 
\begin{equation}
	\label{bijective4}
	\bm b_X(x) = \frac{\bm \alpha_X(x) - W/|\mathbb X|}{\sum_{x^\prime\in\mathbb X}\bm \alpha_X(x^\prime)},
	\quad
	u_X=\frac{W}{\sum_{x\in\mathbb X}\bm \alpha_X(x)},
\end{equation}
by utilizing the additivity condition $\sum_{x\in\mathbb{X}} \bm b_{X}(x) + u_X = 1$.

\rev{Besides, it is noteworthy that comprehensive elaborations on the concepts within the Subjective Logic theory are available in \cite{josang2001logic,josang2016subjective}.}

\subsection{Derivation of Optimization Objectives in EDL}
\label{app:optimization-objective-app}
As aforementioned in \secref{edl}, to perform model optimization, EDL integrates the conventional MSE loss function over the class probability $\bm p_X$ which is assumed to follow the Dirichlet PDF specified in the bijection, thus derives the optimization objective as
\begin{equation}
	\small
	\begin{aligned}
		\mathcal{L}_\text{edl}
		=&\sum_{(\bm z, \bm y)\in \mathcal{D}}\mathbb{E}_{\bm p_X\sim\text{Dir}(\bm p_X, \bm \alpha_X)}\left[\Vert \bm y - \bm p_X\Vert_2^2\right] \\
		=&\sum_{(\bm z, \bm y)\in \mathcal{D}}\mathbb{E}_{\bm p_X\sim\text{Dir}(\bm p_X, \bm \alpha_X)}\sum_{x\in\mathbb X}\left(\bm y_x^2 - 2\bm y_x \bm p_X(x) + \bm p_X(x)^2\right) \\
		=&\sum_{(\bm z, \bm y)\in \mathcal{D}}\sum_{x\in\mathbb X}(\bm y_x^2 - 2\bm y_x \mathbb{E}_{\bm p_X\sim\text{Dir}(\bm p_X, \bm \alpha_X)}[\bm p_X(x)] \\		&+ \mathbb{E}_{\bm p_X\sim\text{Dir}(\bm p_X, \bm \alpha_X)}\left[\bm p_X(x)^2\right] ).
	\end{aligned}
\end{equation}
Using the identity $\mathbb{E}[x^2]=\mathbb{E}[x]^2+\text{Var}[x]$, we know that
\begin{equation}
	\small
	\begin{aligned}
		\mathcal{L}_\text{edl}
		=&\sum_{(\bm z, \bm y)\in \mathcal{D}}\sum_{x\in\mathbb X}\left(\bm y_x -\mathbb{E}_{\bm p_X\sim\text{Dir}(\bm p_X, \bm \alpha_X)}[\bm p_X(x)]\right)^2 \\
		&+ \text{Var}_{\bm p_X\sim\text{Dir}(\bm p_X, \bm \alpha_X)}[\bm p_X(x)].
	\end{aligned}
\end{equation}
Since the Dirichlet distribution has the following properties:
\begin{equation}
	\small
	\begin{aligned}
	\mathbb{E}[\bm p_X(x)]&=\frac{\bm \alpha_X(x)}{S_X},\\
	\text{Var}[\bm p_X(x)]&=\frac{\bm \alpha_X(x)(S_X - \bm\alpha_X(x))}{S_X^2(S_X + 1)},
	\end{aligned}
\end{equation}
where $S_X = \sum_{i=1}^C \bm \alpha_X(x)$, we can explicitly express $\mathcal{L}_\text{edl}$ by $\bm\alpha_X(x)$ and $S_X$ as
\begin{equation}
	\small
	\mathcal{L}_\text{edl}
	=\sum_{(\bm z, \bm y)\in \mathcal{D}}\sum_{x\in\mathbb X}\left(\bm y_x -\frac{\bm \alpha_X(x)}{S_X}\right)^2
	+ \frac{\bm\alpha_X(x)(S_X - \bm\alpha_X(x))}{S_X^2(S_X + 1)}.
\end{equation}

Furthermore, EDL introduces an auxiliary regularization term to suppress the evidence of non-target classes by minizing the Kullback-Leibler (KL) divergence between a modified Dirichlet distribution and a uniform distribution.
This regularization term has demonstrated promising empirical results and has been elucidated by \cite{deng2023uncertainty} using the PAC-Bayesian theory~\cite{mcallester1998some}.
Specifically, the regularization term has the following form:
\begin{equation}
	\small
	\begin{aligned}
		\mathcal{L}_\text{kl}
		=&\frac{1}{|\mathcal{D}|}\sum_{(\bm z, \bm y)\in \mathcal{D}}\text{KL}\left(\text{Dir}(\bm p_X, \tilde{\bm \alpha}_X), \text{Dir}(\bm p_X, \bm 1)\right) \\
		=&\frac{1}{|\mathcal{D}|}\sum_{(\bm z, \bm y)\in \mathcal{D}} \bigg(\log\frac{\Gamma(\tilde{S}_X)}{\Gamma(C)\prod_{x\in\mathbb X}\Gamma\left(\tilde{\bm \alpha}_X(x)\right)} + \\
		&\sum_{x\in\mathbb X}\left(\tilde{\bm \alpha}_X(x)-1\right)
		\left(\psi\left(\tilde{\bm \alpha}_X(x)\right) - \psi(\tilde{S}_X)\right)\bigg),
	\end{aligned}
\end{equation}
where $\bm 1$ denotes a $C$-dimensional ones vector, $\Gamma$ denotes the Gamma function, and $\psi$ denotes the digamma function. $\tilde{\bm \alpha}_X=\bm y + (\bm 1 - \bm y)\odot \bm \alpha_X$ a modified Dirichlet parameter vector whose value of the target class has been set to 1, $\odot$ denotes Hadamard product, and $\tilde{S}_X=\sum_{x\in\mathbb X} \tilde{\bm \alpha}_X(x)$.

Next, we provide the calculation process of the modified KL-divergence-based regularization term $\mathcal{L}_\text{kl}(\lambda)$ in \eqnref{redl-kl}:
\begin{equation}
	\small
	\label{ssl-kl-term-app}
	\begin{aligned}
		&\text{KL}\left(\text{Dir}(\bm P_X, \bm \alpha^\lambda_X), \text{Dir}(\bm P_X, \lambda\cdot\bm 1)\right) \\
		=&\int_{\bm P_X\in\mathcal{N}_C}
		\text{Dir}\left(\bm P_X, \bm \alpha_X^\lambda\right)\log\frac{\text{Dir}\left(\bm P_X, \bm \alpha_X^\lambda\right)}{\text{Dir}\left(\bm P_X, \lambda\cdot\bm 1\right)}d\bm P_X \\
		=&\int_{\bm P_X\in\mathcal{N}_C} \text{Dir}\left(\bm P_X, \bm \alpha_X^\lambda\right)\\
		&\log\bigg(\frac{\Gamma(S^\lambda_X)/\Gamma(C\lambda)}{\prod_{x\in\mathbb X}\Gamma\left(\bm \alpha_X^\lambda(x)\right)/\Gamma(\lambda)}\prod_{x\in\mathbb X} \bm P_X(x)^{\bm \alpha^\lambda_X(x)-\lambda}\bigg)d\bm P_X \\
		=& \log\frac{\Gamma(S^\lambda_X)/\Gamma(C\lambda)}{\prod_{x\in\mathbb X}\Gamma\left(\bm \alpha_X^\lambda(x)\right)/\Gamma(\lambda)} \\
		&+\sum_{x\in\mathbb X}(\bm \alpha_X^\lambda(x)-\lambda)\int_{\bm P_X}\text{Dir}\left(\bm P_X, \bm \alpha_X^\lambda\right)\log \bm P_X(x) d \bm P_X \\
		=& \log\frac{\Gamma(S^\lambda_X)}{\prod_{x\in\mathbb X}\Gamma\left(\bm \alpha_X^\lambda(x)\right)} + \\
		&\sum_{x\in\mathbb X}\left(\bm \alpha_X^\lambda(x)-\lambda\right)
		\left(\psi\left(\bm \alpha_X^\lambda(x)\right) - \psi(S^\lambda_X)\right) + K \\
		\approx& \log\frac{\Gamma(S^\lambda_X)}{\prod_{x\in\mathbb X}\Gamma\left(\bm \alpha_X^\lambda(x)\right)} + \\
		&\sum_{x\in\mathbb X}\left(\bm \alpha_X^\lambda(x)-\lambda\right)
		\left(\psi\left(\bm \alpha_X^\lambda(x)\right) - \psi(S^\lambda_X)\right)
	\end{aligned}
\end{equation}
where $\bm 1$ denotes a $C$-dimensional ones vector, $\mathcal{N}_C$ is a $C$-dimensional unit simplex, $\Gamma$ denotes the gamma function, and $\psi$ denotes the digamma function.
$\bm \alpha_X^\lambda=\lambda\bm y + (\bm 1 - \bm y)\odot \bm \alpha_X$ represents a modified Dirichlet parameter vector whose value of the target class has been set to $\lambda$ instead of $1$ in EDL, $\odot$ denotes Hadamard product, and $S_X^\lambda=\sum_{x\in\mathbb X} \bm\alpha^\lambda_X(x)$.
$K=C\log\Gamma(\lambda) -\Gamma(C\lambda)$ is a scalar which does not affect the optimization result.

\subsection{Derivation for Uncertainty Measures}
\label{app:measures}
This subsection provides the derivation of several uncertainty measures, including expected entropy, mutual information, and differential entropy, of Dirichlet-based uncertainty models. The following content is adapted from the Appendix of \cite{malinin2018predictive} and \cite{deng2023uncertainty}.

\noindent\textbf{Expected Entropy.}
Let $X$ be a random variable defined in $\mathbb{X}$, where $\mathbb{X}$ is a domain consisting of multiple mutually disjoint values. Let $\bm p$ be a probability distribution over $\mathbb{X}$, and let $\text{Dir}(\bm p, \bm \alpha)$ be a Dirichlet distribution parameterized by the concentration parameter vector $\bm \alpha$.
If $X$ represents the category index of an input sample, $x\in\mathbb{X}=\{1,...,C\}$ denotes the value of $X$, satisfying $p(X=x)=\bm p(x)$, then the expected entropy of the random variable $X$ over the Dirichlet distribution $\text{Dir}(\bm p, \bm \alpha)$ can be derived as follows:
\begin{equation}
	\small
	\label{expected-entropy}
	\begin{aligned}
		&\mathbb{E}_{\bm p\sim\text{Dir}(\bm p,\bm \alpha)}[\mathcal{H}[\bm p(x)]] \\
		=&\int_{\bm p\in\mathcal{N}_C} \text{Dir}(\bm p, \bm \alpha)\left(-\sum_{x\in\mathbb X}\bm p(x)\ln \bm p(x)\right)d\bm p \\
%		=&-\sum_{x\in\mathbb X}\int_{\bm p\in\mathcal{N}_C}\text{Dir}(\bm p,\bm \alpha)\left(-\bm p(x)\ln \bm p(x)\right)d\bm p \\
		=&-\sum_{x\in\mathbb X}\int_{\bm p\in\mathcal{N}_C}\frac{\Gamma(S)}{\prod_{x^\prime\in\mathbb{X}}\Gamma(\bm \alpha(x^\prime))} \\
		&\prod_{x^\prime\in\mathbb X}\bm p(x^\prime)^{\bm \alpha(x^\prime)-1}
		\left(-\bm p(x)\ln \bm p(x)\right)d\bm p \\
		=&-\sum_{x\in\mathbb X}\int_{\bm p\in\mathcal{N}_C}\frac{\bm \alpha(x)}{S}\frac{\Gamma(S)}{\Gamma(\bm \alpha(x) + 1)\prod_{x^\prime\neq x}\Gamma(\bm \alpha(x^\prime))}\\
		&\prod_{x^\prime\neq x}\bm p(x^\prime)^{\bm \alpha(x^\prime)-1}
		\bm p(x)^{\bm \alpha(x)}\ln \bm p(x)d\bm p \\
		=&-\sum_{x\in\mathbb X}\frac{\bm \alpha(x)}{S}\int_{\bm p\in\mathcal{N}_C}\mathbb{E}_{\bm p \sim\text{Dir}(\bm p, \bm\alpha + \bm 1_x)}[\ln \bm p(x)]d\bm p \\
		=&-\sum_{x\in\mathbb X}\frac{\bm \alpha(x)}{S}\left(\psi(\bm \alpha(x) + 1) - \psi(S + 1)\right),
	\end{aligned}
\end{equation}
where $S=\sum_{x\in\mathbb X}\bm \alpha(x)$, $\mathcal{N}_C$ is a $C$-dimensional unit simplex, $\psi$ denotes the digamma function, and $\bm 1_x$ denotes a one-hot vector with the $x$-th element being set to 1.
The last third equation comes from the property of Gamma function that $\Gamma(n)=(n-1)!$.
In some literature, the expected entropy is used to measure the \textit{data uncertainty}.

\noindent\textbf{Mutual Information.}
In the Dirichlet-based uncertainty methods, the mutual information between the labels $\bm y$ and the class probability $\bm p$, which can be regarded as the difference between the total amount of uncertainty and the data uncertainty, can be approximately computed as:
\begin{equation}
	\small
	\begin{aligned}
		&\underbrace{I[\bm y, \bm p]}_{\text{Distributional Uncertainty}} \\
		\approx& \underbrace{\mathcal{H}\left[\mathbb{E}_{\bm p\sim\text{Dir}(\bm p, \bm \alpha)}[\bm p(x)]\right]}_{\text{Total Uncertainty}} - \underbrace{\mathbb{E}_{\bm p\sim\text{Dir}(\bm p, \bm \alpha)}\left[\mathcal{H}[\bm p(x)]\right]}_{\text{Expected Data Uncertainty}} \\
		=&-\sum_{x\in\mathbb X}\frac{\bm \alpha(x)}{S}\ln\frac{\bm \alpha(x)}{S} + \sum_{x\in\mathbb X}\frac{\bm \alpha(x)}{S}\left(\psi(\bm \alpha(x) + 1)- \psi(S + 1)\right) \\
		=&-\sum_{x\in\mathbb X}\frac{\bm \alpha(x)}{S}\left(\ln\frac{\bm \alpha(x)}{S} - \psi(\bm \alpha(x) + 1) + \psi(S + 1)\right).
	\end{aligned}
\end{equation}
The calculation of the expected data uncertainty utilizes the result of \eqnref{expected-entropy}. The mutual information is often used to measure the \textit{distributional uncertainty}.

\noindent\textbf{Differential Entropy.}
The derivation of the differential entropy of the Dirichlet distribution is given by:
\begin{equation}
	\small
	\begin{aligned}
		&\mathcal{H}[\text{Dir}(\bm p, \bm \alpha)] \\
		=& - \int_{\bm p\in\mathcal{N}_C} \text{Dir}(\bm p, \bm \alpha) \ln \text{Dir}(\bm p, \bm \alpha) d\bm p \\
		=& - \int_{\bm p\in\mathcal{N}_C}\text{Dir}(\bm p, \bm \alpha)\bigg(\ln \Gamma(S)-\sum_{x\in\mathbb X}\Gamma(\bm \alpha(x)) \\
		&+\sum_{x\in\mathbb X}(\bm \alpha(x)-1)\ln \bm p(x)\bigg)d\bm p \\
		=&\sum_{x\in\mathbb X}\ln\Gamma(\bm \alpha(x)) - \ln \Gamma(S) - \sum_{x\in\mathbb X}(\bm \alpha(x) - 1)\mathbb{E}_{\bm p\sim\text{Dir}(\bm p, \bm \alpha)}[\ln \bm p (x)] \\
		=&\sum_{x\in\mathbb X}\ln\Gamma(\bm \alpha(x)) - \ln \Gamma(S) - \sum_{x\in\mathbb X}(\bm \alpha(x) - 1)(\psi(\bm \alpha(x) - \psi(S))).
	\end{aligned}
\end{equation}
Differential entropy is also a prevalent measure of \textit{distributional uncertainty}. A lower entropy indicates that the model yields a sharper distribution, whereas a higher value signifies a more uniform Dirichlet distribution.

\section{Supplementary Introduction}

\subsection{Uncertainty Reasoning Frameworks}
\label{app:comparison}

\textbf{Comparison with Dempster-Shafer Theory (DST)~\cite{shafer1976mathematical}.} The DST, often referred to as evidence theory, was initially introduced by Dempster within the realm of statistical inference~\cite{dempster2008upper}, and Shafer later expanded this theory into a comprehensive framework for representing epistemic uncertainty~\cite{shafer1976mathematical}.
DST has been pivotal in shaping subjective logic by challenging the traditional additivity principle of probability theory.
Specifically, DST allows the sum of probabilities for all mutually exclusive events to be less than one.
This feature enables both DST and subjective logic to explicitly represent uncertainty about probabilities by allocating belief mass to the entire domain.
The difference between DST and subjective logic is that, subjective logic encourages the evidence distribution of samples with high uncertainty to fall back onto a prior, while DST does not include a flexible base rate representing the prior distribution.

\noindent\textbf{Comparison with Imprecise Dirichlet Model (IDM)~\cite{walley1996inferences}.}  The IDM for multinomial variables derives upper and lower probabilities by adjusting the minimum and maximum base rates in the Beta/Dirichlet PDF for each possible value within the domain.
Unlike subjective logic, which employs a prior weight to influence the base rate's effect, IDM creates an interval of expected probabilities by setting the base rate to its maximum (equal to one) for upper probabilities, and to zero for lower probabilities.
It is important to note, however, that the intervals provided by IDM are not strictly bounded, meaning the actual probabilities may fall outside these estimated ranges.

\noindent\textbf{Comparison with Fuzzy Logic~\cite{hajek2013metamathematics}.}
In Fuzzy Logic, variables are defined by terms that have imprecise and partially overlapping meanings. For instance, when considering the variable temperature, potential values might include ``Low (\qty{0}{\degreeCelsius} to \qty{20}{\degreeCelsius})", ``Medium (\qty{15}{\degreeCelsius} to \qty{30}{\degreeCelsius})", and ``High (\qty{25}{\degreeCelsius} to \qty{40}{\degreeCelsius})".
Despite the inherent fuzziness of these values, temperature can still be represented in an exact and crisp manner using a \textit{fuzzy membership function}. For example, one could state ``The temperature is 0.3 Low and 0.7 Medium", which quantitatively expresses the degree to which the temperature belongs to each vague category.
Conversely, in subjective logic, values are inherently crisp, but subjective opinions incorporate an uncertainty mass to capture ambiguity.
Fuzzy logic and subjective logic address different aspects of uncertainty, and there is potential to integrate these two approaches by representing fuzzy membership functions using subjective opinions~\cite{josang2008continuous}.

\noindent\textbf{Comparison with Kleene’s Three-Valued Logic~\cite{fitting1994kleene}.}
In Kleene’s Three-Valued Logic, propositions are categorized as either TRUE, FALSE, or UNKNOWN.
A significant limitation of this system is that it broadly labels all non-absolute propositions as UNKNOWN, without providing a detailed quantification of the degree of uncertainty.
According to this logic, if two propositions \( x_1 \) and \( x_2 \) are both classified as UNKNOWN, their conjunction \( (x_1 \wedge x_2) \) is also deemed UNKNOWN.
This approach leads to a clear issue when dealing with the conjunction of numerous UNKNOWN propositions.
Specifically, if \( n \) (a sufficiently large number) propositions \( x_1, x_2, \ldots, x_n \) are all UNKNOWN, their conjunction \( (x_1 \wedge x_2 \wedge \ldots \wedge x_n) \) will also be considered UNKNOWN, even if the probability of the conjunction being TRUE is extremely low.
For example, it is reasonable to classify the proposition ``the next coin toss will result in heads" as UNKNOWN, but it is counterintuitive to label ``the outcome of the next 1,000 coin tosses will all be heads" as UNKNOWN, which should be (almost) FALSE.
Subjective logic addresses this paradox effectively. When a series of vacuous opinions are combined, the resulting base rate diminishes towards zero, which in turn minimizes the projected probability.
This mechanism ensures that highly unlikely conjunctions are appropriately identified as being close to FALSE.

\subsection{Datasets}
\label{app:datasets}
Following \cite{deng2023uncertainty}, we conduct experiments on the following groups of image classification dataset: 
(1) \textbf{MNIST}~\cite{lecun1998mnist}, \textbf{FMNIST}~\cite{xiao2017fashion}, \textbf{KMNIST}~\cite{clanuwat2018deep}; 
(2) \textbf{CIFAR-10}~\cite{krizhevsky2009learning}, \textbf{SVHN}~\cite{netzer2018street}, \textbf{CIFAR-100}~\cite{krizhevsky2009learning}, \textbf{GTSRB}~\cite{stallkamp2012man}, \textbf{LFWPeople}~\cite{huang2008labeled}, \textbf{Places365}~\cite{zhou2017places}, \textbf{Food-101}~\cite{bossard14};
(3) \textbf{mini-ImageNet}~\cite{vinyals2016matching}, \textbf{CUB}~\cite{wah2011caltech}.
Within each group, we designate the first dataset as in-distribution training data, while utilizing the subsequent ones as OOD data.
Moreover, to evaluate the effectiveness of our method on video-modality data, we also conduct an open-set action recognition experiment by taking \textbf{UCF-101}~\cite{soomro2012ucf101} as ID data and \textbf{HMDB-51}~\cite{kuehne2011hmdb} and \textbf{MiT-v2}~\cite{monfort2021multi} as OOD data following \cite{bao2021evidential}.
Below are the detailed introductions of the involved datasets:

\textbf{MNIST}\cite{lecun1998mnist}.
MNIST consists of handwritten digits ranging from 0 to 9.
Specifically, MNIST contains 60,000 training images and 10,000 testing images, which have been normalized to fit into $28\times28$ pixel bounding boxes.
We use the proportion of [0.8, 0.2] to partition the training samples into training and validation sets.

\textbf{FMNIST}\cite{xiao2017fashion}. FashionMNIST is a dataset designed as a more challenging replacement for MNIST. Created by Zalando Research, FMNIST features grayscale images of various clothing items such as shirts, trousers, sneakers, and bags. The dataset is structured similarly to MNIST, containing 60,000 training images and 10,000 testing images, each of which is $28\times28$ pixels in size. We use FMNIST as OOD data when training models on MNIST.

\textbf{KMNIST}\cite{clanuwat2018deep}.
Kuzushiji-MNIST is another drop-in replacement for MNIST, consisting of a training set with 60,000 handwritten Kuzushiji (cursive Japanese) Hiragana characters and a testing set comprising 10,000 ones. Similar to MNIST, the handwritten characters have been processed to fit into $28\times28$ pixel resolution grayscale images. We also use KMNIST as OOD data when using MNIST as ID data.

\textbf{CIFAR-10}\cite{krizhevsky2009learning}.
CIFAR-10 comprises 60,000 $32\times32$ color distributed across 10 distinct classes such as airplanes, birds, cats, ships, and more, with each class containing 6,000 images. Among them, 50,000 are designated for training and the remaining 10,000 for testing. We partition the training images into training and validation sets using a split ratio of [0.95, 0.05].

\textbf{SVHN}\cite{netzer2018street}.
Street View House Numbers dataset consists of digit images of house numbers from Google Street View. Specifically, it contains 73257 digits for training and 26032 digits for testing. We use SVHN as OOD data when training models on CIFAR10.

\textbf{CIFAR-100}\cite{krizhevsky2009learning}.
CIFAR-100 is just like the CIFAR-10, except it has 100 classes containing 600 images each. There are 500 training images and 100 testing images per class. We use CIFAR-100 as OOD data when using CIFAR-10 as ID data.

\textbf{GTSRB}\cite{stallkamp2012man}.
The German Traffic Sign Recognition Benchmark dataset comprises 43 different traffic sign classes, with a total of 39,209 training images and 12,630 test images. These images are taken under various lighting conditions and environments. In our study, we use the testing set as OOD data when using CIFAR-10 as ID data.

\textbf{LFWPeople}\cite{huang2008labeled}.
Labeled Faces in the Wild, a popular face photograph database, contains over 13,000 images of faces collected from the web, each labeled with the name of the person pictured. Of these, 1,680 individuals have two or more distinct photos in the dataset. We use the testing set from the version provided by torchvision as OOD data when using CIFAR-10 as ID data.

\textbf{Places365}\cite{zhou2017places}.
Places365 includes 1.8 million training images and 36,000 validation images, spanning 365 scene categories. The distribution of images across these categories reflects real-world occurrence frequencies. We use the validation set as OOD data when using CIFAR-10 as ID data.

\textbf{Food-101}\cite{bossard14}.
Food-101 comprises 101 food categories, totaling 101,000 images. Each category includes 250 manually reviewed test images and 750 training images. All images were rescaled to a maximum side length of 512 pixels. We utilize the testing set as OOD data when using CIFAR-10 as ID data.

\textbf{mini-ImageNet}\cite{vinyals2016matching}.
This database is designed for few-shot learning evaluation. 
mini-ImageNet comprises 50,000 $84\times84$ color images for training and 10,000 ones for testing, evenly distributed across 100 classes, and these 100 classes are subdivided into sets of 64, 16, and 20 for meta-training, meta-validation, and meta-testing tasks, respectively.

\textbf{CUB}\cite{wah2011caltech}.
The Caltech-UCSD Birds dataset contains 11,788 images of 200 subcategories belonging to birds, 5,994 for training and 5,794 for testing. We use CUB as OOD data when using mini-ImageNet as ID data in the few-shot setting.

\textbf{UCF-101}\cite{soomro2012ucf101}.
UCF-101 is an action recognition data set of realistic action videos, collected from YouTube. Specifically, UCF-101 contains 13320 videos distributed across 101 action categories. For experiments of video-modality setting, we train models on UCF-101 training split and take its testing set as known samples in inference.
Following \cite{bao2021evidential}, despite there exists a few overlapping classes between UCF-101 and the OOD datasets, HMDB-51 and MiT-v2, we do not manually clean the data for standardizing the evaluation.

\textbf{HMDB-51}\cite{kuehne2011hmdb}.
HMDB-51 is collected mostly from movies, and a small proportion from Prelinger archive, YouTube and Google videos. Specifically, HMDB-51 contains 6,849 clips of 51 action categories, each containing a minimum of 101 clips. We use its testing set as unknown samples in the video-modality setting.

\textbf{MiT-v2}\cite{monfort2021multi}.
Multi-Moments in Time has 305 classes and 30,500 testing videos. We also use the testing set as unknown samples for experiments in the video-modality setting.

\section{Additional Experiment Results}
\label{app:additional-experiment}

\subsection{Classical Setting}
\label{app:additional-classical}

In \tabref{classical-ood-aupr-app} and \tabref{classical-ood-auroc-app}, we provide the AUPR and AUROC scores of OOD detection in the classical setting, measured by MP (Max projected probability), UM (Uncertainty Mass), DE (Differential Entropy), and MI (Mutual Information), respectively.
\tabref{ece-app} compares EDL-related works with the temperature scaling method~\cite{guo2017calibration} in the classical setting, including results evaluated by the Expected Calibration Error (ECE) with 15 bins and the Brier score. Although temperature scaling achieves impressive results when evaluated by the ECE metric, there still exists a performance gap with our method on OOD detection ability.

Besides, we believe that employing the AUPR  scores for evaluation purposes aligns more closely with our objectives than using ECE or Brier score. As delineated in \secref{classical}, our primary criterion for assessing uncertainty estimation is the model's ability in differentiating between ID and OOD samples, as well as between correctly classified and misclassified samples.
Despite that ECE is frequently employed to assess the degree of correspondence between the model's confidence and the true correctness likelihood, a confidence distribution accompanied by a low ECE does not inherently ensure a clear distinction between correct and incorrect predictions. For instance, in a balanced two-class dataset, if a binary classifier categorizes all samples into a single class with a consistent confidence output of 50\%, the ECE would be zero, yet this result lacks practical significance.

\begin{table}[]
	\centering
	\caption{Comparison of temperature scaling method with EDL-related works in the classical setting, including results evaluated by Expected Calibration Error (ECE) with 15 bins and Brier score. Downward arrows ($\downarrow$) indicate that lower values correspond to better performance for these metrics.}
	\label{tab:ece-app}
	\renewcommand{\arraystretch}{1.1}
	\small
	\resizebox{\linewidth}{!}{
		\begin{tabular}{@{}c|cc|c|c@{}}
			\toprule
			\multirow{2}{*}{Method} & \multicolumn{2}{c|}{Confidence Calibration} & Mis Detect & OOD Detect \\
			& ECE (15 bins)$\downarrow$ & Brier score$\downarrow$ & AUPR & AURP (Mean) \\ \midrule
			Temp Scale & \textbf{1.06$\pm$0.10} & 18.44$\pm$0.49 & {\ul 98.89$\pm$0.05} & 82.07$\pm$2.23 \\
			EDL & 11.56$\pm$0.93 & 27.34$\pm$0.71 & 98.74$\pm$0.07 & 82.32$\pm$0.98 \\
			$\mathcal{I}$-EDL & 44.35$\pm$1.27 & 59.73$\pm$1.31 & 98.71$\pm$0.11 & 82.01$\pm$1.47 \\
			\textbf{R-EDL} & {\ul 3.47$\pm$0.31} & {\ul 18.15$\pm$0.50} & \textbf{98.98$\pm$0.05} & {\ul 83.73$\pm$1.07} \\
			\textbf{Re-EDL} & 5.72$\pm$0.32 & \textbf{14.95$\pm$0.47} & 98.81$\pm$0.05 & \textbf{85.46$\pm$1.41} \\ \bottomrule
	\end{tabular}}
\end{table}

\subsection{Few-shot Setting}
\label{app:additional-fewshot}
\tabref{fsl-ood-aupr-app} shows few-shot results of OOD detection measured by more uncertainty metrics, \ie, MP (Max projected probability), UM (Uncertainty Mass), DE (Differential Entropy), and MI (Mutual Information).
%\tabref{smooth} compares our method and label smoothing in the few-shot setting.
All results consistently demonstrate the superior OOD detection performance of our proposed method.

\pa{\subsection{Ablation Study}
\label{app:ablation}
Due to the space limitation, here we provide the complete versions of \twotabref{loss-form}{activation} as \twotabref{app-loss-form}{app-activation}.}

%\begin{table*}[]
%	\centering
%	\caption{Implementation details of experiments in the classical setting.}
%	\label{tab:detail-classical}
%	\renewcommand{\arraystretch}{1.1}
%	\small
%	\resizebox{0.8\linewidth}{!}{
%	\begin{tabular}{@{}ccccccc@{}}
%		\toprule
%		ID dataset & OOD dataset & Optimizer & Learning rate & (decay,step) for lr & Max Epoch & $\lambda$ \\ \midrule
%		MNIST & FMNIST, KMNIST & Adam & $1\times10^{-3}$ & (0.1, 15) & 60 & 0.1 \\
%		CIFAR-10 & \begin{tabular}[c]{@{}c@{}}SVHN, CIFAR-100, GTSRB,\\ LFWPeople, Places365, Food101\end{tabular} & Adam & $1\times10^{-4}$ & - & 200 & 0.8 \\ \bottomrule
%	\end{tabular}
%	}
%\end{table*}

\begin{table*}[]
	\centering
	\caption{List of the hyperparameter $\lambda$ of experiments in the few-shot setting.}
	\label{tab:detail-fewshot}
	\renewcommand{\arraystretch}{1.1}
	\small
	\resizebox{0.8\linewidth}{!}{
		\begin{tabular}{@{}c|c|ccc|ccc@{}}
			\toprule
			Method & Setting & 5-Way 1-Shot & 5-Way 5-Shot & 5-Way 20-Shot & 10-Way 1-Shot & 10-Way 5-Shot & 10-Way 20-Shot \\ \midrule
			R-EDL & $\lambda$ & 0.7 & 0.2 & 0.3 & 0.8 & 0.6 & 0.7 \\
			Re-EDL & $\lambda$ & 0.2 & 0.1 & 1.2 & 0.8 & 0.3 & 0.3 \\ \bottomrule
		\end{tabular}
	}
\end{table*}

\begin{table*}[]
	\centering
	\caption{AUROC scores of OOD detection in the classical setting, measured by MP (Max projected probability), UM (Uncertainty Mass), DE (Differential Entropy), and MI (Mutual Information).}
	\label{tab:classical-ood-auroc-app}
	\renewcommand{\arraystretch}{1.1}
	\small
	\resizebox{\textwidth}{!}{
	\begin{tabular}{@{}c|cccc|cccc|cccc@{}}
		\toprule
		\multirow{2}{*}{Method} & \multicolumn{4}{c|}{CIFAR10$\rightarrow$SVHN} & \multicolumn{4}{c|}{CIFAR10$\rightarrow$CIFAR100} & \multicolumn{4}{c}{CIFAR10$\rightarrow$GTSRB} \\
		& MP & UM & DE & MI & MP & UM & DE & MI & MP & UM & DE & MI \\ \midrule
		EDL & 84.91$\pm$1.49 & 84.98$\pm$1.62 & 84.90$\pm$1.51 & 84.97$\pm$1.62 & 84.46$\pm$0.34 & 84.38$\pm$0.42 & 84.48$\pm$0.35 & 84.39$\pm$0.41 & 83.00$\pm$1.00 & 82.94$\pm$0.99 & 83.03$\pm$1.01 & 82.94$\pm$0.99 \\
		$\mathcal{I}$-EDL & 87.44$\pm$1.93 & 87.22$\pm$1.68 & 87.49$\pm$1.96 & 87.24$\pm$1.68 & 83.95$\pm$0.24 & 83.82$\pm$0.26 & 83.99$\pm$0.24 & 83.83$\pm$0.26 & 84.04$\pm$1.76 & 83.73$\pm$1.71 & 84.04$\pm$1.77 & 83.75$\pm$1.71 \\
		\textbf{R-EDL} & {\ul 87.47$\pm$1.23} & {\ul 87.47$\pm$1.24} & {\ul 87.54$\pm$0.96} & {\ul 87.47$\pm$1.24} & {\ul 85.26$\pm$0.36} & {\ul 85.26$\pm$0.35} & {\ul 84.90$\pm$0.45} & {\ul 85.26$\pm$0.35} & {\ul 85.50$\pm$0.78} & {\ul 85.50$\pm$0.79} & {\ul 85.24$\pm$0.63} & {\ul 85.50$\pm$0.79} \\
		\textbf{Re-EDL} & \textbf{89.72$\pm$0.81} & \textbf{92.22$\pm$1.14} & \textbf{91.79$\pm$1.08} & \textbf{92.19$\pm$1.13} & \textbf{85.06$\pm$0.23} & \textbf{86.67$\pm$0.14} & \textbf{86.40$\pm$0.14} & \textbf{86.65$\pm$0.14} & \textbf{87.42$\pm$1.55} & \textbf{89.84$\pm$2.05} & \textbf{89.36$\pm$1.96} & \textbf{89.80$\pm$2.04} \\ \midrule
		\multirow{2}{*}{Method} & \multicolumn{4}{c|}{CIFAR10$\rightarrow$LFWPeople} & \multicolumn{4}{c|}{CIFAR10$\rightarrow$Places365} & \multicolumn{4}{c}{CIFAR10$\rightarrow$Food101} \\
		& MP & UM & DE & MI & MP & UM & DE & MI & MP & UM & DE & MI \\ \midrule
		EDL & 72.55$\pm$3.56 & 72.57$\pm$3.69 & 72.52$\pm$3.57 & 72.57$\pm$3.69 & 84.82$\pm$0.53 & 84.76$\pm$0.67 & 84.86$\pm$0.56 & 84.76$\pm$0.66 & 85.04$\pm$0.60 & 84.99$\pm$0.59 & 85.05$\pm$0.60 & 84.99$\pm$0.59 \\
		$\mathcal{I}$-EDL & 73.24$\pm$1.08 & {\ul 73.30$\pm$1.09} & 73.20$\pm$1.07 & 73.29$\pm$1.09 & 84.06$\pm$0.56 & 83.88$\pm$0.65 & 84.13$\pm$0.60 & 83.89$\pm$0.64 & 84.85$\pm$0.61 & 84.59$\pm$0.52 & 84.88$\pm$0.57 & 84.61$\pm$0.52 \\
		\textbf{R-EDL} & \textbf{75.81$\pm$2.46} & \textbf{75.81$\pm$2.46} & \textbf{75.94$\pm$2.41} & \textbf{75.81$\pm$2.46} & \textbf{85.79$\pm$0.55} & {\ul 85.79$\pm$0.54} & {\ul 85.30$\pm$0.60} & {\ul 85.79$\pm$0.54} & \textbf{85.44$\pm$1.33} & {\ul 85.45$\pm$1.32} & {\ul 85.20$\pm$1.38} & {\ul 85.44$\pm$1.32} \\
		\textbf{Re-EDL} & {\ul 73.63$\pm$4.29} & 73.29$\pm$4.17 & {\ul 73.43$\pm$4.28} & {\ul 73.31$\pm$4.19} & {\ul 85.75$\pm$0.60} & \textbf{87.84$\pm$0.58} & \textbf{87.45$\pm$0.59} & \textbf{87.81$\pm$0.58} & {\ul 85.20$\pm$0.82} & \textbf{86.56$\pm$1.04} & \textbf{86.31$\pm$0.97} & \textbf{86.54$\pm$1.03} \\ \bottomrule
\end{tabular}}
\end{table*}

\begin{table*}[]
	\centering
	\caption{AUPR scores of OOD detection in the few-shot setting, measured by MP (Max projected probability), UM (Uncertainty Mass), DE (Differential Entropy), and MI (Mutual Information).}
	\label{tab:fsl-ood-aupr-app}
	\renewcommand{\arraystretch}{1.1}
	\small
	\resizebox{\textwidth}{!}{
	\begin{tabular}{@{}c|cccc|cccc|cccc@{}}
		\toprule
		\multirow{2}{*}{Method} & \multicolumn{4}{c|}{5-Way 1-Shot} & \multicolumn{4}{c|}{5-Way 5-Shot} & \multicolumn{4}{c}{5-Way 20-Shot} \\
		& MP & UM & DE & MI & MP & UM & DE & MI & MP & UM & DE & MI \\ \midrule
		EDL & 66.78$\pm$0.12 & 65.41$\pm$0.13 & 69.00$\pm$0.12 & 66.11$\pm$0.13 & 74.46$\pm$0.10 & 76.53$\pm$0.14 & 77.40$\pm$0.12 & 76.69$\pm$0.13 & 80.01$\pm$0.10 & 79.78$\pm$0.12 & 80.35$\pm$0.11 & 79.83$\pm$0.12 \\
		$\mathcal{I}$-EDL & 71.79$\pm$0.12 & 74.76$\pm$0.13 & 74.04$\pm$0.13 & 74.70$\pm$0.13 & 82.04$\pm$0.10 & 82.48$\pm$0.10 & 82.30$\pm$0.10 & 82.47$\pm$0.10 & 84.29$\pm$0.09 & 85.40$\pm$0.09 & 85.12$\pm$0.09 & 85.35$\pm$0.09 \\
		\textbf{R-EDL} & {\ul 72.91$\pm$0.12} & {\ul 74.84$\pm$0.13} & {\ul 74.34$\pm$0.13} & {\ul 74.76$\pm$0.13} & \textbf{83.65$\pm$0.10} & \textbf{84.22$\pm$0.10} & \textbf{84.05$\pm$0.10} & \textbf{84.13$\pm$0.10} & {\ul 84.85$\pm$0.09} & {\ul 85.57$\pm$0.09} & {\ul 85.43$\pm$0.09} & {\ul 85.53$\pm$0.09} \\
		\textbf{Re-EDL} & \textbf{73.20$\pm$0.12} & \textbf{75.61$\pm$0.13} & \textbf{75.65$\pm$0.13} & \textbf{75.30$\pm$0.13} & {\ul 83.10$\pm$0.10} & {\ul 83.35$\pm$0.10} & {\ul 83.26$\pm$0.10} & {\ul 83.31$\pm$0.10} & \textbf{85.19$\pm$0.09} & \textbf{86.06$\pm$0.09} & \textbf{85.93$\pm$0.09} & \textbf{86.06$\pm$0.09} \\ \midrule
		\multirow{2}{*}{Method} & \multicolumn{4}{c|}{10-Way 1-Shot} & \multicolumn{4}{c|}{10-Way 5-Shot} & \multicolumn{4}{c}{10-Way 20-Shot} \\
		& MP & UM & DE & MI & MP & UM & DE & MI & MP & UM & DE & MI \\ \midrule
		EDL & 59.19$\pm$0.09 & 67.81$\pm$0.12 & 67.78$\pm$0.12 & 67.84$\pm$0.12 & 71.06$\pm$0.10 & 76.28$\pm$0.10 & 75.74$\pm$0.10 & 76.19$\pm$0.10 & 74.50$\pm$0.08 & 76.89$\pm$0.09 & 76.70$\pm$0.08 & 76.86$\pm$0.09 \\
		$\mathcal{I}$-EDL & {\ul 71.60$\pm$0.10} & {\ul 71.95$\pm$0.10} & {\ul 71.57$\pm$0.10} & {\ul 71.95$\pm$0.10} & 80.63$\pm$0.11 & 82.29$\pm$0.10 & 81.06$\pm$0.09 & 81.96$\pm$0.10 & 81.34$\pm$0.07 & 82.52$\pm$0.07 & 82.16$\pm$0.07 & 82.41$\pm$0.07 \\
		\textbf{R-EDL} & \textbf{72.83$\pm$0.10} & \textbf{73.08$\pm$0.10} & \textbf{73.17$\pm$0.10} & \textbf{73.08$\pm$0.10} & {\ul 82.39$\pm$0.09} & \textbf{83.37$\pm$0.09} & \textbf{82.98$\pm$0.09} & \textbf{83.28$\pm$0.09} & {\ul 82.22$\pm$0.08} & {\ul 82.72$\pm$0.07} & {\ul 82.48$\pm$0.08} & {\ul 82.68$\pm$0.08} \\
		\textbf{Re-EDL} & 71.48$\pm$0.10 & 71.39$\pm$0.09 & 70.84$\pm$0.09 & 71.38$\pm$0.09 & \textbf{82.46$\pm$0.09} & {\ul 83.33$\pm$0.09} & {\ul 82.87$\pm$0.08} & {\ul 83.26$\pm$0.08} & \textbf{82.79$\pm$0.07} & \textbf{84.09$\pm$0.07} & \textbf{83.79$\pm$0.07} & \textbf{84.06$\pm$0.07} \\ \bottomrule
\end{tabular}}
\end{table*}

\begin{table*}[]
	\centering
	\caption{Validation of the essential EDL setting in the cross entropy (CE) loss formulation by simply replacing softmax classification head with projected probability. Note that this is the complete version of \tabref{loss-form}.}
	\label{tab:app-loss-form}
	\renewcommand{\arraystretch}{1.1}
	\small
	\resizebox{0.95\textwidth}{!}{
	\begin{tabular}{@{}c|c|cc|ccccccc@{}}
		\toprule
		\multirow{2}{*}{Loss} & \multirow{2}{*}{Method} & \multicolumn{2}{c|}{CIFAR10} & $\rightarrow$SVHN & $\rightarrow$CIFAR100 & $\rightarrow$GTSRB & $\rightarrow$LFWPeople & $\rightarrow$Places365 & $\rightarrow$Food101 & Mean \\
		&  & Cls Acc & Mis Detect & OOD Detect & OOD Detect & OOD Detect & OOD Detect & OOD Detect & OOD Detect & OOD Detect \\ \midrule
		\multirow{2}{*}{CE} & Softmax & 90.11$\pm$0.19 & \textbf{98.91$\pm$0.06} & 80.05$\pm$3.04 & 85.86$\pm$0.67 & 82.81$\pm$4.17 & 88.17$\pm$3.32 & 68.03$\pm$2.72 & 77.07$\pm$2.26 & 80.33$\pm$2.70 \\
		& Re-EDL & \textbf{90.48$\pm$0.25} & {\ul 98.90$\pm$0.07} & {\ul 85.54$\pm$1.29} & \textbf{88.65$\pm$0.37} & {\ul 88.52$\pm$1.38} & {\ul 89.67$\pm$1.27} & \textbf{74.42$\pm$0.67} & {\ul 80.55$\pm$1.02} & {\ul 84.56$\pm$1.00} \\ \midrule
		\multirow{2}{*}{MSE} & Softmax & 90.10$\pm$0.25 & 98.87$\pm$0.04 & 83.62$\pm$3.29 & 86.54$\pm$0.18 & 82.77$\pm$2.66 & 88.22$\pm$2.28 & 69.57$\pm$2.33 & 77.09$\pm$3.39 & 81.30$\pm$2.36 \\
		& Re-EDL & {\ul 90.13$\pm$0.25} & 98.81$\pm$0.05 & \textbf{89.94$\pm$1.40} & {\ul 88.31$\pm$0.16} & \textbf{90.53$\pm$2.04} & \textbf{89.71$\pm$2.08} & {\ul 73.42$\pm$1.05} & \textbf{80.83$\pm$1.72} & \textbf{85.46$\pm$1.41} \\ \bottomrule
\end{tabular}}
\end{table*}

\begin{table*}[]
	\centering
	\caption{Comparison of common evidence functions. Note that this is the complete version of \tabref{activation}.}
	\label{tab:app-activation}
	\renewcommand{\arraystretch}{1.1}
	\small
	\resizebox{0.95\textwidth}{!}{
	\begin{tabular}{@{}c|cc|ccccccc@{}}
		\toprule
		\multirow{2}{*}{\begin{tabular}[c]{@{}c@{}}Evidence\\ Functions\end{tabular}} & \multicolumn{2}{c|}{CIFAR10} & $\rightarrow$SVHN & $\rightarrow$CIFAR100 & $\rightarrow$GTSRB & $\rightarrow$LFWPeople & $\rightarrow$Places365 & $\rightarrow$Food101 & Mean \\
		& Cls Acc & Mis Detect & OOD Detect & OOD Detect & OOD Detect & OOD Detect & OOD Detect & OOD Detect & OOD Detect \\ \midrule
		ReLU & 83.56$\pm$6.10 & 98.76$\pm$0.14 & {\ul 84.35$\pm$4.59} & 85.29$\pm$2.73 & 85.18$\pm$2.49 & {\ul 88.45$\pm$3.01} & 68.62$\pm$3.81 & {\ul 78.97$\pm$3.57} & 81.81$\pm$3.37 \\
		Softplus & \textbf{90.13$\pm$0.25} & {\ul 98.81$\pm$0.05} & \textbf{89.94$\pm$1.40} & \textbf{88.31$\pm$0.16} & \textbf{90.53$\pm$2.04} & \textbf{89.71$\pm$2.08} & \textbf{73.42$\pm$1.05} & \textbf{80.83$\pm$1.72} & \textbf{85.46$\pm$1.41} \\
		Exp & {\ul 89.81$\pm$0.18} & \textbf{98.83$\pm$0.06} & 83.72$\pm$2.55 & {\ul 87.16$\pm$0.30} & {\ul 86.43$\pm$2.31} & 87.80$\pm$3.63 & {\ul 71.72$\pm$1.31} & 77.50$\pm$2.08 & {\ul 82.39$\pm$2.03} \\ \bottomrule
\end{tabular}}
\end{table*} 

\subsection{Visualization of PR and ROC Curves}
\label{app:curve}
\begin{figure*}
	\centering
	\subfloat[Max Probability]{
		\includegraphics[width=0.24\linewidth]{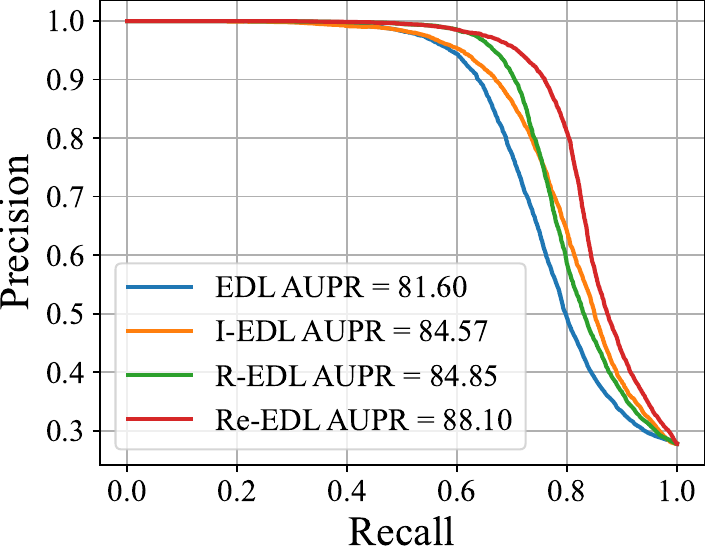}}
	\subfloat[Uncertainty Mass]{
		\includegraphics[width=0.24\linewidth]{graph/aupr/alpha0.pdf}}
	\subfloat[Differential Entropy]{
		\includegraphics[width=0.24\linewidth]{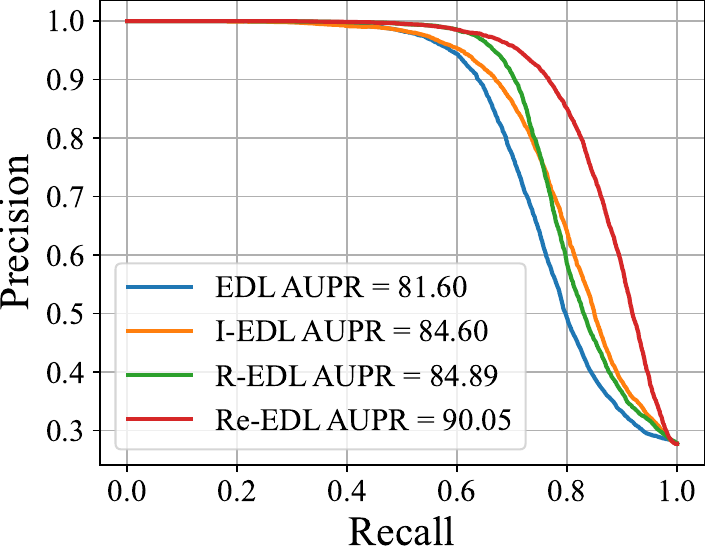}} 
	\subfloat[Mutual Information]{
		\includegraphics[width=0.24\linewidth]{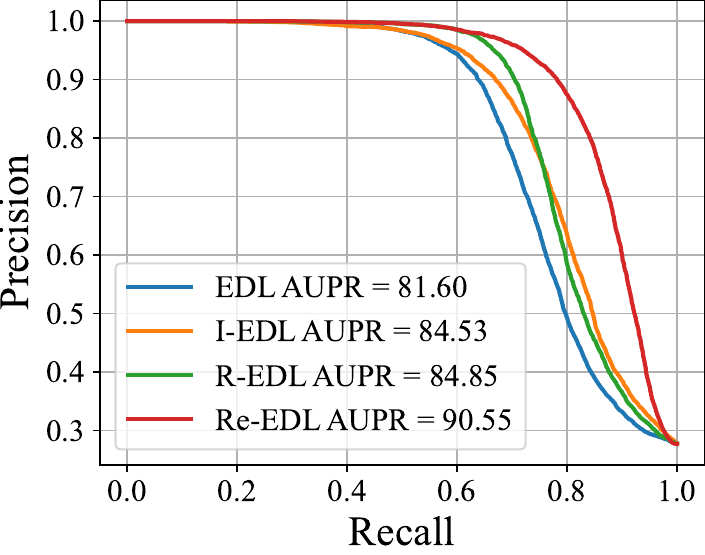}}
	\caption{Precision-Recall (PR) curves of differentiating OOD data (SVHN) from ID data (CIFAR-10).}
	\label{aupr-curve}
\end{figure*}

\begin{figure*}
	\centering
	\subfloat[Max Probability]{
		\includegraphics[width=0.24\linewidth]{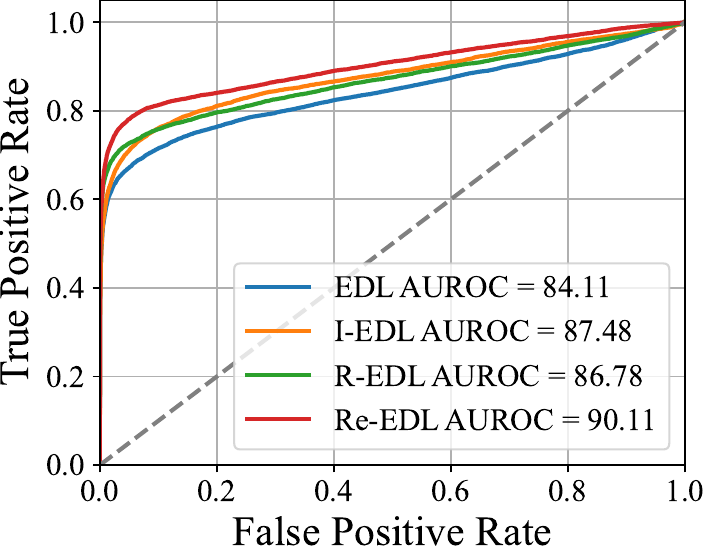}}
	\subfloat[Uncertainty Mass]{
		\includegraphics[width=0.24\linewidth]{graph/auroc/alpha0.pdf}}
	\subfloat[Differential Entropy]{
		\includegraphics[width=0.24\linewidth]{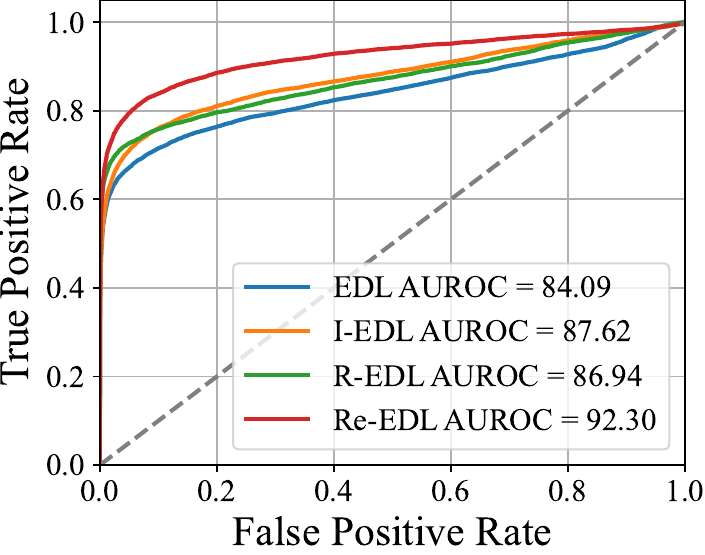}} 
	\subfloat[Mutual Information]{
		\includegraphics[width=0.24\linewidth]{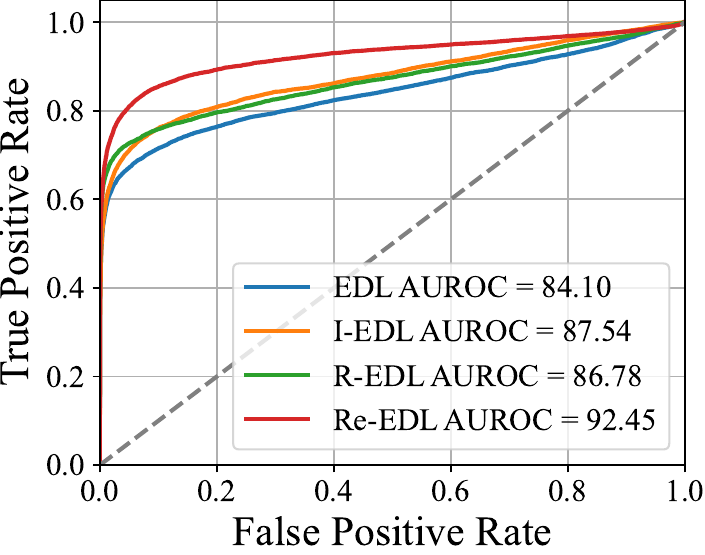}}
	\caption{Receiver Operating Characteristic (ROC) curves of differentiating OOD data (SVHN) from ID data (CIFAR-10).}
	\label{auroc-curve}
\end{figure*}

Figures \firef{aupr-curve} and \firef{auroc-curve} depict the Precision-Recall (PR) and Receiver Operating Characteristic (ROC) curves, respectively, for differentiating OOD data (SVHN) from ID data (CIFAR-10) using four different uncertainty metrics in a classical setting.
Given that we conducted five runs for each method, we selected the run that most closely matched the average performance to plot these curves.
As shown in the figures, $\mathcal{I}$-EDL and R-EDL both outperform the traditional EDL method.
Notably, R-EDL exhibits higher precision at low recall levels, indicating that it tends to make more conservative but accurate predictions when detecting OOD samples.
Our proposed Re-EDL method, meanwhile, demonstrates the best overall performance.

\subsection{Visualization of Uncertainty Distributions}
\label{app:visual}
\begin{figure*}
	\centering
	\subfloat[EDL]{
		\includegraphics[width=0.24\linewidth]{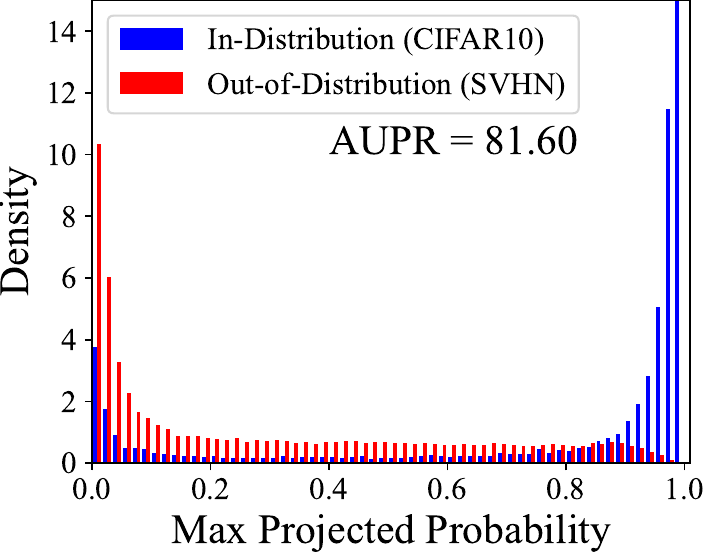}}
	\subfloat[$\mathcal{I}$-EDL]{
		\includegraphics[width=0.24\linewidth]{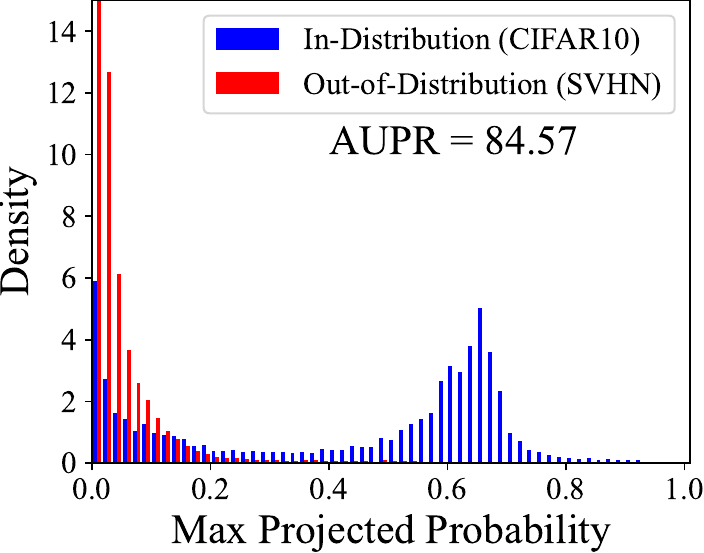}}
	\subfloat[R-EDL]{
		\includegraphics[width=0.24\linewidth]{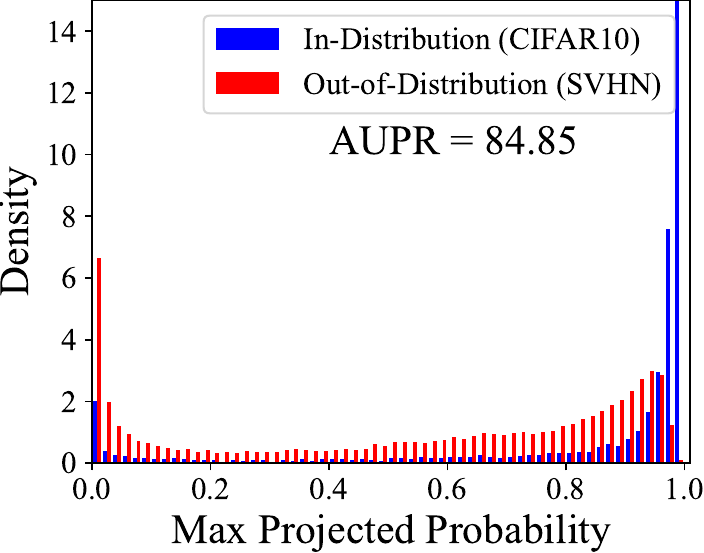}} 
	\subfloat[Re-EDL]{
		\includegraphics[width=0.24\linewidth]{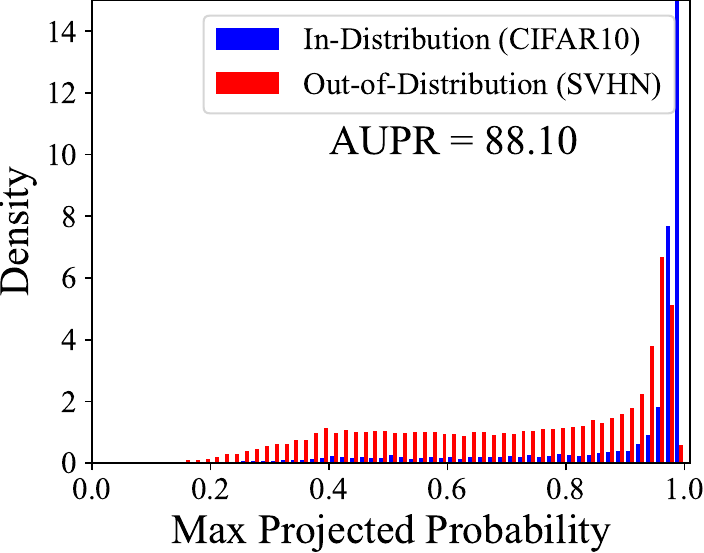}}
	\caption{Uncertainty distribution measured by max projected probability on CIFAR10.}
	\label{visual-1}
\end{figure*}

\fourfigref{visual-1}{visual-2}{visual-3}{visual-4} show density plots of the normalized uncertainty measures for CIFAR-10 against SVHN, and CIFAR-10 against CIFAR-100.
%while \fourfigref{visual-5}{visual-6}{visual-7}{visual-8} show density plots for MNIST against FMNIST, and MNIST against KMNIST.
The uncertainty measures include max projected probability, uncertainty mass, differential entropy, and mutual information.
We apply min-max normalization on each uncertainty value $u$, \ie, $u_\text{norm}=(u - \min u)(\max u - \min u)$.

\begin{figure*}
	\centering
	\subfloat[EDL]{
		\includegraphics[width=0.24\linewidth]{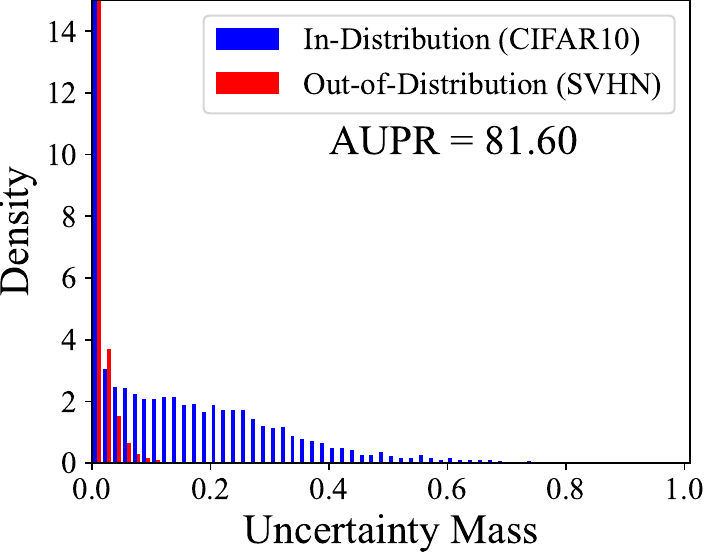}}
	\subfloat[$\mathcal{I}$-EDL]{
		\includegraphics[width=0.24\linewidth]{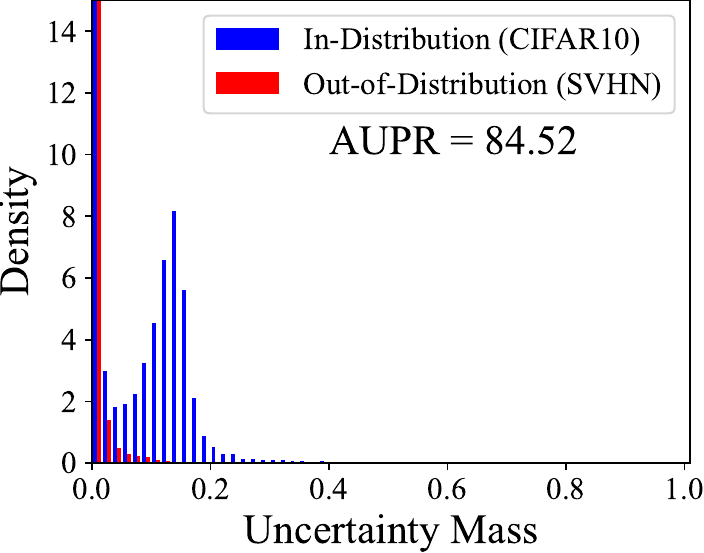}}
	\subfloat[R-EDL]{
		\includegraphics[width=0.24\linewidth]{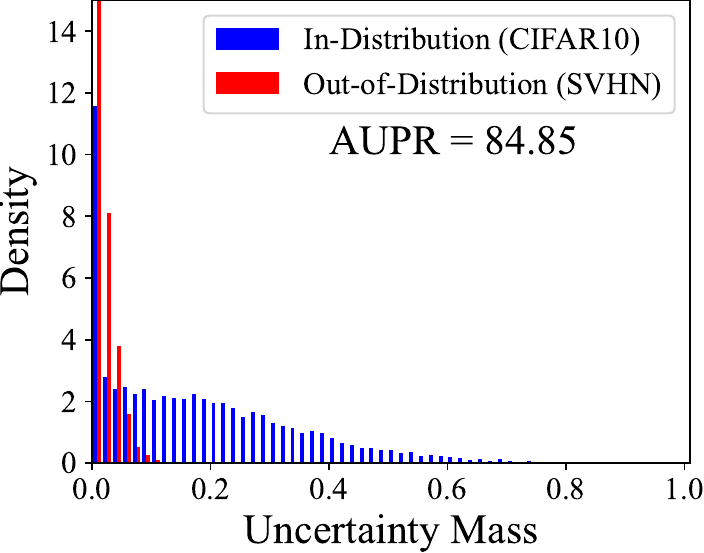}}
	\subfloat[Re-EDL]{
		\includegraphics[width=0.24\linewidth]{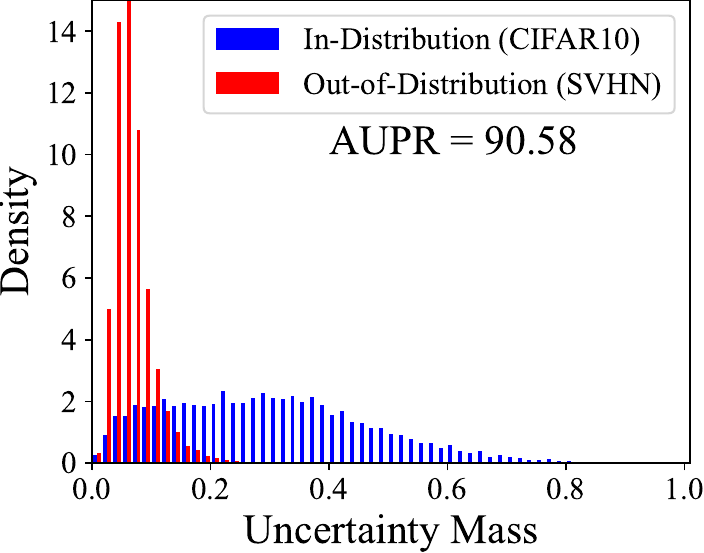}}
	\caption{Uncertainty distribution measured by uncertainty mass on CIFAR10.}
	\label{visual-2}
\end{figure*}

\begin{figure*}
	\centering
	\subfloat[EDL]{
		\includegraphics[width=0.24\linewidth]{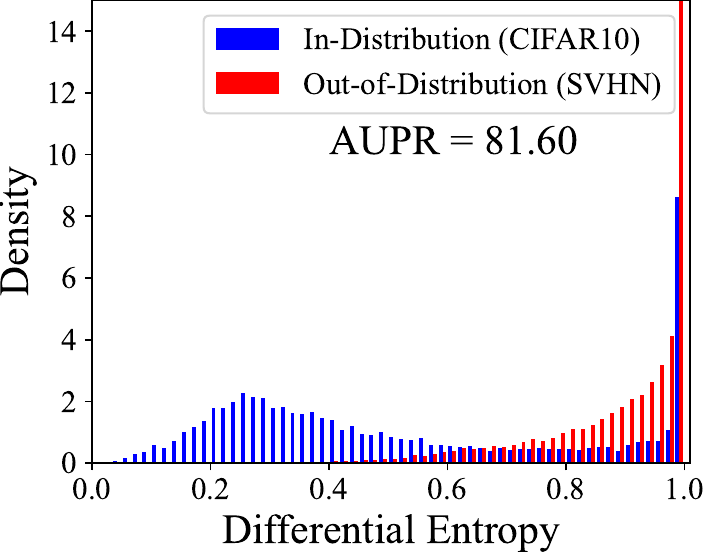}}
	\subfloat[$\mathcal{I}$-EDL]{
		\includegraphics[width=0.24\linewidth]{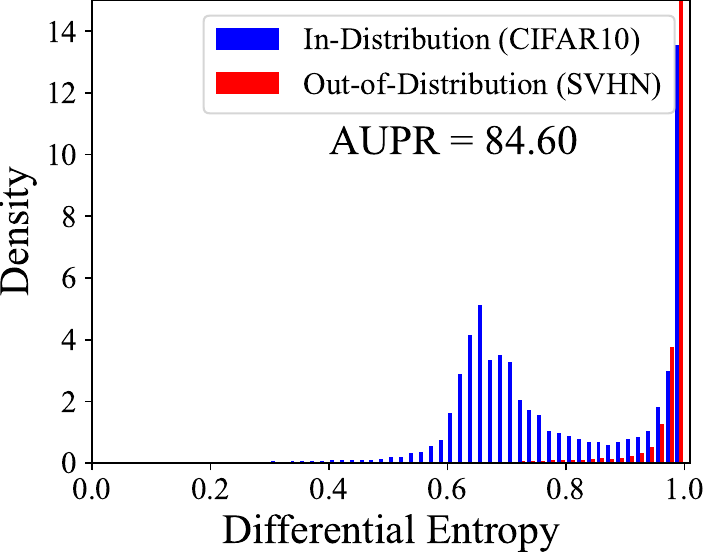}}
	\subfloat[R-EDL]{
		\includegraphics[width=0.24\linewidth]{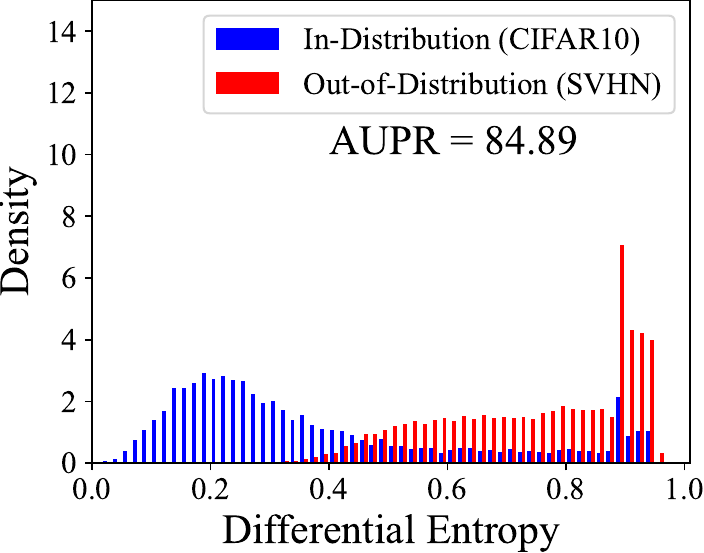}}
	\subfloat[Re-EDL]{
		\includegraphics[width=0.24\linewidth]{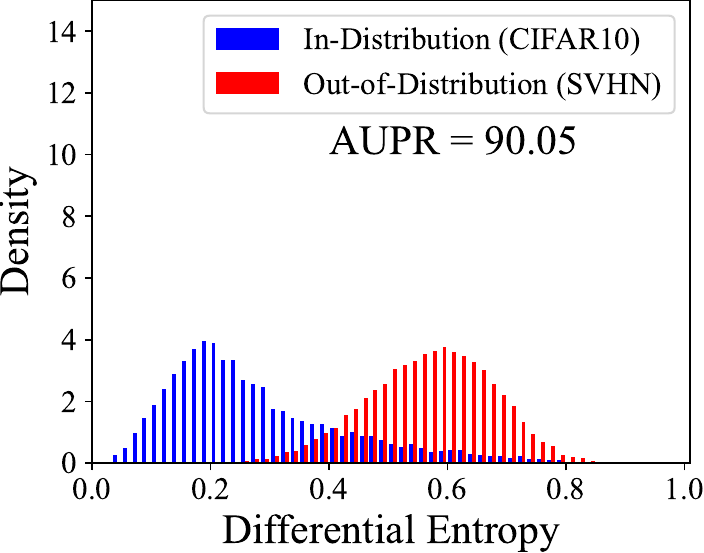}}
	\caption{Uncertainty distribution measured by differential entropy on CIFAR10.}
	\label{visual-3}
\end{figure*}

\begin{figure*}
	\centering
	\subfloat[EDL]{
		\includegraphics[width=0.24\linewidth]{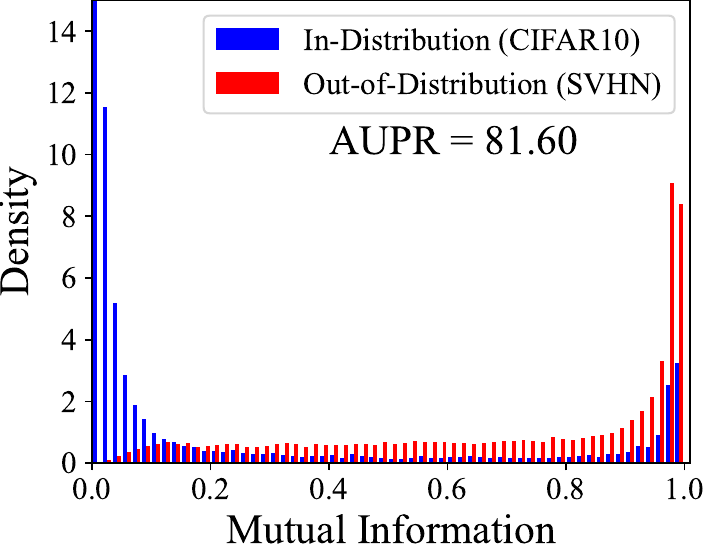}}
	\subfloat[$\mathcal{I}$-EDL]{
		\includegraphics[width=0.24\linewidth]{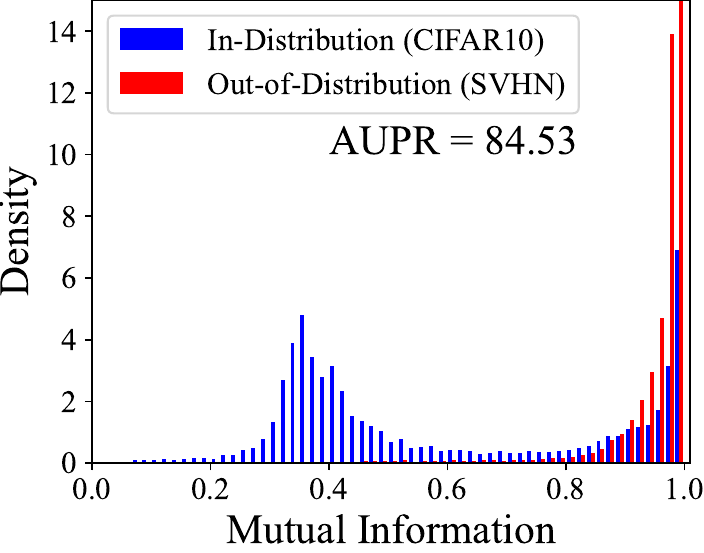}}
	\subfloat[R-EDL]{
		\includegraphics[width=0.24\linewidth]{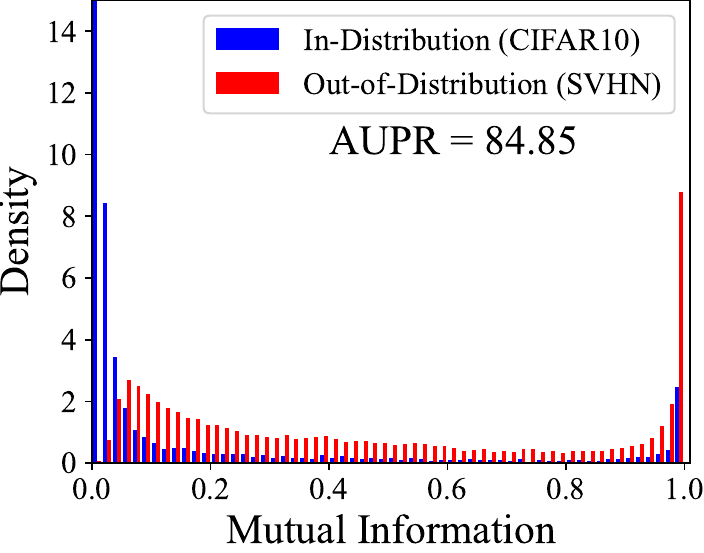}}
	\subfloat[Re-EDL]{
		\includegraphics[width=0.24\linewidth]{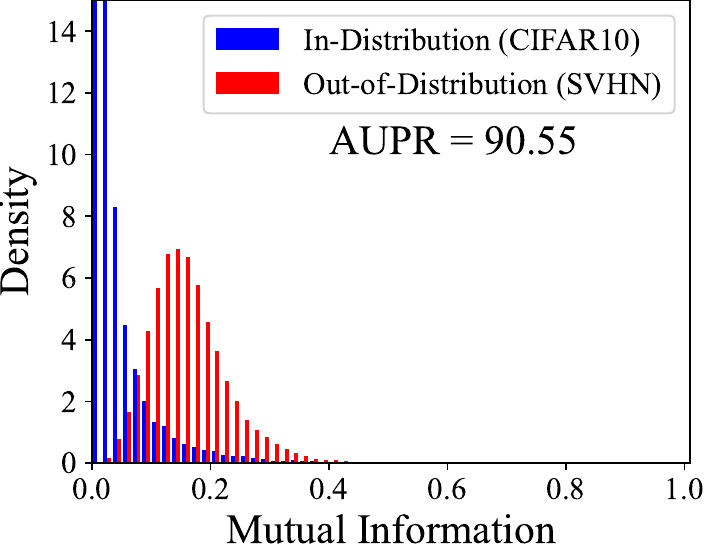}}
	\caption{Uncertainty distribution measured by mutual information on CIFAR10.}
	\label{visual-4}
\end{figure*}

In \firef{visual-1} and \firef{visual-2}, where max projected probability and uncertainty mass are used as measures of uncertainty, we examine the distributions at the far left of the x-axis for each subplot.
The observations indicate that within the EDL, $\mathcal{I}$-EDL, and R-EDL methods, a prevalent issue is the assignment of nearly zero confidence (or extremely high uncertainty) to a substantial number of ID samples.
Consistent with our discussion in \secref{kl-div-loss}, this phenomenon suggests that the adoption of the KL-Div-minimizing regularization excessively suppresses the amplitude of evidence. Consequently, this leads to an overly small total evidence for some ID samples, resulting in excessive uncertainty.
%As a result, these ID samples will be always erroneously identified as OOD data, irrespective of the uncertainty threshold settings.
In contrast, based on R-EDL, Re-EDL substantially avoids this issue by further deprecating the regularization on non-target evidence, thus achieving a significant performance enhancement.
Additionally, in \firef{visual-3} and \firef{visual-4}, R-EDL and Re-EDL also assign relatively lower uncertainty to ID data and thus exhilarates better separability.

The density plots of $\mathcal{I}$-EDL show different shapes with other methods, since $\mathcal{I}$-EDL utilizes the Fisher information matrix to measure the amount of information that the categorical probabilities carry about the concentration parameters of the corresponding Dirichlet distribution, thus allowing a certain class label with higher evidence to have a larger variance.
Consequently, the predictions made by the $\mathcal{I}$-EDL approach are typically less extreme, resulting in a bimodal distribution on the uncertainty density plot where the two peaks are generally closer to the center of the density axis.
%compared to the EDL and R-EDL methods.
Moreover, we deduce that the similarity in the shapes of the uncertainty density plots among EDL, R-EDL, and Re-EDL may stem from the fact that our modifications to EDL only consist of relaxations of nonessential settings, without introducing any extra mechanisms.

\newpage

% your own custom macro containing the $\backslash${\tt{includegraphics}} command to make things
%\bf{If you include a photo:}\vspace{-33pt}
%\begin{IEEEbiography}[{\includegraphics[width=1in,height=1.25in,clip,keepaspectratio]{fig1}}]{Michael Shell}
%Use $\backslash${\tt{begin\{IEEEbiography\}}} and then for the 1st argument use $\backslash${\tt{includegraphics}} to declare and link the author photo.
%Use the author name as the 3rd argument followed by the biography text.
%\end{IEEEbiography}

\end{document}